\documentclass[a4paper,fleqn]{cas-dc}

\usepackage[numbers,sort&compress]{natbib}
\usepackage{float}
\usepackage{orcidlink}
\usepackage{amsmath,amssymb,amsfonts}
\usepackage{bm}
\usepackage{booktabs}
\usepackage{algorithm}
\usepackage{algorithmic}
\usepackage{float}
\usepackage{tabularx}
\usepackage{physics}
\usepackage{caption}
\usepackage{subfigure}
\usepackage{amsmath}
\usepackage{amssymb}
\usepackage{graphicx}

\newlength{\extralength}
\shorttitle{CoInS-Net: Continuous Interpolation-Segmentation Network}
\shortauthors{Sun et al.}

\title[mode=title]{CoInS-Net: A Continuous Position-Aware Network for Joint Medical Image Interpolation and Segmentation}

\author[1]{Yujia Sun\ensuremath{^\dagger}~\orcidlink{0009-0007-8431-9156}}
\ead{yujiasun88@gmail.com}

\author[2]{Ningfeng Que\ensuremath{^\dagger}~\orcidlink{0009-0000-1790-1241}}
\ead{2501112641@stu.pku.edu.cn}

\author[2]{Peiting Shi~\orcidlink{0009-0003-5905-1582}}
\ead{2401112573@stu.pku.edu.cn}

\author[1]{Rongrong Fu~\orcidlink{0009-0008-8879-3370}}
\author[1]{Yingying Yang~\orcidlink{0009-0005-8854-9288}}
\author[1]{Xinhang Li~\orcidlink{0009-0007-3339-3815}}

\author[1]{Yin Dai\ensuremath{^*}~\orcidlink{0000-0001-7812-9750}}
\ead{daiyin@bmie.neu.edu.cn}

\nonumnote{\ensuremath{^\dagger}Yujia Sun and Ningfeng Que contributed equally to this work.}
\nonumnote{\ensuremath{^*}Corresponding author: Yin Dai. E-mail: daiyin@bmie.neu.edu.cn.}

\affiliation[1]{organization={College of Medicine and Biological  Information Engineering, Northeastern University},
    city={Shenyang},
    postcode={110169},
    country={China}}

\affiliation[2]{organization={State Key Laboratory of Natural and Biomimetic Drugs, Department of Biomedical Engineering, College of Future Technology, Peking University},
    city={Beijing},
    postcode={100871},
    country={China}}

\begin{document}
\let\WriteBookmarks\relax
\def\floatpagepagefraction{1}
\def\textpagefraction{.001}

\maketitle

\begin{abstract}
Accurate medical image interpolation and anatomical structure segmentation are fundamental for computer-aided diagnosis and treatment planning. Anisotropic medical volumes with sparse through-plane sampling often suffer from structural discontinuity and boundary blur, hindering reliable clinical image analysis. Most existing methods implement interpolation and segmentation independently, which introduces redundant computation and fails to fully exploit complementary cross-slice structural information between sequential slices. To address these issues, we propose a continuous position-aware interaction network, termed CoInS-Net, for joint frame interpolation and lesion segmentation. Unlike conventional cascaded interpolation-then-segmentation paradigms, the framework enables bidirectional interaction under a shared Swin encoder with continuous spatial coordinate queries. A spatially continuous position interpolation module generates target-position features at every scale from the relative coordinate and physical spacing, and a prototype-based task mutual interaction module lets the segmentation and interpolation branches exchange global structure through a small set of shared prototypes rather than dense feature mixing. A multi-scale task-cooperative decoder further separates each scale into shared and task-specific components, so the two tasks reinforce common anatomy while preserving their distinct requirements down to the boundary level, without extra annotations. Experiments on four public medical imaging datasets with diverse modalities and anatomical regions demonstrate that the proposed method outperforms conventional single-task schemes. It achieves 1.12 to 1.67 dB PSNR gains over the best interpolation baseline and 1.8 to 2.9 percentage point Dice improvements over the best segmentation baseline. The framework also reduces inference latency by nearly half compared with separate single-task pipelines, enhancing the stability and continuity of sequential medical image analysis. The joint optimization framework effectively realizes mutual promotion between interpolation and segmentation tasks, providing a reliable and universal technical scheme for intelligent clinical medical image analysis.

\end{abstract}

\section{Introduction}

Clinical computed tomography and magnetic resonance imaging volumes are typically highly anisotropic, with substantially higher in-plane resolution than through-plane resolution. Sparse sampling along the z-axis weakens inter-slice anatomical continuity. Organ boundaries, small structures, and lesions often appear blurred, displaced, or abruptly changed across adjacent slices, which increases the difficulty of both intermediate-slice reconstruction and target-position segmentation \cite{lam2024,Zhang2018,Song2025}.

When the two tasks are modeled independently, interpolation may produce blurred boundaries or anatomically implausible structures due to the lack of semantic constraints \cite{Uhm2026,isensee2021nature,siddique2021,Cao2023Swin,cao2022mia}. Segmentation models cannot fully exploit structural transitions across adjacent slices. The two tasks are inherently complementary. Segmentation semantics can constrain anatomically plausible image synthesis, while inter-slice variations learned through interpolation can facilitate target-structure recognition. Joint modeling of medical slice interpolation and semantic segmentation is therefore highly desirable.

Existing methods that combine medical slice interpolation with semantic segmentation can be broadly divided into two categories. The first category adopts a serial auxiliary strategy. Intermediate images are first generated through frame interpolation, and the densified data are then used for segmentation. In such methods, interpolation serves as a preprocessing step or a form of data augmentation, while the two tasks remain relatively independent. The interpolation process is difficult to optimize according to segmentation requirements, and generation errors may propagate to the segmentation stage \cite{Wu2022,Cheung2024,Sarmad2023,Huang2006}.

The second category integrates interpolation and segmentation into a unified network, enabling joint optimization through cascaded structures, shared encoders, or feature fusion \cite{zhao2023,Marinov2023Mirror}. Although this design strengthens the association between the two tasks, existing methods still mostly rely on generic feature sharing or use interpolation to assist segmentation in a one-way manner. They lack task-specific bidirectional interaction, failing to fully exploit semantic segmentation constraints to guide image generation and to effectively use cross-slice variations to inform target localization in segmentation. As illustrated in Fig.~\ref{fig:interaction_paradigms}, our method differs from serial auxiliary and conventional joint-learning paradigms by establishing controlled bidirectional interactions between slice interpolation and semantic segmentation.

Simply incorporating interpolation and segmentation into the same network is insufficient to fully exploit their complementarity. The regions where these two tasks most require mutual collaboration are often organ interfaces, lesion boundaries, thin structures, and partial-volume regions with substantial inter-slice variations. In these regions, interpolation requires segmentation semantics to provide anatomical constraints, avoiding excessive boundary smoothing or anatomically implausible structures. Conversely, segmentation requires inter-slice transition information captured during interpolation to enhance structural localization of the target. The key to a joint model therefore lies not merely in feature sharing, but in establishing controllable bidirectional interactions tailored to task-specific requirements.

Such interactive joint modeling still faces several challenges. First, target slice positions are not fixed in anisotropic volumetric data, requiring continuous modeling that incorporates both relative position and physical spacing. Second, the information required by interpolation and segmentation is not completely identical. Simple feature sharing or concatenation may easily introduce negative transfer, so anatomical semantics and inter-slice variation information need to be exchanged selectively. Finally, both tasks rely on boundary details, while additional boundary annotations would reduce the practicality of the method.

To address these issues, we propose CoInS-Net, a continuous position-aware interaction network, for joint medical slice interpolation and semantic segmentation. The network employs a shared Swin Transformer encoder \cite{Liu2021Swin} to extract multi-scale features from adjacent slices, and uses two task-specific decoders to simultaneously predict the image and segmentation mask at an arbitrary target position. Specifically, the spatially continuous position interpolation module models the target position and physical spacing to achieve dynamic fusion of adjacent-slice features at every scale. The prototype-based task mutual interaction module exchanges task-specific global information between interpolation and segmentation through a compact set of sample-adaptive prototypes. The multi-scale task-cooperative decoder selectively exchanges spatial information between the two branches through learned gating mechanisms, jointly enhancing image details and segmentation boundaries without requiring additional boundary annotations.

The main contributions of this paper are summarized as follows:
\begin{enumerate}
    \item We construct a joint framework for arbitrary-position slice interpolation and tumor segmentation in sparse anisotropic medical volumetric data, enabling end-to-end synchronous prediction of intermediate images and their corresponding segmentation masks.
    \item We design the spatially continuous position interpolation module, which generates target-position features at every scale from the relative coordinate and physical spacing, with exact endpoint recovery and input-order symmetry built in by construction.
    \item We propose the prototype-based task mutual interaction module, which performs bidirectional information exchange between interpolation and segmentation through a compact set of sample-adaptive prototypes, avoiding the unstructured mixing and quadratic cost of dense cross-task attention.
    \item We develop the multi-scale task-cooperative decoder, which routes cross-task spatial information through learned sigmoid gates at each upsampling scale, enabling selective feature exchange that refines image details and segmentation boundaries without additional boundary supervision.
\end{enumerate}
\par\smallskip
\noindent
\begin{minipage}{\columnwidth}
    \centering
    \includegraphics[width=\columnwidth]{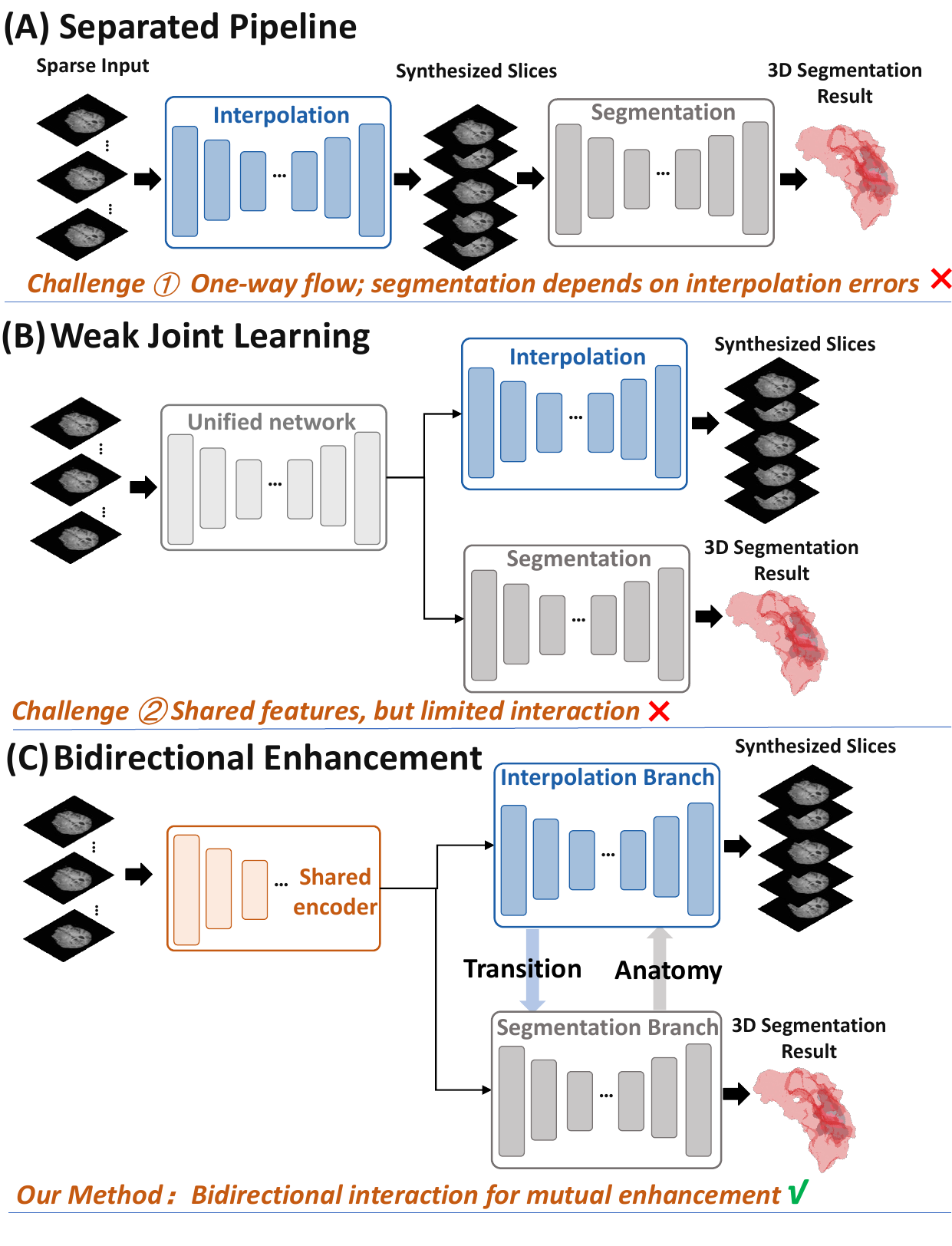}
    \captionof{figure}{Comparison of three interaction paradigms.}
    \label{fig:interaction_paradigms}
\end{minipage}
\par\smallskip

This section reviews prior studies related to slice interpolation for anisotropic medical volumes, volumetric medical image segmentation, and joint learning of slice interpolation and segmentation. The discussion is organized around the three key issues addressed in this work: arbitrary-position prediction, task-specific bidirectional interaction, and collaborative refinement of image details and segmentation boundaries.

\subsection{Slice Interpolation for Anisotropic Medical Volumes}

To alleviate low through-plane resolution of medical volumetric data, recent slice interpolation methods have gradually evolved from fixed-scale reconstruction \cite{Schreibmann2006} toward position- and scale-aware synthesis \cite{zhou2018tmi,Bai2024,Chao2022,Yu2024, que2025adaptive,Guo2020,Uhm2026}. SAINT incorporates voxel spacing into the slice synthesis process, allowing a single model to accommodate different through-plane upsampling factors \cite{Peng2020SAINT}. ArSSR represents three-dimensional medical images as continuous implicit voxel functions and enables arbitrary-scale reconstruction through coordinate queries \cite{Wu2023ArSSR}. SA-INR further models local voxel relationships using spatial attention, allowing continuous sampling at arbitrary slice spacings \cite{Wang2024SAINR}. I$^{3}$Net jointly exploits inter-slice variation, in-plane frequency information, and cross-view features to improve recovery of anatomical structures and high-frequency details in anisotropic medical images \cite{Song2024I3Net}.

These studies demonstrate that arbitrary-scale and continuous-position reconstruction have received considerable attention. Existing methods are primarily designed for image reconstruction. Target coordinates and voxel spacing are generally treated as scale controls, coordinate-query variables, or generation conditions, while the main objective remains image-level reconstruction fidelity. They neither simultaneously predict segmentation results at the target position nor exploit segmentation semantics to constrain the fusion of adjacent-slice features.

The contribution of this work is not merely arbitrary-position slice interpolation. Within a unified interpolation-segmentation framework, we jointly encode the relative target position and physical slice spacing as a continuous position query, which dynamically determines the contributions of the two adjacent slices to the target-position representation at every scale.

\subsection{Medical Volumetric Image Segmentation}

Medical volumetric image segmentation methods commonly employ three-dimensional convolutional networks \cite{isensee2021nature}, Transformer-based architectures \cite{siddique2021,Cao2023Swin,cao2022mia}, or 2.5D models to exploit volumetric context. AFTer-UNet separates in-plane feature extraction from through-plane relationship modeling and combines intra-slice and inter-slice contextual information for segmentation of anisotropic medical volumes \cite{Yan2022AFTerUNet}. Other 2.5D approaches similarly incorporate cross-slice attention or multi-slice feature aggregation to improve utilization of neighboring slices while avoiding the computational cost of fully three-dimensional networks \cite{Zhang2025FM,Zhang2025CT,Yang2026D,Dong2024,wang2024pattern,Shen2023Multi}.

These methods generally assume that the image at the target position has already been observed. Neighboring slices are used to complement the context of the target slice rather than to infer its semantic structure when the target image itself is unavailable. When the target position lies between two sparsely observed slices, organs or lesions may continuously translate, deform, emerge, or disappear along the through-plane direction. Segmentation at such an unobserved position therefore requires more than conventional neighboring-slice aggregation. It also requires explicit modeling of the structural transitions learned during interpolation.

Boundary-aware segmentation constitutes another closely related research direction. Existing methods improve localization of ambiguous structures through auxiliary boundary detection, contour prediction, or signed-distance-map estimation. BA-UNet jointly learns region segmentation and boundary detection and exchanges information between the two tasks \cite{Wang2022BANet}. Shape and boundary-aware multi-branch models further employ signed distance maps to emphasize regions near object boundaries \cite{Liu2022ShapeBoundary}. These studies confirm the importance of boundary information for medical image segmentation. They typically introduce explicit boundary-related targets or auxiliary prediction branches. Frequency information and semantic boundaries are usually modeled only within the segmentation task, without exploiting the correspondence between high-frequency image details and semantic boundary responses across interpolation and segmentation.

\subsection{Joint Learning of Slice Interpolation and Segmentation}

Slice interpolation and semantic segmentation are inherently complementary. Interpolation can recover missing observations in sparsely sampled medical volumes, whereas segmentation semantics can provide anatomical constraints for intermediate-slice synthesis. Existing studies related to these two tasks have mainly adopted serial auxiliary strategies. Slice Imputation generates multiple intermediate images and corresponding labels through frame interpolation, transforms anisotropic volumes into denser representations, and subsequently applies a segmentation network to the augmented data \cite{Wu2022SliceImputation}. This approach demonstrates that intermediate-slice synthesis can improve segmentation of anisotropic medical volumes.

Interpolation and segmentation remain sequentially organized in such pipelines. Errors introduced during image synthesis can propagate to the downstream segmentation stage, while segmentation predictions cannot provide feedback to correct anatomically implausible interpolation results. The multi-task learning component \cite{caruana1997multitask,ruder2017overview,zhao2023} in Slice Imputation is used to jointly optimize auxiliary classification and adversarial discrimination objectives within the synthesis model, rather than to establish bidirectional interaction between dedicated interpolation and segmentation decoders.

Joint reconstruction and segmentation frameworks \cite{Marinov2023Mirror,Dai2025Multi,Solano2025, sun2026tascswinmt,chen2021mia,Marullo2023,Nohel2025Unified,li2022physics} commonly employ shared encoders, cascaded prediction, feature concatenation, or separate task heads \cite{misra2016crossstitch,ma2018modeling,tang2020progressive}. Although these designs strengthen the connection between image reconstruction and semantic prediction, they do not fully consider the different information requirements of slice interpolation and segmentation. Interpolation primarily focuses on intensity variation, texture transition, and structural evolution between adjacent slices, whereas segmentation emphasizes category semantics, anatomical localization, and object boundaries. Unstructured mixing of these features may introduce task-irrelevant information and consequently lead to negative transfer.

Existing methods mostly emphasize the use of reconstructed or interpolated images to improve segmentation, while paying limited attention to how segmentation semantics can conversely constrain image synthesis. In anatomically challenging regions such as organ interfaces, lesion boundaries, thin structures, and areas with substantial inter-slice variation, interpolation requires semantic cues to determine whether a generated structure is anatomically plausible. Conversely, segmentation requires transition cues learned by the interpolation branch to estimate the spatial state of a target structure at an intermediate position.

To address these limitations, our method routes the exchange between the two branches through a compact set of sample-adaptive prototypes, so that segmentation semantics and inter-slice transition cues are communicated selectively rather than through direct and unstructured feature sharing. Furthermore, a task-cooperative decoder routes cross-task spatial information through learned gates at each upsampling scale, allowing the model to improve both image structures and segmentation boundaries without introducing an independent boundary-prediction objective or explicit boundary supervision.

\section{Method}
\label{sec:method}

\subsection{Problem Formulation and Architecture Overview}
\label{subsec:overview}

Given two endpoint slices $\bm{I}_0$ and $\bm{I}_1$ from the same volumetric scan at physical depths $z_0$ and $z_1$, we aim to jointly predict the image $\widehat{\bm{I}}_t$ and segmentation map $\widehat{\bm{P}}_t$ at an arbitrary intermediate position parameterized by
\begin{equation}
t = \frac{z_t - z_0}{z_1 - z_0}, \quad t \in (0, 1).
\label{eq:position}
\end{equation}
Both outputs are produced in a single forward pass, enabling end-to-end mutual optimization between the two tasks.

\begin{figure*}[!tbp]
\centering
\fontsize{7}{9}\selectfont
\setlength{\tabcolsep}{0.1pt}
\includegraphics[width=1.0\textwidth]{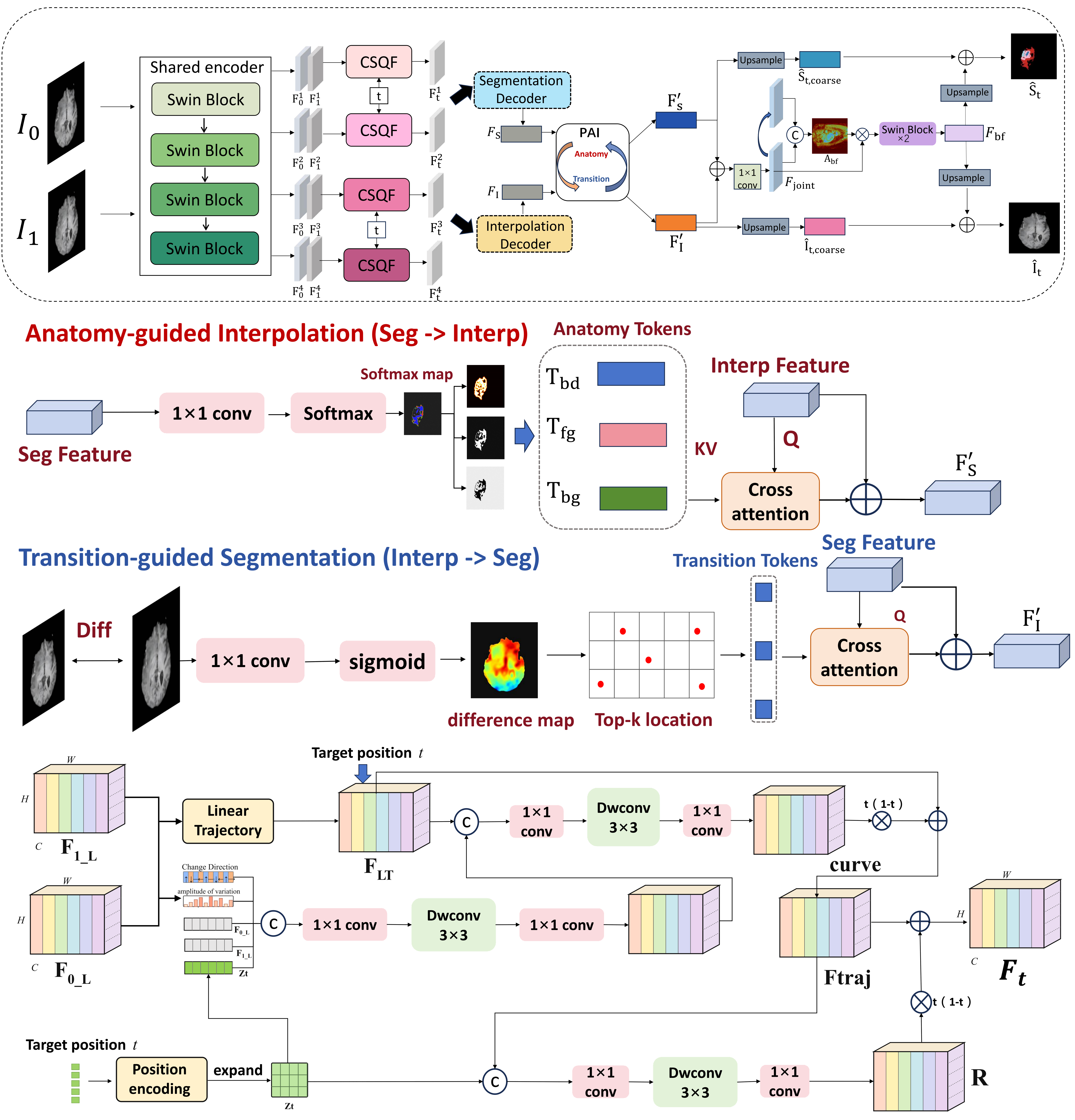}
\caption{Overall Framework of CoInS-Net}
\label{fig:framework}
\end{figure*}

The proposed network, illustrated in Fig.~\ref{fig:framework}, comprises four stages. A weight-shared Swin Transformer encoder extracts multi-scale features from each endpoint independently (Sec.~\ref{subsec:encoder}). The Physical Position-Aware Feature Generator (PPA-IFG) synthesizes target-position features at every scale using continuous coordinate queries and physical spacing awareness (Sec.~\ref{subsec:ppaifg}). The Cross-Task Prototype Interaction module (CTPI) establishes bidirectional category-level information exchange between the interpolation and segmentation branches (Sec.~\ref{subsec:ctpi}). The Multi-Task Collaborative Decoder (MTC-Decoder) recovers spatial details through dual-branch decoding with a local affinity consistency constraint (Sec.~\ref{subsec:decoder}). The training objective is described in Sec.~\ref{subsec:loss}.

\subsection{Shared Multi-Scale Encoder}
\label{subsec:encoder}

We encode each endpoint slice independently through a weight-shared 2D Swin Transformer:
\begin{equation}
\bm{F}_s^l = \mathcal{E}_l(\bm{I}_s), \quad s \in \{0, 1\},\; l = 1, \ldots, L,
\label{eq:encoder}
\end{equation}
where $s$ indexes the endpoint slice and $l$ the feature scale. Weight sharing ensures that corresponding channels across the two endpoints are directly comparable, which is essential for the difference operators and position-conditioned blending used in subsequent modules. Independent encoding also preserves the identity of each endpoint, avoids the positional ambiguity introduced by direct channel concatenation of two slices, and naturally supports input-order swapping during training.

\subsection{Physical Position-Aware Feature Generator}
\label{subsec:ppaifg}

The PPA-IFG module generates continuous intermediate features at every encoder scale, conditioned on both the normalized position $t$ and the physical inter-slice spacing. Unlike video frame interpolation methods that operate at fixed temporal intervals, medical volumes exhibit variable slice thickness across patients and protocols. We therefore introduce an explicit physical spacing signal to distinguish cases where the same relative position $t$ corresponds to different amounts of anatomical change.

\subsubsection{Physical Position Encoding}

We define the normalized spacing ratio
\begin{equation}
\rho = \frac{|z_1 - z_0|}{s_{xy}},
\label{eq:spacing_ratio}
\end{equation}
where $s_{xy}$ is the mean in-plane pixel spacing. A position token is then constructed as
\begin{equation}
\bm{e}_t = \mathrm{MLP}\!\left[\,t,\; 1{-}t,\; t(1{-}t),\; \log(1{+}\rho),\; \gamma(t)\,\right],
\label{eq:pos_token}
\end{equation}
with the sinusoidal component $\gamma(t) = [\sin(\pi t), \cos(\pi t), \sin(2\pi t), \cos(2\pi t)]$. We restrict the frequency to low-order harmonics to avoid unstable fitting on limited medical training data. The polynomial term $t(1{-}t)$ encodes the query's distance from both endpoints and peaks at the midpoint where nonlinear correction is most needed.

\subsubsection{Linear Position Anchor}

At scale $l$, the linear blend of endpoint features provides a baseline that captures slowly varying anatomy:
\begin{equation}
\bm{B}_t^l = (1 - t)\,\bm{F}_0^l + t\,\bm{F}_1^l.
\label{eq:linear_anchor}
\end{equation}
This anchor alone cannot express the nonlinear structural transitions---organ appearance, disappearance, bifurcation, or deformation---that dominate along the through-plane axis of thick-slice acquisitions.

\subsubsection{Dual-Endpoint Local Position-Aware Aggregation}

To capture nonlinear transitions, we aggregate information from local neighborhoods of both endpoint features, guided by the position token. For each spatial location $x$, the query is formed from the anchor and position encoding:
\begin{equation}
\bm{q}_t^l(x) = W_q^l\!\left[\bm{B}_t^l(x),\; \bm{e}_t\right].
\label{eq:query}
\end{equation}
Keys and values are extracted from local offsets $\delta \in \mathcal{N}$ of both endpoints $s \in \{0,1\}$:
\begin{equation}
\bm{k}_{s,\delta}^l(x) = W_k^l\,\bm{F}_s^l(x{+}\delta), \quad
\bm{v}_{s,\delta}^l(x) = W_v^l\,\bm{F}_s^l(x{+}\delta).
\label{eq:kv}
\end{equation}
The attention score incorporates a position-dependent bias that favors the closer endpoint:
\begin{equation}
r_{s,\delta}^l(x) = \frac{{\bm{q}_t^l(x)}^\top \bm{k}_{s,\delta}^l(x)}{\sqrt{d}} + \log(w_s + \epsilon) + b_\delta,
\label{eq:attn_score}
\end{equation}
where $w_0 = 1{-}t$, $w_1 = t$, and $b_\delta$ is a learnable spatial bias. After softmax normalization over all $(s, \delta)$ pairs, the aggregated feature is
\begin{equation}
\bm{A}_t^l(x) = \sum_{s \in \{0,1\}} \sum_{\delta \in \mathcal{N}} a_{s,\delta}^l(x)\,\bm{v}_{s,\delta}^l(x).
\label{eq:aggregation}
\end{equation}
This mechanism jointly encodes position priors, content relevance, and local structural variation, without forcing the medical slice transitions into a 2D optical flow model that is inappropriate for thick-slice acquisitions where structures appear and disappear rather than translate.

\subsubsection{Endpoint-Preserving Feature Generation}

The final position-aware feature at scale $l$ combines the linear anchor with a gated nonlinear correction:
\begin{equation}
\bm{P}_t^l = \bm{B}_t^l + t(1{-}t)\,\Phi_A^l\!\left[\bm{A}_t^l,\; \bm{B}_t^l,\; \bm{F}_1^l {-} \bm{F}_0^l,\; \bm{e}_t\right],
\label{eq:ppa_output}
\end{equation}
where $\Phi_A^l$ is a lightweight convolutional block. The multiplicative gate $t(1{-}t)$ guarantees exact endpoint recovery:
\begin{equation}
\bm{P}_0^l = \bm{F}_0^l, \qquad \bm{P}_1^l = \bm{F}_1^l,
\label{eq:endpoint_recovery}
\end{equation}
so the network learns only the residual that linear blending misses. The gate peaks at $t{=}0.5$, where nonlinear correction is most needed, and naturally attenuates near the endpoints where the linear trend dominates.

\subsection{Task-Specific Feature Adaptation}
\label{subsec:adapter}

The shared position-aware features $\bm{P}_t^l$ are projected into task-specific subspaces through lightweight adapters:
\begin{equation}
\bm{U}_I^l = \mathcal{A}_I^l(\bm{P}_t^l), \qquad
\bm{U}_S^l = \mathcal{A}_S^l(\bm{P}_t^l),
\label{eq:adapters}
\end{equation}
where each adapter consists of a $1{\times}1$ convolution, layer normalization, and GELU activation. This separation establishes distinct feature spaces for the interpolation branch ($\bm{U}_I^l$) and the segmentation branch ($\bm{U}_S^l$), preventing the two tasks from competing for representational capacity in a single shared space while keeping their inputs aligned through the common position-aware backbone.

\subsection{Cross-Task Prototype Interaction}
\label{subsec:ctpi}

The CTPI module is the core contribution of the proposed framework. It enables bidirectional information exchange between the interpolation and segmentation branches at the category level, using compact prototypes as intermediaries rather than dense pixel-to-pixel attention. The module operates in two steps: cross-task prototype construction and bidirectional prototype-conditioned interaction.

\subsubsection{Cross-Task Prototype Construction}

We first obtain a coarse semantic probability map from the segmentation features at scale $l$:
\begin{equation}
\bm{Q}^l = \mathrm{Softmax}\!\left(H_{\mathrm{aux}}^l(\bm{U}_S^l)\right),
\label{eq:coarse_seg}
\end{equation}
where $Q_c^l(x)$ indicates the probability of position $x$ belonging to class $c$. This map serves only to establish spatial category correspondences and is not used as the final segmentation output.

For each class $c$, we aggregate a segmentation semantic prototype and an interpolation appearance prototype using the same probability weights:
\begin{align}
\bm{p}_{S,c}^l &= \frac{\sum_x Q_c^l(x)\,\phi_S^l(\bm{U}_S^l(x))}{\sum_x Q_c^l(x) + \epsilon},
\label{eq:seg_proto} \\
\bm{p}_{I,c}^l &= \frac{\sum_x Q_c^l(x)\,\phi_I^l(\bm{U}_I^l(x))}{\sum_x Q_c^l(x) + \epsilon},
\label{eq:interp_proto}
\end{align}
where $\phi_S^l$ and $\phi_I^l$ are linear projections. Using the same class weights for both prototypes ensures explicit category-level correspondence: $\bm{p}_{S,c}^l$ captures the semantic identity and discriminative structure of class $c$, while $\bm{p}_{I,c}^l$ captures its intensity distribution, texture, and morphological appearance.

These two prototypes are fused into a cross-task joint prototype:
\begin{equation}
\bm{m}_c^l = \psi^l\!\left[\bm{p}_{I,c}^l,\; \bm{p}_{S,c}^l,\; \bm{p}_{I,c}^l \odot \bm{p}_{S,c}^l\right],
\label{eq:joint_proto}
\end{equation}
where $\odot$ denotes element-wise multiplication that highlights channels where both tasks co-activate, and $\psi^l$ is a two-layer MLP. The concatenation preserves complementary information from both tasks while the element-wise product identifies their shared response patterns.

To avoid noise from absent classes, we restrict subsequent operations to the valid class set $\mathcal{V}^l = \{c \mid \sum_x Q_c^l(x) > \kappa\}$. During early training, the coarse probability is stabilized by mixing with downsampled ground-truth labels: $\widetilde{\bm{Q}}^l = \eta\,\bm{Y}_t^l + (1{-}\eta)\,\mathrm{sg}(\bm{Q}^l)$, where $\eta$ anneals from 1 to 0 over the first 20\% of training epochs.

\subsubsection{Bidirectional Prototype-Conditioned Interaction}

The exchange between tasks proceeds in two symmetric directions, each using the other task's prototypes to selectively enhance its own features.

\textit{Segmentation-conditioned interpolation.} Each spatial location of the interpolation feature queries the segmentation semantic prototypes to determine its category affinity:
\begin{equation}
\alpha_{I,c}^l(x) = \mathrm{Softmax}_{c \in \mathcal{V}^l}\!\left(\frac{\mathrm{sim}\!\left(W_{q,I}^l\,\bm{U}_I^l(x),\; W_{k,S}^l\,\bm{p}_{S,c}^l\right)}{\tau}\right),
\label{eq:seg2interp_attn}
\end{equation}
where $\mathrm{sim}(\cdot,\cdot)$ denotes cosine similarity and $\tau$ is a temperature parameter. The aggregated enhancement from the joint prototypes is
\begin{equation}
\bm{C}_I^l(x) = \sum_{c \in \mathcal{V}^l} \alpha_{I,c}^l(x)\,W_{v,I}^l\,\bm{m}_c^l,
\label{eq:seg2interp_agg}
\end{equation}
yielding the enhanced interpolation feature:
\begin{equation}
\widetilde{\bm{U}}_I^l = \Phi_I^l\!\left[\bm{U}_I^l,\; \bm{C}_I^l\right].
\label{eq:enhanced_interp}
\end{equation}
This direction provides anatomical boundary awareness and category-level structural priors to the interpolation branch, reducing over-smoothing across tissue interfaces and preventing unrealistic mixing of different anatomical regions.

\textit{Interpolation-conditioned segmentation.} Symmetrically, the segmentation features query the interpolation appearance prototypes:
\begin{equation}
\alpha_{S,c}^l(x) = \mathrm{Softmax}_{c \in \mathcal{V}^l}\!\left(\frac{\mathrm{sim}\!\left(W_{q,S}^l\,\bm{U}_S^l(x),\; W_{k,I}^l\,\bm{p}_{I,c}^l\right)}{\tau}\right),
\label{eq:interp2seg_attn}
\end{equation}
\begin{equation}
\bm{C}_S^l(x) = \sum_{c \in \mathcal{V}^l} \alpha_{S,c}^l(x)\,W_{v,S}^l\,\bm{m}_c^l,
\label{eq:interp2seg_agg}
\end{equation}
\begin{equation}
\widetilde{\bm{U}}_S^l = \Phi_S^l\!\left[\bm{U}_S^l,\; \bm{C}_S^l\right].
\label{eq:enhanced_seg}
\end{equation}
This direction feeds intensity, texture, and morphological transition cues back into the segmentation branch, improving category recognition in low-contrast regions and enhancing boundary localization for small structures and lesions.

The attention weights $\alpha_{r,c}^l(x)$ dynamically select which prototypes are relevant at each spatial position, while the value projections $W_{v,r}^l\,\bm{m}_c^l$ implicitly determine which channels to transmit. This avoids the quadratic spatial cost of dense cross-attention and provides category-level interpretability: the interaction is mediated by anatomical class prototypes rather than arbitrary spatial correspondences.

\subsection{Multi-Task Collaborative Decoder}
\label{subsec:decoder}

The CTPI module operates on deep, low-resolution features where semantic abstraction is strongest. The MTC-Decoder complements it by maintaining cross-task cooperation during the high-resolution upsampling stages, where boundary precision and fine structural detail are determined.

\subsubsection{Dual-Branch Cooperative Decoding}

Let $\bm{D}_I^{l+1}$ and $\bm{D}_S^{l+1}$ denote the interpolation and segmentation decoder features at scale $l{+}1$. At each scale, a lightweight shared cooperation feature is first constructed:
\begin{equation}
\bm{C}^l = \Phi_C^l\!\left[\mathrm{Up}(\bm{D}_I^{l+1}),\; \mathrm{Up}(\bm{D}_S^{l+1})\right],
\label{eq:coop_feature}
\end{equation}
where $\mathrm{Up}(\cdot)$ denotes bilinear upsampling. Each branch then integrates its own upsampled features, the position-aware skip connection from PPA-IFG, and the shared cooperation signal:
\begin{align}
\bm{D}_I^l &= \Phi_{D,I}^l\!\left[\mathrm{Up}(\bm{D}_I^{l+1}),\; \bm{P}_t^l,\; \bm{C}^l\right],
\label{eq:decode_interp} \\
\bm{D}_S^l &= \Phi_{D,S}^l\!\left[\mathrm{Up}(\bm{D}_S^{l+1}),\; \bm{P}_t^l,\; \bm{C}^l\right],
\label{eq:decode_seg}
\end{align}
where $\Phi_{D,I}^l$ and $\Phi_{D,S}^l$ are task-specific convolutional blocks with independent parameters. The skip connections from $\bm{P}_t^l$ supply position-conditioned multi-scale context, while $\bm{C}^l$ provides a lightweight communication channel through which the two branches share local structural information without full feature fusion.

The final predictions are obtained from the highest-resolution decoder features:
\begin{equation}
\widehat{\bm{I}}_t = H_I(\bm{D}_I^1), \qquad
\widehat{\bm{P}}_t = \mathrm{Softmax}\!\left(H_S(\bm{D}_S^1)\right),
\label{eq:output_heads}
\end{equation}
where $H_I$ and $H_S$ are task-specific output heads.

\subsubsection{Spatial Affinity Consistency Constraint}

Rather than forcing image gradients to align with segmentation boundaries---which is unreliable in MRI where real anatomical boundaries may have weak intensity contrast and many irrelevant edges exist---we impose a softer constraint: the two branches should agree on local spatial relationships.

From the high-resolution interpolation decoder features $\bm{D}_I^1$, we compute a normalized structural embedding and define a local affinity:
\begin{equation}
\bm{Z}_I(x) = \frac{\phi_A(\bm{D}_I^1(x))}{\|\phi_A(\bm{D}_I^1(x))\|_2 + \epsilon},
\label{eq:interp_embed}
\end{equation}
\begin{equation}
A_I(x, q) = \exp\!\left(-\frac{\|\bm{Z}_I(x) - \bm{Z}_I(q)\|_2^2}{\tau_a}\right),
\label{eq:interp_affinity}
\end{equation}
where $\phi_A$ is a linear projection and $\tau_a$ is a temperature. This affinity approaches 1 when two positions share similar interpolation structure.

From the segmentation output, we define a complementary affinity based on class probability agreement:
\begin{equation}
A_S(x, q) = \sum_c \widehat{P}_{t,c}(x)\,\widehat{P}_{t,c}(q),
\label{eq:seg_affinity}
\end{equation}
which is high when the two positions are predicted to belong to the same class and low at class boundaries.

The spatial consistency loss aligns these two views of local structure over a neighborhood pixel-pair set $\mathcal{E} = \{(x, q) \mid q \in \mathcal{N}(x)\}$:
\begin{equation}
\mathcal{L}_{\mathrm{sc}} = \frac{1}{|\mathcal{E}|} \sum_{(x,q) \in \mathcal{E}} \mathrm{SmoothL1}\!\left(A_I(x,q) - A_S(x,q)\right).
\label{eq:sc_loss}
\end{equation}
This constraint requires only that the two branches form consistent judgments about which neighboring pixels belong together, without forcing their features to be identical. To prevent background-dominated pixel pairs from overwhelming the loss, we apply balanced sampling across same-class pairs, cross-class pairs, and boundary-adjacent pairs during training.

\subsection{Training Objective}
\label{subsec:loss}

The total loss consists of three components corresponding to interpolation reconstruction, segmentation supervision, and cross-task spatial consistency:
\begin{equation}
\mathcal{L}_{\mathrm{total}} = \mathcal{L}_{\mathrm{int}} + \lambda_{\mathrm{seg}}\,\mathcal{L}_{\mathrm{seg}} + \lambda_{\mathrm{sc}}\,\mathcal{L}_{\mathrm{sc}}.
\label{eq:total_loss}
\end{equation}

\subsubsection{Interpolation Reconstruction Loss}

The interpolation loss combines a robust pixel-level term with a structural similarity term:
\begin{equation}
\mathcal{L}_{\mathrm{int}} = \mathcal{L}_{\mathrm{char}} + \alpha_{\mathrm{ssim}}\,\mathcal{L}_{\mathrm{ssim}}.
\label{eq:interp_loss}
\end{equation}
The Charbonnier loss provides outlier-robust pixel supervision:
\begin{equation}
\mathcal{L}_{\mathrm{char}} = \frac{1}{|\Omega|} \sum_{x \in \Omega} \sqrt{\left(\widehat{I}_t(x) - I_t(x)\right)^2 + \epsilon^2},
\label{eq:charbonnier}
\end{equation}
while the SSIM loss $\mathcal{L}_{\mathrm{ssim}} = 1 - \mathrm{SSIM}(\widehat{\bm{I}}_t, \bm{I}_t)$ preserves local contrast, luminance, and tissue texture continuity. We do not employ adversarial or perceptual losses to avoid hallucinating anatomical structures that do not exist in the ground truth.

\subsubsection{Segmentation Loss}

The segmentation loss combines Dice and cross-entropy terms to address class imbalance:
\begin{equation}
\mathcal{L}_{\mathrm{seg}} = \mathcal{L}_{\mathrm{dice}} + \alpha_{\mathrm{ce}}\,\mathcal{L}_{\mathrm{ce}},
\label{eq:seg_loss}
\end{equation}
where the Dice loss
\begin{equation}
\mathcal{L}_{\mathrm{dice}} = 1 - \frac{2\sum_{x,c} \widehat{P}_{t,c}(x)\,Y_{t,c}(x) + \epsilon}{\sum_{x,c} \widehat{P}_{t,c}(x) + \sum_{x,c} Y_{t,c}(x) + \epsilon}
\label{eq:dice}
\end{equation}
provides region-level overlap supervision robust to foreground-background imbalance, and the cross-entropy loss $\mathcal{L}_{\mathrm{ce}} = -\frac{1}{|\Omega|}\sum_x \sum_c Y_{t,c}(x)\log\widehat{P}_{t,c}(x)$ supplies stable per-pixel gradient signals.

\subsubsection{Loss Weights and Warm-Up Schedule}

The spatial consistency weight follows a linear warm-up schedule:
\begin{equation}
\lambda_{\mathrm{sc}}(e) = \lambda_{\mathrm{sc}}^{\max}\,\min\!\left(1,\; \frac{e}{E_w}\right),
\label{eq:warmup}
\end{equation}
where $e$ is the current epoch and $E_w$ is the warm-up period. This prevents the two branches from imposing erroneous mutual constraints during early training when neither has converged. We set $\alpha_{\mathrm{ssim}}{=}0.15$, $\alpha_{\mathrm{ce}}{=}1$, $\lambda_{\mathrm{seg}}{=}1$, and $\lambda_{\mathrm{sc}}^{\max}{=}0.1$ throughout all experiments.

The cross-task prototypes in CTPI are trained end-to-end through the interpolation and segmentation losses without a separate contrastive objective. The category-level alignment emerges naturally from the shared class probability weights in Eqs.~\eqref{eq:seg_proto}--\eqref{eq:interp_proto} and the downstream task supervision.

\subsection{Training Strategy}
\label{subsec:training}

\subsubsection{Training Sample Construction}

From a relatively thin-slice or high-resolution reference volume, we select three slices $I_i$, $I_k$, $I_j$ with $i < k < j$ and define $\bm{I}_0 = I_i$, $\bm{I}_1 = I_j$, $\bm{I}_t = I_k$ with $t = (z_k - z_i)/(z_j - z_i)$. The corresponding segmentation label is $\bm{Y}_t = Y_k$. To simulate realistic thick-slice acquisition rather than simple slice deletion, we apply slice-profile blurring along the z-axis before sparse downsampling.

\subsubsection{Multi-Position and Multi-Interval Training}

We randomly sample varying endpoint intervals (e.g., $j{-}i \in \{2, 3, 4, 5, 6\}$) during training, exposing the network to diverse non-integer relative positions rather than only midpoint interpolation. Input order is randomly swapped as $(\bm{I}_0, \bm{I}_1, t) \leftrightarrow (\bm{I}_1, \bm{I}_0, 1{-}t)$ to reinforce the algebraic symmetry built into PPA-IFG and reduce order dependence without requiring an explicit consistency loss.

\subsubsection{Two-Stage Optimization}

Training proceeds in two phases. In the first phase (approximately 15\% of total epochs), CTPI influence is attenuated and the network optimizes $\mathcal{L}_{\mathrm{int}} + \lambda_{\mathrm{seg}}\mathcal{L}_{\mathrm{seg}}$ to establish stable per-task representations. In the second phase, the full CTPI module and spatial consistency loss are activated, and the prototype construction transitions from ground-truth-guided to prediction-guided via the annealing schedule described in Sec.~\ref{subsec:ctpi}.

\section{Experiments}
\label{sec:experiments}

In this section, we first present the experimental setup including implementation details, datasets, preprocessing and evaluation metrics. We then report quantitative comparative results, provide qualitative visualizations, and evaluate performance under varying sampling spacings and interpolation positions.

\subsection{Experimental Setup}
\subsubsection{Implementation Details}
All experiments were conducted on a Linux workstation equipped with an NVIDIA RTX 4090 GPU and 24 GB system memory. The proposed CoInS-Net was implemented in PyTorch and trained with the AdamW optimizer. Mixed-precision training was employed to improve memory efficiency and training stability. The initial learning rate was set to $1\times10^{-4}$ with a cosine annealing scheduler. Training was conducted for up to 100 epochs with early stopping based on validation performance. The model checkpoint achieving the highest combined score of interpolation SSIM and segmentation Dice was retained for final evaluation.

\subsubsection{Datasets and Preprocessing}
We evaluated CoInS-Net on four public medical image datasets with different modalities, anatomical regions and task complexities to verify the generalization robustness of our multi-task framework. These include the Task02\_Heart dataset from Medical Segmentation Decathlon with 21 3D NIfTI volumes for single-class left atrial segmentation on cardiac MRI\footnote{\url{http://medicaldecathlon.com/dataaws/}}, the Task07\_Pancreas dataset from the same benchmark with 282 abdominal CT volumes for dual-class pancreas and tumor segmentation, the AMOS2022 dataset with cross-modal CT and MRI scans and fine-grained annotations of 15 abdominal structures, where the model receives only one modality per case and is required to adapt to both input modalities\footnote{\url{https://amos22.grand-challenge.org/Instructions/}}, and the BraTS 2024 post-treatment glioma dataset with multi-region pathological MRI annotations, from which the t1n modality is adopted for experiments\footnote{\url{https://www.kaggle.com/datasets/awsaf49/brats20-dataset-training-validation}}. All datasets were split at the patient level into training, validation and test with a 7:1:2 ratio.

A consistent data preprocessing pipeline was adopted for all datasets to eliminate potential biases. First, NIfTI files are read via the nibabel library and converted into PyTorch tensors with batch-channel-height-width dimension order. Second, per-image normalization is applied to scale pixel values into the range 0 to 1, eliminating the impact of differences in scanning devices and imaging parameters. Third, data augmentation including random horizontal and vertical flipping, random rotation, and random scaling is applied to the training set to mitigate overfitting and enhance model robustness. For the validation and test sets, only format conversion and normalization are performed.

\subsubsection{Evaluation Metrics}
Segmentation and interpolation performance are quantitatively assessed using standard metrics. Interpolation quality is measured by Peak Signal-to-Noise Ratio and Structural Similarity Index, while segmentation accuracy is evaluated by Dice Coefficient and Recall.

Peak Signal-to-Noise Ratio is calculated as:
\begin{equation}
\begin{aligned}
\text{MSE} &= \frac{1}{H \times W} \sum_{h=1}^{H} \sum_{w=1}^{W} (\hat{X}_2(h,w) - X_2(h,w))^2 \\
\text{PSNR} &= 10 \times \log_{10}\left(\frac{\text{max}(X_2)^2}{\text{MSE}}\right)
\end{aligned}
\end{equation}
Here $H$ and $W$ denote frame height and width, $X_2$ is the ground truth frame, $\hat{X}_2$ is the predicted frame, and $\text{max}(X_2)$ is the maximum pixel value of the ground truth.

Structural Similarity Index is defined as:
\begin{equation}
\begin{aligned}
A_{\text{SSIM}} &= (2\mu_{X_2}\mu_{\hat{X}_2}+C_1)(2\sigma_{X_2\hat{X}_2}+C_2) \\
B_{\text{SSIM}} &= (\mu_{X_2}^2+\mu_{\hat{X}_2}^2+C_1)(\sigma_{X_2}^2+\sigma_{\hat{X}_2}^2+C_2) \\
\text{SSIM}(X_2, \hat{X}_2) &= \frac{A_{\text{SSIM}}}{B_{\text{SSIM}}}
\end{aligned}
\end{equation}
Here $\mu$ denotes pixel mean, $\sigma^2$ denotes pixel variance, $\sigma_{X_2\hat{X}_2}$ denotes cross-covariance, and $C_1,C_2$ are regularization constants.

Dice Coefficient is computed as:
\begin{equation}
\begin{aligned}
I_i &= \sum_{h=1}^{H} \sum_{w=1}^{W} (\hat{Y}_i(h,w) \times Y_i(h,w)) \\
P_i &= \sum_{h=1}^{H} \sum_{w=1}^{W} \hat{Y}_i(h,w)^2, \quad G_i = \sum_{h=1}^{H} \sum_{w=1}^{W} Y_i(h,w)^2 \\
\text{Dice}(\hat{Y}_i, Y_i) &= \frac{2I_i}{P_i + G_i}
\end{aligned}
\end{equation}
Here $\hat{Y}_i$ and $Y_i$ are predicted and ground truth segmentation masks respectively.

Recall is defined as:
\begin{equation}
\text{Recall} = \frac{I_i}{G_i}
\end{equation}
Notations are consistent with the Dice coefficient definition.

Statistical significance was assessed using Welch's t-test across five independent experiments. Single, double and triple asterisks represent $p<0.05$, $p<0.01$, and $p<0.001$ respectively compared with our proposed method. The best performance in each metric column is marked in bold.

\subsection{Quantitative Performance Comparison}
We evaluate CoInS-Net on four medical image datasets against representative single-task state-of-the-art methods. Interpolation baselines cover residual channel attention architectures, spatial-aware slice synthesis, multi-contrast arbitrary-scale upsampling and implicit neural representation designs, including McASSR~\cite{McASSR}, SAINT~\cite{SAINT}, RCAN~\cite{RCAN}, ACVTT~\cite{uhm2026anisotropic}, $\text{I}^3$Net~\cite{Song2024I3Net}, CycleINR~\cite{CycleINR} and SFCLI-Net~\cite{li2025sfclinet}. Segmentation baselines include convolutional U-shaped networks, transformer backbones, Mamba-based models, diffusion methods and test-time adaptation schemes, namely nnU-Net~\cite{isensee2021nature}, UNet++~\cite{zhou2018tmi}, Swin UNETR~\cite{cao2022mia}, SegMamba-V2~\cite{xing2025segmambav2}, HiDiff~\cite{chen2024hidiff}, Anatomy-Aware Seg~\cite{wang2024accurate} and SicTTA~\cite{wu2026sictta}.

As shown in Tables \ref{tab:interp_all} and \ref{tab:seg_all}, CoInS-Net consistently outperforms all single-task state-of-the-art baselines on both tasks across the four datasets. For interpolation, it yields 1.12 to 1.67 dB PSNR gains over the best baseline SFCLI-Net with universally higher SSIM, verifying that anatomical constraints effectively preserve tissue structure integrity and boundary sharpness. For segmentation, it improves Dice by 1.8 to 2.9 percentage points compared with SegMamba-V2, and the synchronous rise in Recall confirms that inter-slice deformation cues reduce lesion omission. The performance margin widens as dataset difficulty increases, demonstrating the strong generalization of the proposed multi-task interaction framework.

\begin{table*}[!tbp]
\centering
\fontsize{7}{9}\selectfont
\setlength{\tabcolsep}{3pt}
\caption{Interpolation Performance on Four Medical Image Datasets}
\label{tab:interp_all}
\begin{tabular}{@{}lcccccccc@{}}
\toprule
\multirow{2}{*}{Model} & \multicolumn{2}{c}{MSD Heart} & \multicolumn{2}{c}{MSD Pancreas} & \multicolumn{2}{c}{AMOS2022} & \multicolumn{2}{c}{BraTS 2024} \\
\cmidrule(lr){2-3} \cmidrule(lr){4-5} \cmidrule(lr){6-7} \cmidrule(lr){8-9}
& PSNR (dB) & SSIM & PSNR (dB) & SSIM & PSNR (dB) & SSIM & PSNR (dB) & SSIM \\
\midrule
RCAN & 35.73$\pm$0.31$^{***}$ & 0.950$\pm$0.0041$^{***}$ & 34.78$\pm$0.29$^{***}$ & 0.933$\pm$0.0039$^{***}$ & 35.15$\pm$0.34$^{***}$ & 0.930$\pm$0.0046$^{***}$ & 34.02$\pm$0.36$^{***}$ & 0.918$\pm$0.0049$^{***}$ \\
SAINT & 35.99$\pm$0.26$^{***}$ & 0.953$\pm$0.0034$^{***}$ & 35.04$\pm$0.24$^{***}$ & 0.936$\pm$0.0032$^{***}$ & 35.42$\pm$0.29$^{***}$ & 0.935$\pm$0.0039$^{***}$ & 34.31$\pm$0.31$^{***}$ & 0.922$\pm$0.0042$^{***}$ \\
McASSR & 36.41$\pm$0.19$^{***}$ & 0.958$\pm$0.0026$^{***}$ & 35.47$\pm$0.18$^{***}$ & 0.942$\pm$0.0024$^{***}$ & 35.80$\pm$0.22$^{***}$ & 0.941$\pm$0.0030$^{***}$ & 34.76$\pm$0.24$^{***}$ & 0.929$\pm$0.0033$^{***}$ \\
ACVTT & 35.44$\pm$0.35$^{***}$ & 0.946$\pm$0.0047$^{***}$ & 34.57$\pm$0.33$^{***}$ & 0.930$\pm$0.0044$^{***}$ & 34.82$\pm$0.38$^{***}$ & 0.925$\pm$0.0051$^{***}$ & 33.76$\pm$0.40$^{***}$ & 0.913$\pm$0.0054$^{***}$ \\
I$^3$Net & 36.11$\pm$0.28$^{***}$ & 0.954$\pm$0.0037$^{***}$ & 35.18$\pm$0.27$^{***}$ & 0.937$\pm$0.0036$^{***}$ & 35.37$\pm$0.31$^{***}$ & 0.933$\pm$0.0042$^{***}$ & 34.45$\pm$0.33$^{***}$ & 0.924$\pm$0.0045$^{***}$ \\
CycleINR & 36.29$\pm$0.23$^{***}$ & 0.957$\pm$0.0031$^{***}$ & 35.43$\pm$0.21$^{***}$ & 0.941$\pm$0.0028$^{***}$ & 35.69$\pm$0.25$^{***}$ & 0.938$\pm$0.0034$^{***}$ & 34.68$\pm$0.27$^{***}$ & 0.927$\pm$0.0037$^{***}$ \\
SFCLI-Net & 36.64$\pm$0.16$^{***}$ & 0.962$\pm$0.0021$^{***}$ & 35.70$\pm$0.15$^{***}$ & 0.945$\pm$0.0020$^{***}$ & 36.03$\pm$0.18$^{***}$ & 0.944$\pm$0.0024$^{***}$ & 35.05$\pm$0.20$^{***}$ & 0.933$\pm$0.0026$^{***}$ \\
Ours & \textbf{37.88$\pm$0.21} & \textbf{0.975$\pm$0.0013} & \textbf{37.01$\pm$0.08} & \textbf{0.958$\pm$0.0005} & \textbf{37.15$\pm$0.14} & \textbf{0.958$\pm$0.0021} & \textbf{36.72$\pm$0.16} & \textbf{0.949$\pm$0.0023} \\
\bottomrule
\end{tabular}
\end{table*}

\begin{table*}[!tbp]
\centering
\fontsize{7}{9}\selectfont
\setlength{\tabcolsep}{3pt}
\caption{Segmentation Performance on Four Medical Image Datasets}
\label{tab:seg_all}
\begin{tabular}{@{}lcccccccc@{}}
\toprule
\multirow{2}{*}{Model} & \multicolumn{2}{c}{MSD Heart} & \multicolumn{2}{c}{MSD Pancreas} & \multicolumn{2}{c}{AMOS2022} & \multicolumn{2}{c}{BraTS 2024} \\
\cmidrule(lr){2-3} \cmidrule(lr){4-5} \cmidrule(lr){6-7} \cmidrule(lr){8-9}
& Dice & Recall & Dice & Recall & Dice & Recall & Dice & Recall \\
\midrule
UNet++ & 0.895$\pm$0.018$^{***}$ & 0.897$\pm$0.020$^{***}$ & 0.905$\pm$0.017$^{***}$ & 0.907$\pm$0.019$^{***}$ & 0.784$\pm$0.020$^{***}$ & 0.792$\pm$0.022$^{***}$ & 0.759$\pm$0.022$^{***}$ & 0.770$\pm$0.024$^{***}$ \\
nnU-Net & 0.911$\pm$0.014$^{***}$ & 0.913$\pm$0.016$^{**}$ & 0.922$\pm$0.013$^{***}$ & 0.923$\pm$0.015$^{***}$ & 0.801$\pm$0.016$^{***}$ & 0.808$\pm$0.018$^{***}$ & 0.778$\pm$0.018$^{***}$ & 0.789$\pm$0.020$^{***}$ \\
Swin UNETR & 0.928$\pm$0.011$^{***}$ & 0.930$\pm$0.013$^{***}$ & 0.939$\pm$0.010$^{***}$ & 0.940$\pm$0.011$^{***}$ & 0.818$\pm$0.013$^{***}$ & 0.825$\pm$0.015$^{***}$ & 0.797$\pm$0.015$^{***}$ & 0.808$\pm$0.017$^{***}$ \\
SegMamba-V2 & 0.947$\pm$0.008$^{***}$ & 0.949$\pm$0.009$^{**}$ & 0.956$\pm$0.007$^{**}$ & 0.957$\pm$0.008$^{***}$ & 0.838$\pm$0.010$^{***}$ & 0.847$\pm$0.011$^{**}$ & 0.819$\pm$0.012$^{*}$ & 0.831$\pm$0.013$^{***}$ \\
HiDiff & 0.932$\pm$0.013$^{**}$ & 0.936$\pm$0.015$^{***}$ & 0.941$\pm$0.012$^{***}$ & 0.944$\pm$0.013$^{***}$ & 0.821$\pm$0.014$^{***}$ & 0.830$\pm$0.016$^{**}$ & 0.800$\pm$0.016$^{*}$ & 0.812$\pm$0.018$^{***}$ \\
Anatomy-Aware & 0.918$\pm$0.016$^{***}$ & 0.920$\pm$0.018$^{***}$ & 0.927$\pm$0.015$^{***}$ & 0.928$\pm$0.017$^{***}$ & 0.806$\pm$0.018$^{***}$ & 0.815$\pm$0.020$^{**}$ & 0.784$\pm$0.020$^{**}$ & 0.795$\pm$0.022$^{***}$ \\
SicTTA & 0.939$\pm$0.010$^{***}$ & 0.941$\pm$0.011$^{***}$ & 0.948$\pm$0.009$^{**}$ & 0.949$\pm$0.010$^{***}$ & 0.829$\pm$0.011$^{***}$ & 0.836$\pm$0.013$^{***}$ & 0.809$\pm$0.013 & 0.819$\pm$0.015$^{***}$ \\
Ours & \textbf{0.968$\pm$0.003} & \textbf{0.970$\pm$0.004} & \textbf{0.974$\pm$0.006} & \textbf{0.975$\pm$0.006} & \textbf{0.867$\pm$0.009} & \textbf{0.875$\pm$0.010} & \textbf{0.842$\pm$0.011} & \textbf{0.852$\pm$0.012} \\
\bottomrule
\end{tabular}
\end{table*}

To further investigate the fine-grained segmentation performance of the proposed method on abdominal anatomical structures, we report per-organ segmentation results on the AMOS2022 dataset. We select the five organs with the largest volume fraction, namely the liver, stomach, spleen, left kidney, and right kidney, for detailed comparison. As shown in Table \ref{tab:amos_per_organ}, CoInS-Net achieves consistent improvements across all five anatomical structures. The liver with the largest volume and relatively uniform intensity obtains the highest segmentation accuracy and the lowest prediction variance, while the stomach with large morphological variability and irregular cavity contours achieves the lowest performance among the selected organs. The spleen and bilateral kidneys with moderate volume obtain intermediate segmentation accuracy. Stable performance gains across organs of different sizes and morphological complexities demonstrate that the continuous position-aware modeling and prototype interaction mechanism can be generalized to diverse abdominal anatomical structures.

\begin{table*}[!tbp]
\centering
\fontsize{7}{9}\selectfont
\setlength{\tabcolsep}{1pt}
\caption{Per-organ Segmentation Performance on AMOS2022 Dataset}
\label{tab:amos_per_organ}
\begin{tabular}{@{}lcccccccccc@{}}
\toprule
\multirow{2}{*}{Model} & \multicolumn{2}{c}{Liver} & \multicolumn{2}{c}{Stomach} & \multicolumn{2}{c}{Spleen} & \multicolumn{2}{c}{Left Kidney} & \multicolumn{2}{c}{Right Kidney} \\
\cmidrule(lr){2-3} \cmidrule(lr){4-5} \cmidrule(lr){6-7} \cmidrule(lr){8-9} \cmidrule(lr){10-11}
& Dice & Recall & Dice & Recall & Dice & Recall & Dice & Recall & Dice & Recall \\
\midrule
UNet++ & 0.849$\pm$0.023$^{***}$ & 0.858$\pm$0.026$^{***}$ & 0.710$\pm$0.028$^{***}$ & 0.721$\pm$0.031$^{***}$ & 0.807$\pm$0.021$^{***}$ & 0.816$\pm$0.024$^{***}$ & 0.782$\pm$0.025$^{***}$ & 0.791$\pm$0.028$^{***}$ & 0.795$\pm$0.019$^{***}$ & 0.804$\pm$0.022$^{***}$ \\
nnU-Net & 0.868$\pm$0.017$^{***}$ & 0.876$\pm$0.020$^{***}$ & 0.729$\pm$0.022$^{***}$ & 0.739$\pm$0.025$^{***}$ & 0.825$\pm$0.019$^{***}$ & 0.833$\pm$0.022$^{***}$ & 0.801$\pm$0.021$^{***}$ & 0.809$\pm$0.024$^{***}$ & 0.813$\pm$0.016$^{***}$ & 0.822$\pm$0.018$^{***}$ \\
Anatomy & 0.873$\pm$0.020$^{***}$ & 0.881$\pm$0.023$^{***}$ & 0.736$\pm$0.025$^{***}$ & 0.746$\pm$0.028$^{***}$ & 0.830$\pm$0.023$^{***}$ & 0.839$\pm$0.026$^{***}$ & 0.806$\pm$0.018$^{***}$ & 0.815$\pm$0.021$^{***}$ & 0.816$\pm$0.020$^{***}$ & 0.825$\pm$0.023$^{***}$ \\
UNETR & 0.889$\pm$0.013$^{***}$ & 0.897$\pm$0.015$^{***}$ & 0.751$\pm$0.019$^{***}$ & 0.760$\pm$0.022$^{***}$ & 0.845$\pm$0.017$^{***}$ & 0.853$\pm$0.020$^{***}$ & 0.822$\pm$0.016$^{***}$ & 0.830$\pm$0.019$^{***}$ & 0.830$\pm$0.012$^{***}$ & 0.838$\pm$0.015$^{***}$ \\
HiDiff & 0.893$\pm$0.015$^{***}$ & 0.901$\pm$0.018$^{***}$ & 0.755$\pm$0.021$^{***}$ & 0.765$\pm$0.024$^{**}$ & 0.849$\pm$0.014$^{***}$ & 0.858$\pm$0.017$^{***}$ & 0.825$\pm$0.018$^{***}$ & 0.833$\pm$0.021$^{***}$ & 0.832$\pm$0.014$^{***}$ & 0.841$\pm$0.016$^{***}$ \\
SicTTA & 0.902$\pm$0.010$^{***}$ & 0.910$\pm$0.012$^{***}$ & 0.768$\pm$0.015$^{***}$ & 0.777$\pm$0.018$^{**}$ & 0.859$\pm$0.016$^{**}$ & 0.867$\pm$0.019$^{**}$ & 0.837$\pm$0.011$^{***}$ & 0.845$\pm$0.014$^{***}$ & 0.842$\pm$0.009$^{***}$ & 0.851$\pm$0.012$^{***}$ \\
SegMamba & 0.914$\pm$0.008$^{***}$ & 0.921$\pm$0.010$^{**}$ & 0.782$\pm$0.014$^{**}$ & 0.791$\pm$0.016$^{**}$ & 0.869$\pm$0.009$^{***}$ & 0.877$\pm$0.012$^{**}$ & 0.847$\pm$0.012$^{**}$ & 0.855$\pm$0.014$^{**}$ & 0.854$\pm$0.008$^{***}$ & 0.862$\pm$0.011$^{***}$ \\
Ours & \textbf{0.937$\pm$0.005} & \textbf{0.944$\pm$0.007} & \textbf{0.808$\pm$0.009} & \textbf{0.818$\pm$0.011} & \textbf{0.892$\pm$0.006} & \textbf{0.900$\pm$0.008} & \textbf{0.870$\pm$0.007} & \textbf{0.879$\pm$0.009} & \textbf{0.886$\pm$0.006} & \textbf{0.895$\pm$0.008} \\
\bottomrule
\end{tabular}
\end{table*}

\subsection{Qualitative Visualization}
\begin{figure*}[!tbp]
\centering
\fontsize{7}{9}\selectfont
\setlength{\tabcolsep}{0.1pt}
\includegraphics[width=1.0\textwidth]{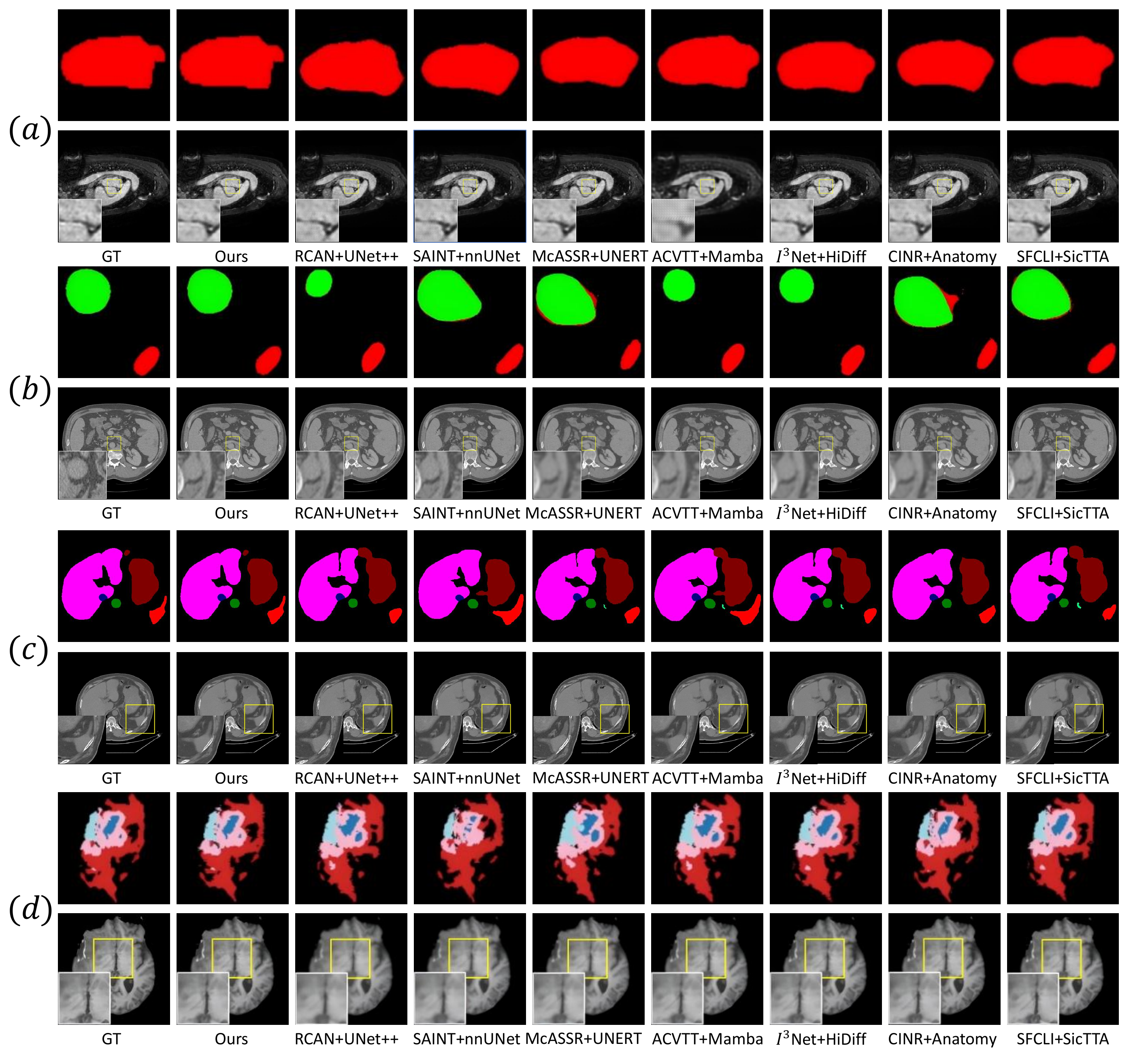}
\caption{Qualitative results on four datasets. (a) MSD Heart; (b) MSD Pancreas; (c) AMOS2022; (d) BraTS2024.}
\label{fig:qualitative_all}
\end{figure*}

Qualitative comparison on the four datasets is shown in Fig. \ref{fig:qualitative_all}. CoInS-Net generates sharper tissue boundaries in interpolated slices and produces more accurate segmentation masks for target regions, while baseline methods tend to produce blurry edges and miss fine structural details.

\subsection{Performance under Different Sampling Spacings}
We evaluate joint interpolation and segmentation performance on MSD Pancreas and BraTS 2024 under eight through-plane spacings to verify generalization across clinical scanning protocols. The 4 mm setting matches the default configuration used in prior comparative experiments. Six cascaded interpolation-segmentation model pairs serve as baselines. SSIM and Dice are adopted as core evaluation metrics.

\begin{figure*}[!tbp]
\centering
\fontsize{7}{9}\selectfont
\setlength{\tabcolsep}{0.1pt}
\includegraphics[width=1.0\textwidth]{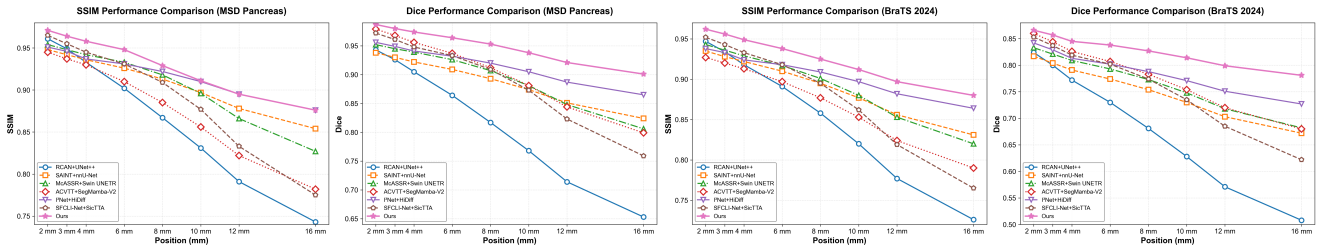}
\caption{Performance comparison across different z-axis sampling spacings on MSD Pancreas and BraTS 2024 datasets.}
\label{fig:diff_pos}
\end{figure*}

Tables \ref{tab:spacing_23}, \ref{tab:spacing_46}, \ref{tab:spacing_810}, and \ref{tab:spacing_1216} and Figure \ref{fig:diff_pos} present results for different spacing ranges. All methods degrade gradually with increasing spacing due to amplified anatomical deformation and inter-slice discontinuity. CoInS-Net consistently outperforms all cascaded baselines on both datasets across all settings, and its performance advantage widens at larger spacings. This demonstrates that the continuous position-aware interaction mechanism effectively alleviates performance drop under sparse sampling. From 2 mm to 16 mm, Dice decreases by 8.7 percentage points on MSD Pancreas and 9.8 percentage points on BraTS 2024 for our method, versus average declines of 15.8 and 18.4 percentage points for baselines. These results verify that continuous position query modeling stabilizes joint task performance under variable acquisition parameters and exhibits superior robustness compared to cascaded approaches.

\begin{table*}[!tbp]
\centering
\fontsize{7}{9}\selectfont
\setlength{\tabcolsep}{3pt}
\caption{Performance comparison under 2 mm and 3 mm z-axis sampling spacings on MSD Pancreas and BraTS 2024 datasets}
\label{tab:spacing_23}
\begin{tabular}{@{}lcccccccc@{}}
\toprule
\multirow{2}{*}{Model} & \multicolumn{4}{c}{2 mm} & \multicolumn{4}{c}{3 mm} \\
\cmidrule(lr){2-5} \cmidrule(lr){6-9}
& \multicolumn{2}{c}{MSD Pancreas} & \multicolumn{2}{c}{BraTS 2024} & \multicolumn{2}{c}{MSD Pancreas} & \multicolumn{2}{c}{BraTS 2024} \\
\cmidrule(lr){2-3} \cmidrule(lr){4-5} \cmidrule(lr){6-7} \cmidrule(lr){8-9}
& SSIM & Dice & SSIM & Dice & SSIM & Dice & SSIM & Dice \\
\midrule
RCAN+UNet++ & 0.961$\pm$0.0028$^{***}$ & 0.943$\pm$0.014$^{***}$ & 0.947$\pm$0.0037$^{***}$ & 0.824$\pm$0.018$^{**}$ & 0.949$\pm$0.0033$^{***}$ & 0.926$\pm$0.016$^{***}$ & 0.934$\pm$0.0042$^{***}$ & 0.800$\pm$0.019$^{***}$ \\
SAINT+nnU-Net & 0.948$\pm$0.0024$^{***}$ & 0.938$\pm$0.011$^{***}$ & 0.935$\pm$0.0031$^{***}$ & 0.817$\pm$0.015$^{***}$ & 0.942$\pm$0.0028$^{***}$ & 0.930$\pm$0.012$^{***}$ & 0.928$\pm$0.0036$^{***}$ & 0.804$\pm$0.016$^{***}$ \\
McASSR+Swin UNETR & 0.954$\pm$0.0018$^{***}$ & 0.952$\pm$0.008$^{***}$ & 0.943$\pm$0.0025$^{***}$ & 0.833$\pm$0.012$^{***}$ & 0.948$\pm$0.0021$^{***}$ & 0.945$\pm$0.009$^{***}$ & 0.936$\pm$0.0029$^{***}$ & 0.821$\pm$0.013$^{**}$ \\
ACVTT+SegMamba-V2 & 0.945$\pm$0.0033$^{***}$ & 0.979$\pm$0.006 & 0.927$\pm$0.0040$^{***}$ & 0.860$\pm$0.010 & 0.937$\pm$0.0038$^{***}$ & 0.968$\pm$0.007$^{*}$ & 0.920$\pm$0.0047$^{***}$ & 0.844$\pm$0.011$^{**}$ \\
I$^3$Net+HiDiff & 0.951$\pm$0.0027$^{***}$ & 0.956$\pm$0.010$^{**}$ & 0.938$\pm$0.0034$^{***}$ & 0.842$\pm$0.014$^{**}$ & 0.946$\pm$0.0031$^{***}$ & 0.949$\pm$0.011$^{**}$ & 0.933$\pm$0.0039$^{***}$ & 0.829$\pm$0.015$^{***}$ \\
SFCLI-Net+SicTTA & 0.965$\pm$0.0015$^{***}$ & 0.972$\pm$0.008$^{**}$ & 0.952$\pm$0.0019$^{***}$ & 0.853$\pm$0.011$^{***}$ & 0.955$\pm$0.0018$^{***}$ & 0.961$\pm$0.009$^{**}$ & 0.943$\pm$0.0022$^{***}$ & 0.837$\pm$0.012$^{*}$ \\
Ours & \textbf{0.971$\pm$0.0003} & \textbf{0.987$\pm$0.004} & \textbf{0.962$\pm$0.0017} & \textbf{0.866$\pm$0.009} & \textbf{0.964$\pm$0.0004} & \textbf{0.980$\pm$0.005} & \textbf{0.956$\pm$0.0020} & \textbf{0.857$\pm$0.010} \\
\bottomrule
\end{tabular}
\end{table*}

\begin{table*}[!tbp]
\centering
\fontsize{7}{9}\selectfont
\setlength{\tabcolsep}{3pt}
\caption{Performance comparison under 4 mm and 6 mm z-axis sampling spacings on MSD Pancreas and BraTS 2024 datasets}
\label{tab:spacing_46}
\begin{tabular}{@{}lcccccccc@{}}
\toprule
\multirow{2}{*}{Model} & \multicolumn{4}{c}{4 mm} & \multicolumn{4}{c}{6 mm} \\
\cmidrule(lr){2-5} \cmidrule(lr){6-9}
& \multicolumn{2}{c}{MSD Pancreas} & \multicolumn{2}{c}{BraTS 2024} & \multicolumn{2}{c}{MSD Pancreas} & \multicolumn{2}{c}{BraTS 2024} \\
\cmidrule(lr){2-3} \cmidrule(lr){4-5} \cmidrule(lr){6-7} \cmidrule(lr){8-9}
& SSIM & Dice & SSIM & Dice & SSIM & Dice & SSIM & Dice \\
\midrule
RCAN+UNet++ & 0.933$\pm$0.0039$^{***}$ & 0.905$\pm$0.017$^{***}$ & 0.918$\pm$0.0049$^{***}$ & 0.772$\pm$0.020$^{***}$ & 0.902$\pm$0.0053$^{***}$ & 0.864$\pm$0.019$^{***}$ & 0.891$\pm$0.0062$^{***}$ & 0.730$\pm$0.023$^{***}$ \\
SAINT+nnU-Net & 0.936$\pm$0.0032$^{***}$ & 0.922$\pm$0.013$^{***}$ & 0.922$\pm$0.0042$^{***}$ & 0.791$\pm$0.017$^{***}$ & 0.926$\pm$0.0038$^{***}$ & 0.909$\pm$0.017$^{***}$ & 0.910$\pm$0.0056$^{***}$ & 0.774$\pm$0.015$^{***}$ \\
McASSR+Swin UNETR & 0.942$\pm$0.0024$^{***}$ & 0.939$\pm$0.010$^{***}$ & 0.929$\pm$0.0033$^{***}$ & 0.809$\pm$0.014$^{***}$ & 0.933$\pm$0.0031$^{***}$ & 0.926$\pm$0.013$^{**}$ & 0.917$\pm$0.0044$^{***}$ & 0.793$\pm$0.017$^{***}$ \\
ACVTT+SegMamba-V2 & 0.930$\pm$0.0044$^{***}$ & 0.956$\pm$0.007$^{**}$ & 0.913$\pm$0.0054$^{***}$ & 0.826$\pm$0.011 & 0.910$\pm$0.0048$^{***}$ & 0.937$\pm$0.011$^{***}$ & 0.897$\pm$0.0067$^{***}$ & 0.807$\pm$0.011$^{*}$ \\
I$^3$Net+HiDiff & 0.937$\pm$0.0036$^{***}$ & 0.941$\pm$0.012$^{**}$ & 0.924$\pm$0.0045$^{***}$ & 0.813$\pm$0.015$^{**}$ & 0.931$\pm$0.0046$^{***}$ & 0.932$\pm$0.018$^{*}$ & 0.918$\pm$0.0051$^{***}$ & 0.802$\pm$0.019$^{**}$ \\
SFCLI-Net+SicTTA & 0.945$\pm$0.0020$^{***}$ & 0.948$\pm$0.009$^{***}$ & 0.933$\pm$0.0026$^{***}$ & 0.819$\pm$0.012$^{*}$ & 0.931$\pm$0.0034$^{***}$ & 0.933$\pm$0.014$^{***}$ & 0.918$\pm$0.0048$^{***}$ & 0.802$\pm$0.017$^{**}$ \\
Ours & \textbf{0.958$\pm$0.0005} & \textbf{0.974$\pm$0.006} & \textbf{0.949$\pm$0.0023} & \textbf{0.845$\pm$0.010} & \textbf{0.948$\pm$0.0011} & \textbf{0.964$\pm$0.008} & \textbf{0.938$\pm$0.0026} & \textbf{0.838$\pm$0.010} \\
\bottomrule
\end{tabular}
\end{table*}

\begin{table*}[!tbp]
\centering
\fontsize{7}{9}\selectfont
\setlength{\tabcolsep}{3pt}
\caption{Performance comparison under 8 mm and 10 mm z-axis sampling spacings on MSD Pancreas and BraTS 2024 datasets}
\label{tab:spacing_810}
\begin{tabular}{@{}lcccccccc@{}}
\toprule
\multirow{2}{*}{Model} & \multicolumn{4}{c}{8 mm} & \multicolumn{4}{c}{10 mm} \\
\cmidrule(lr){2-5} \cmidrule(lr){6-9}
& \multicolumn{2}{c}{MSD Pancreas} & \multicolumn{2}{c}{BraTS 2024} & \multicolumn{2}{c}{MSD Pancreas} & \multicolumn{2}{c}{BraTS 2024} \\
\cmidrule(lr){2-3} \cmidrule(lr){4-5} \cmidrule(lr){6-7} \cmidrule(lr){8-9}
& SSIM & Dice & SSIM & Dice & SSIM & Dice & SSIM & Dice \\
\midrule
RCAN+UNet++ & 0.867$\pm$0.0048$^{***}$ & 0.817$\pm$0.023$^{***}$ & 0.858$\pm$0.0061$^{***}$ & 0.681$\pm$0.021$^{***}$ & 0.831$\pm$0.0061$^{***}$ & 0.768$\pm$0.028$^{***}$ & 0.820$\pm$0.0067$^{***}$ & 0.628$\pm$0.026$^{***}$ \\
SAINT+nnU-Net & 0.913$\pm$0.0065$^{**}$ & 0.893$\pm$0.018$^{***}$ & 0.895$\pm$0.0059$^{***}$ & 0.754$\pm$0.018$^{***}$ & 0.900$\pm$0.0053$^{**}$ & 0.874$\pm$0.021$^{***}$ & 0.877$\pm$0.0063$^{***}$ & 0.730$\pm$0.020$^{***}$ \\
McASSR+Swin UNETR & 0.918$\pm$0.0039$^{*}$ & 0.907$\pm$0.020$^{**}$ & 0.901$\pm$0.0054$^{***}$ & 0.773$\pm$0.016$^{***}$ & 0.896$\pm$0.0048$^{***}$ & 0.881$\pm$0.016$^{***}$ & 0.880$\pm$0.0062$^{***}$ & 0.748$\pm$0.022$^{***}$ \\
ACVTT+SegMamba-V2 & 0.885$\pm$0.0071$^{***}$ & 0.912$\pm$0.012$^{***}$ & 0.877$\pm$0.0067$^{***}$ & 0.783$\pm$0.013$^{**}$ & 0.856$\pm$0.0075$^{***}$ & 0.881$\pm$0.019$^{***}$ & 0.853$\pm$0.0074$^{***}$ & 0.754$\pm$0.015$^{***}$ \\
I$^3$Net+HiDiff & 0.922$\pm$0.0058$^{*}$ & 0.920$\pm$0.027 & 0.909$\pm$0.0061$^{***}$ & 0.788$\pm$0.017$^{***}$ & 0.910$\pm$0.0059 & 0.905$\pm$0.018$^{***}$ & 0.897$\pm$0.0066$^{***}$ & 0.771$\pm$0.021$^{*}$ \\
SFCLI-Net+SicTTA & 0.909$\pm$0.0051$^{***}$ & 0.909$\pm$0.015$^{***}$ & 0.895$\pm$0.0057$^{***}$ & 0.774$\pm$0.019$^{***}$ & 0.877$\pm$0.0064$^{***}$ & 0.873$\pm$0.024$^{***}$ & 0.862$\pm$0.0069$^{***}$ & 0.735$\pm$0.023$^{**}$ \\
Ours & \textbf{0.929$\pm$0.0021} & \textbf{0.953$\pm$0.009} & \textbf{0.925$\pm$0.0031} & \textbf{0.827$\pm$0.011} & \textbf{0.911$\pm$0.0028} & \textbf{0.938$\pm$0.011} & \textbf{0.912$\pm$0.0037} & \textbf{0.814$\pm$0.013} \\
\bottomrule
\end{tabular}
\end{table*}

\begin{table*}[!tbp]
\centering
\fontsize{7}{9}\selectfont
\setlength{\tabcolsep}{3pt}
\caption{Performance comparison under 12 mm and 16 mm z-axis sampling spacings on MSD Pancreas and BraTS 2024 datasets}
\label{tab:spacing_1216}
\begin{tabular}{@{}lcccccccc@{}}
\toprule
\multirow{2}{*}{Model} & \multicolumn{4}{c}{12 mm} & \multicolumn{4}{c}{16 mm} \\
\cmidrule(lr){2-5} \cmidrule(lr){6-9}
& \multicolumn{2}{c}{MSD Pancreas} & \multicolumn{2}{c}{BraTS 2024} & \multicolumn{2}{c}{MSD Pancreas} & \multicolumn{2}{c}{BraTS 2024} \\
\cmidrule(lr){2-3} \cmidrule(lr){4-5} \cmidrule(lr){6-7} \cmidrule(lr){8-9}
& SSIM & Dice & SSIM & Dice & SSIM & Dice & SSIM & Dice \\
\midrule
RCAN+UNet++ & 0.791$\pm$0.0069$^{***}$ & 0.714$\pm$0.031$^{***}$ & 0.777$\pm$0.0078$^{***}$ & 0.571$\pm$0.029$^{***}$ & 0.743$\pm$0.0076$^{***}$ & 0.653$\pm$0.035$^{***}$ & 0.726$\pm$0.0085$^{***}$ & 0.508$\pm$0.032$^{***}$ \\
SAINT+nnU-Net & 0.882$\pm$0.0073$^{*}$ & 0.851$\pm$0.025$^{***}$ & 0.856$\pm$0.0071$^{***}$ & 0.703$\pm$0.024$^{***}$ & 0.859$\pm$0.0079$^{***}$ & 0.824$\pm$0.029$^{***}$ & 0.831$\pm$0.0082$^{***}$ & 0.672$\pm$0.027$^{***}$ \\
McASSR+Swin UNETR & 0.866$\pm$0.0058$^{***}$ & 0.848$\pm$0.022$^{**}$ & 0.853$\pm$0.0069$^{***}$ & 0.718$\pm$0.020$^{***}$ & 0.827$\pm$0.0065$^{***}$ & 0.806$\pm$0.026$^{***}$ & 0.820$\pm$0.0076$^{***}$ & 0.682$\pm$0.025$^{***}$ \\
ACVTT+SegMamba-V2 & 0.822$\pm$0.0083$^{***}$ & 0.844$\pm$0.022$^{**}$ & 0.824$\pm$0.0087$^{***}$ & 0.720$\pm$0.018$^{***}$ & 0.782$\pm$0.0091$^{***}$ & 0.799$\pm$0.026$^{***}$ & 0.790$\pm$0.0095$^{***}$ & 0.680$\pm$0.021$^{***}$ \\
I$^3$Net+HiDiff & 0.895$\pm$0.0068 & 0.887$\pm$0.029$^{*}$ & 0.882$\pm$0.0075$^{***}$ & 0.751$\pm$0.024$^{***}$ & 0.876$\pm$0.0074 & 0.865$\pm$0.033 & 0.864$\pm$0.0083 & 0.727$\pm$0.028$^{*}$ \\
SFCLI-Net+SicTTA & 0.833$\pm$0.0074$^{***}$ & 0.823$\pm$0.028$^{***}$ & 0.819$\pm$0.0081$^{***}$ & 0.685$\pm$0.026$^{***}$ & 0.775$\pm$0.0082$^{***}$ & 0.759$\pm$0.032$^{***}$ & 0.765$\pm$0.0089$^{***}$ & 0.622$\pm$0.029$^{***}$ \\
Ours & \textbf{0.895$\pm$0.0035} & \textbf{0.921$\pm$0.013} & \textbf{0.897$\pm$0.0044} & \textbf{0.799$\pm$0.015} & \textbf{0.876$\pm$0.0042} & \textbf{0.901$\pm$0.015} & \textbf{0.880$\pm$0.0051} & \textbf{0.781$\pm$0.017} \\
\bottomrule
\end{tabular}
\end{table*}

\begin{figure*}[!tbp]
\centering
\fontsize{7}{9}\selectfont
\noindent
\makebox[0.16\textwidth][c]{8 mm-image}\hspace{-4pt}
\makebox[0.16\textwidth][c]{8 mm-label}\hspace{-2pt}
\makebox[0.16\textwidth][c]{6 mm-image}\hspace{-1pt}
\makebox[0.16\textwidth][c]{6 mm-label}\hspace{0pt}
\makebox[0.16\textwidth][c]{4 mm-image}\hspace{0pt}
\makebox[0.16\textwidth][c]{4 mm-label}
\\[-1pt]

\raisebox{0.04\textwidth}{\rotatebox{90}{\textbf{GT}}}\hspace{2pt}%
\subfigure{\includegraphics[width=0.16\textwidth]{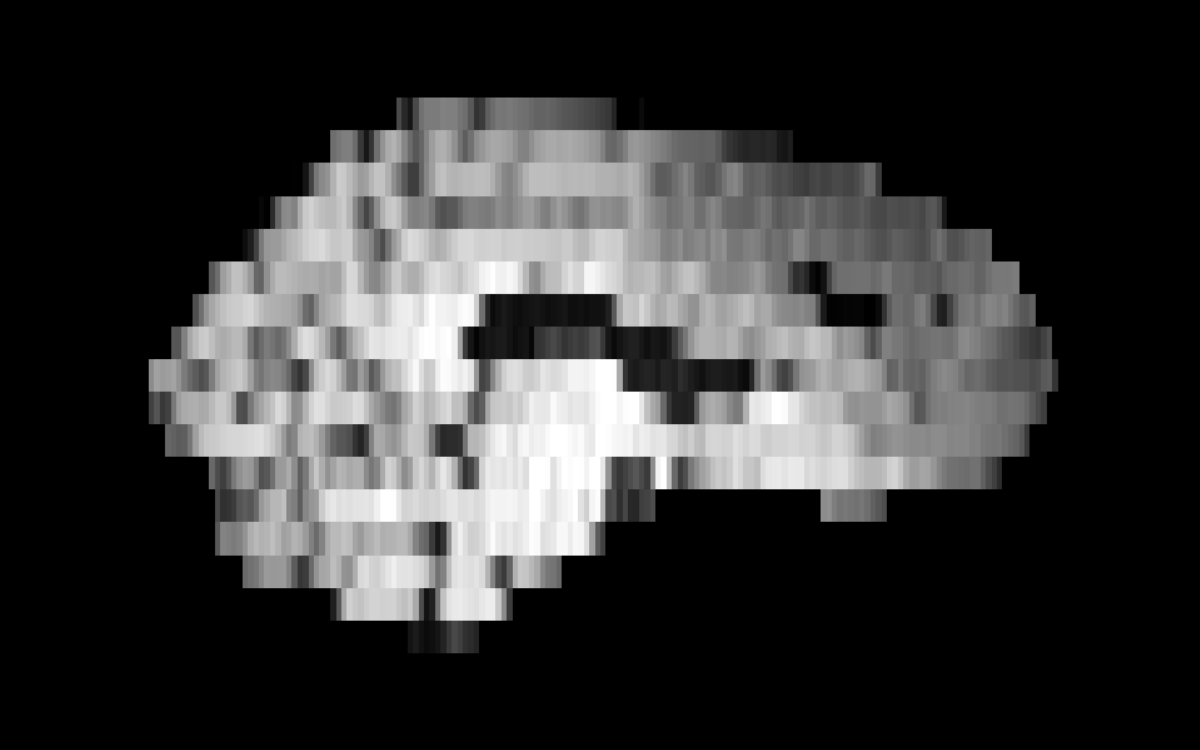}}
\hspace{-4pt}
\subfigure{\includegraphics[width=0.16\textwidth]{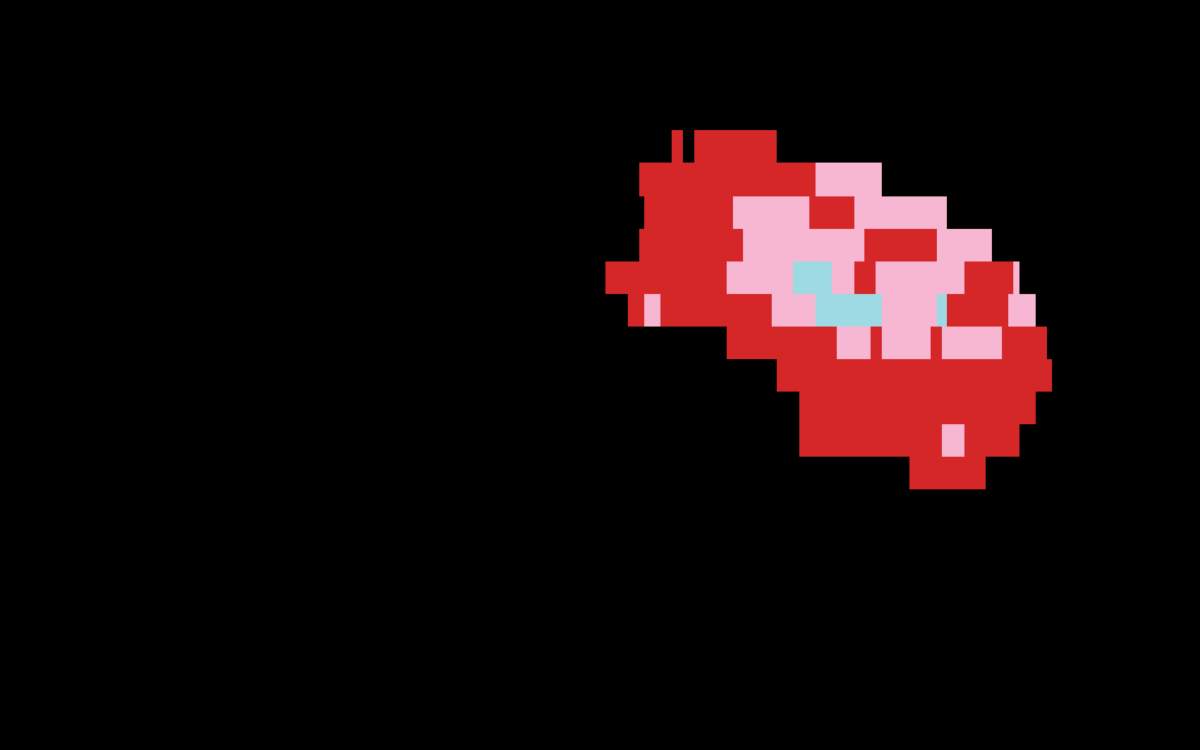}}
\hspace{-4pt}
\subfigure{\includegraphics[width=0.16\textwidth]{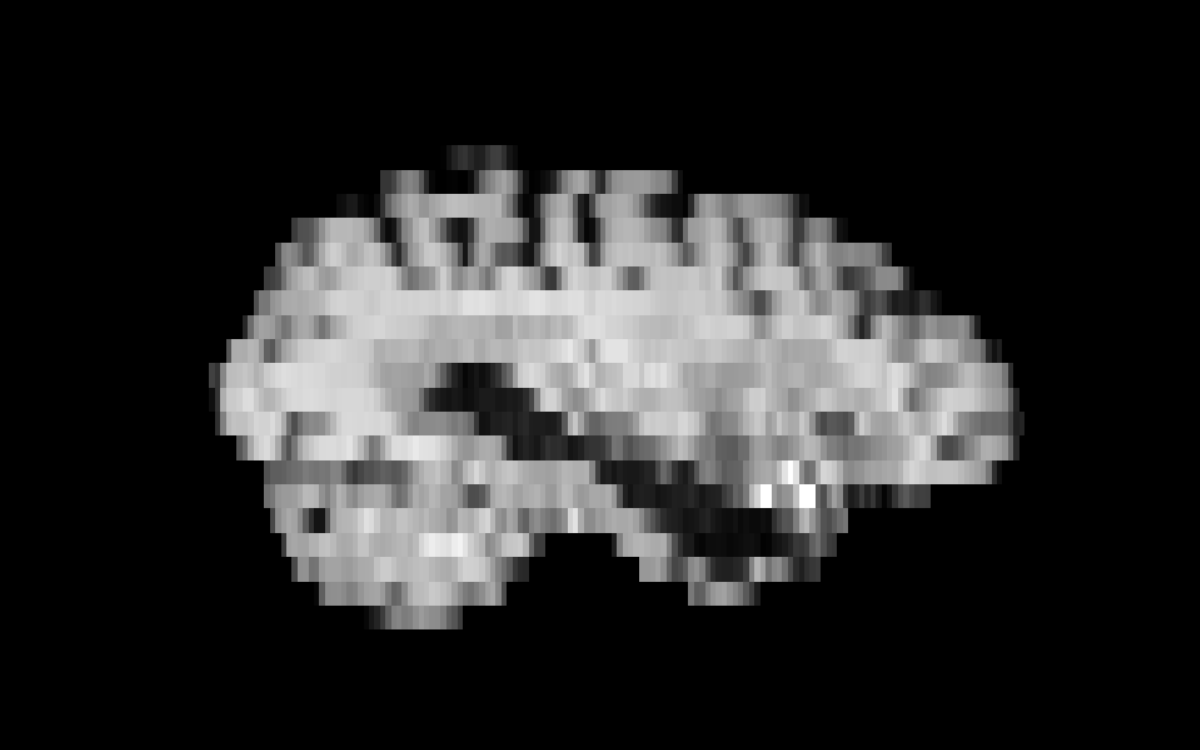}}
\hspace{-4pt}
\subfigure{\includegraphics[width=0.16\textwidth]{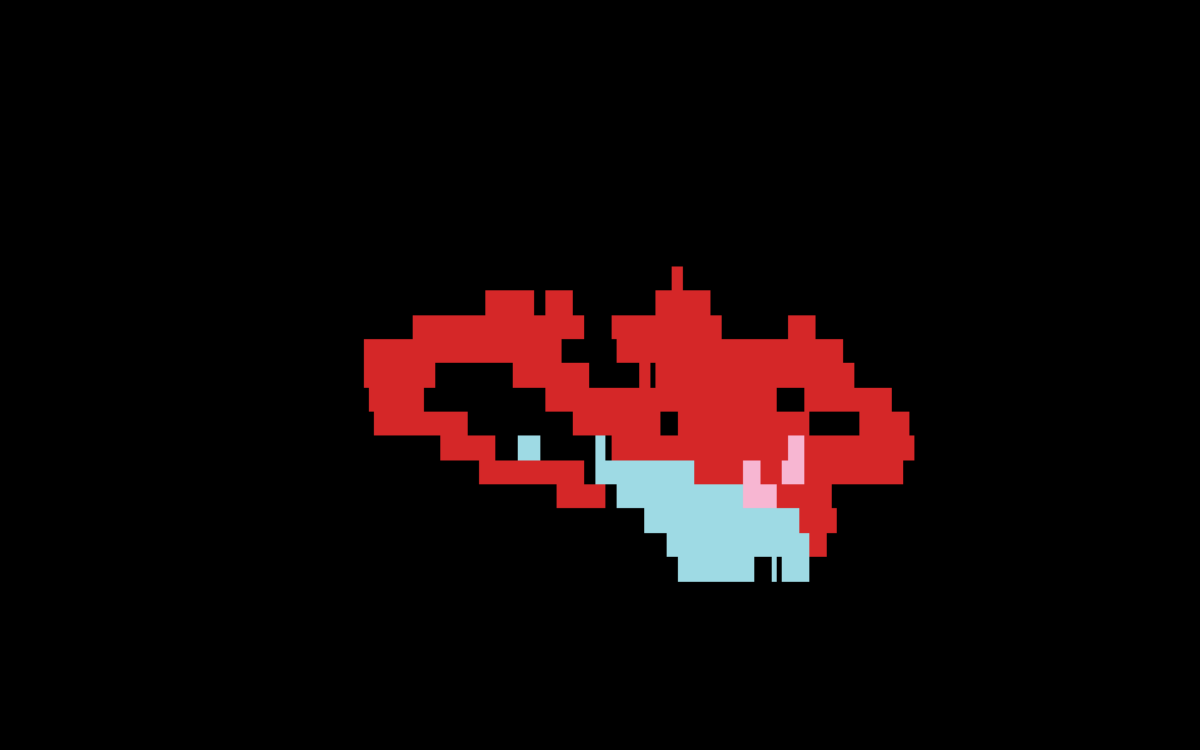}}
\hspace{-4pt}
\subfigure{\includegraphics[width=0.16\textwidth]{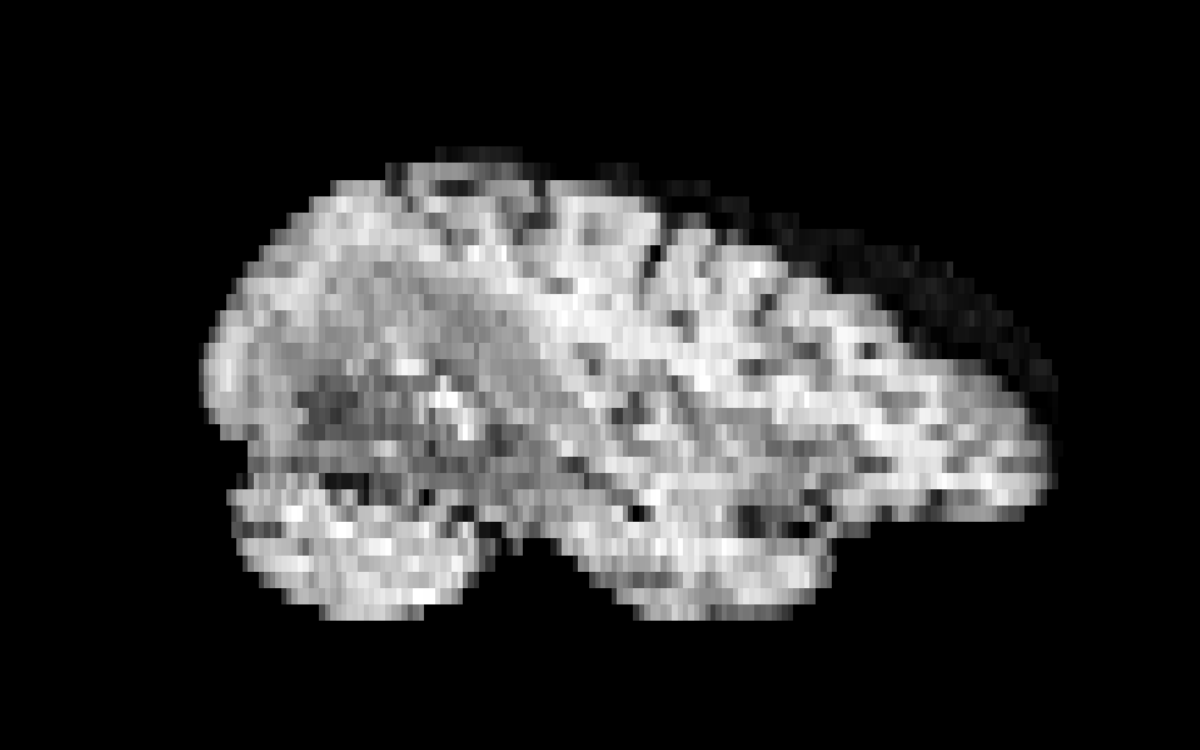}}
\hspace{-4pt}
\subfigure{\includegraphics[width=0.16\textwidth]{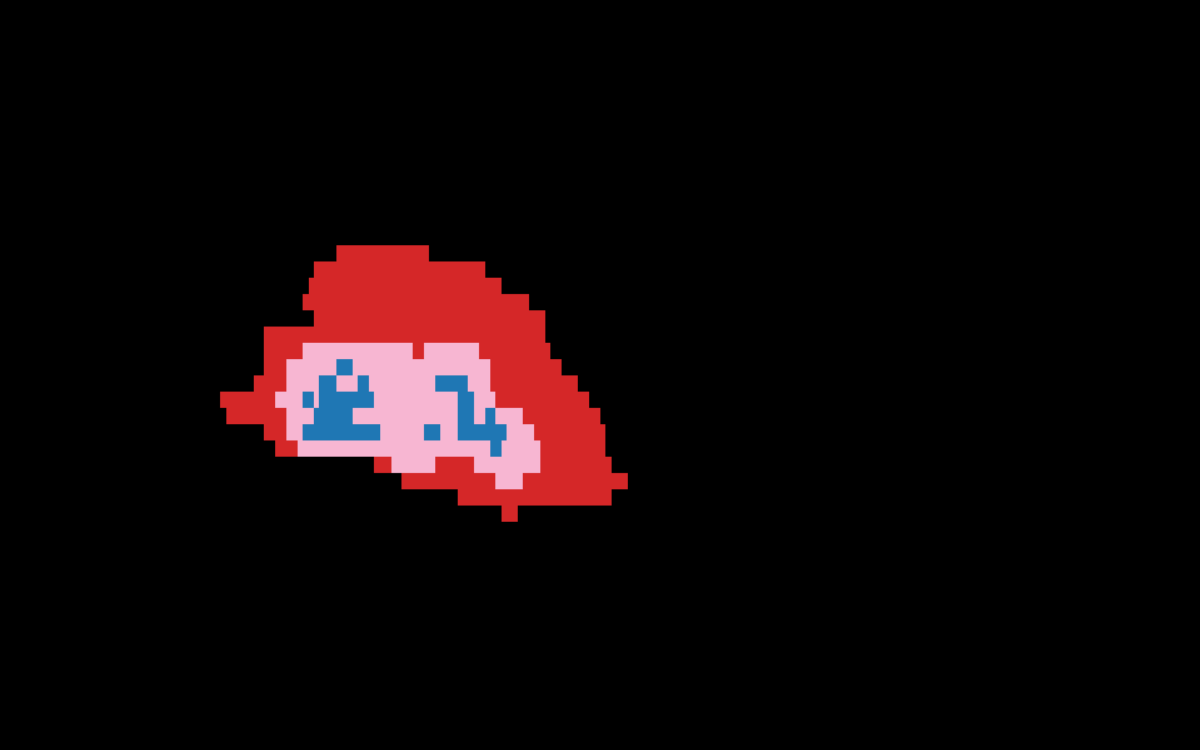}}
\\[-7pt]

\raisebox{0.0\textwidth}{\rotatebox{90}{\textbf{SFCLI+SicTTA}}}\hspace{2pt}%
\subfigure{\includegraphics[width=0.16\textwidth]{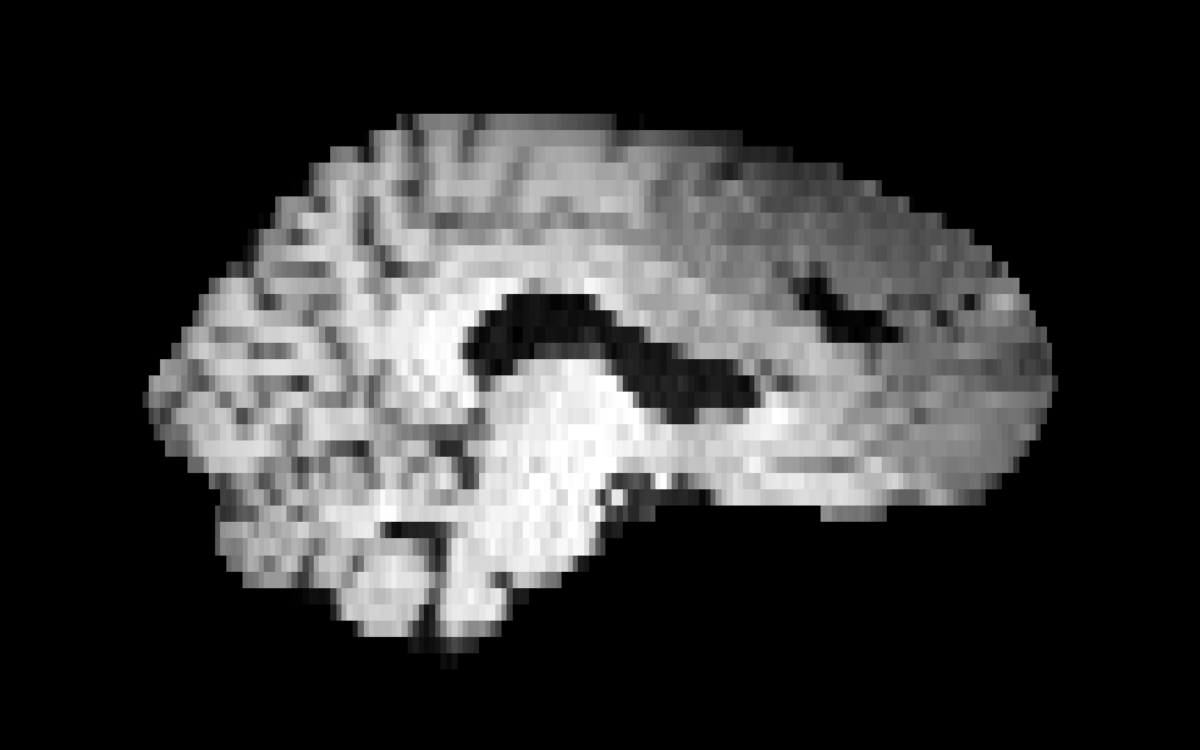}}
\hspace{-4pt}
\subfigure{\includegraphics[width=0.16\textwidth]{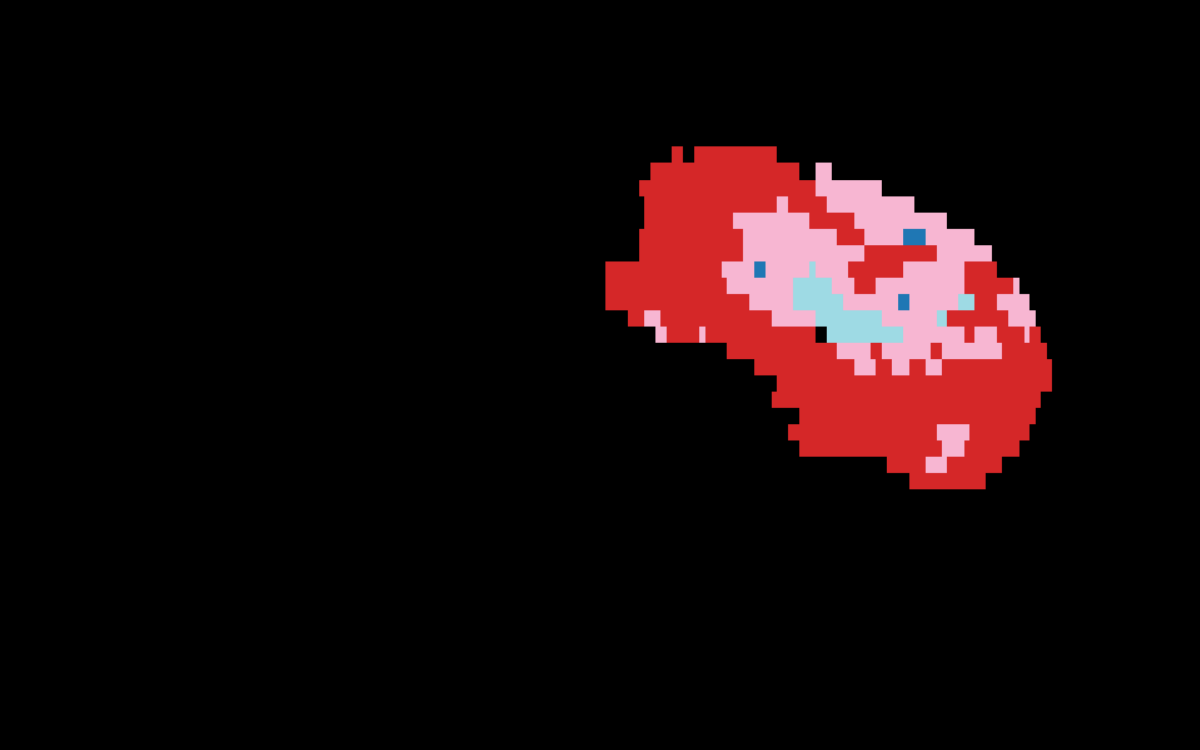}}
\hspace{-4pt}
\subfigure{\includegraphics[width=0.16\textwidth]{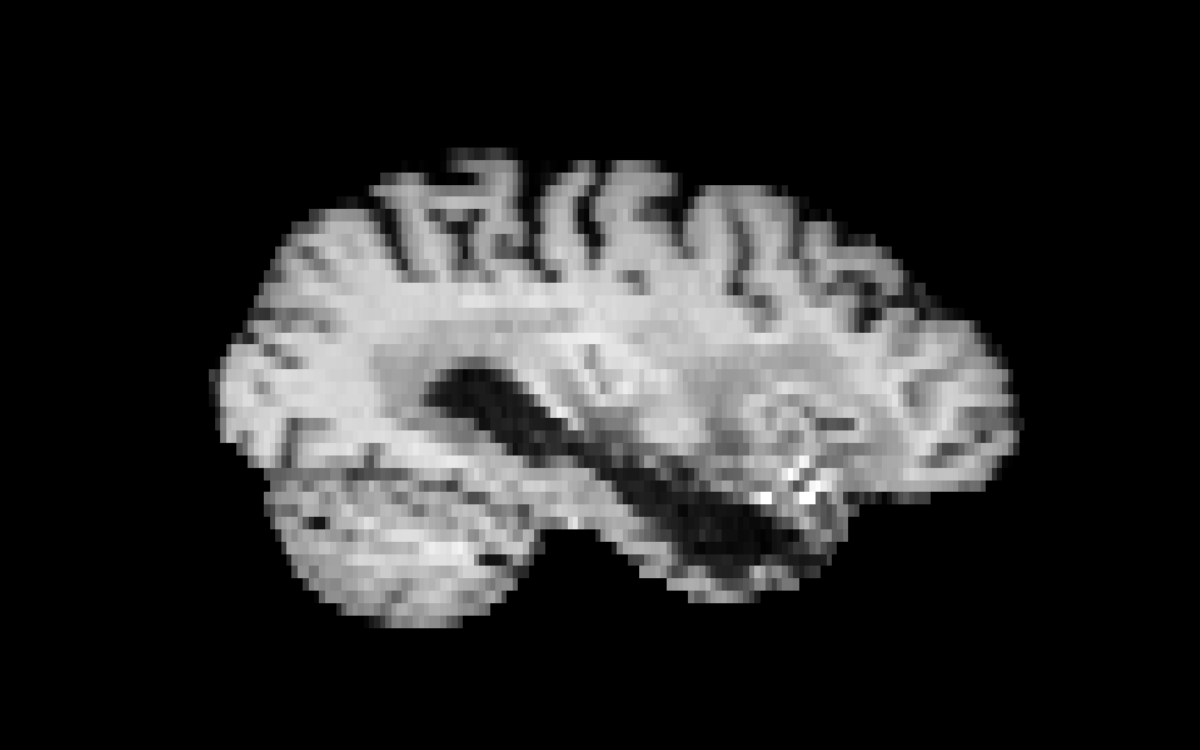}}
\hspace{-4pt}
\subfigure{\includegraphics[width=0.16\textwidth]{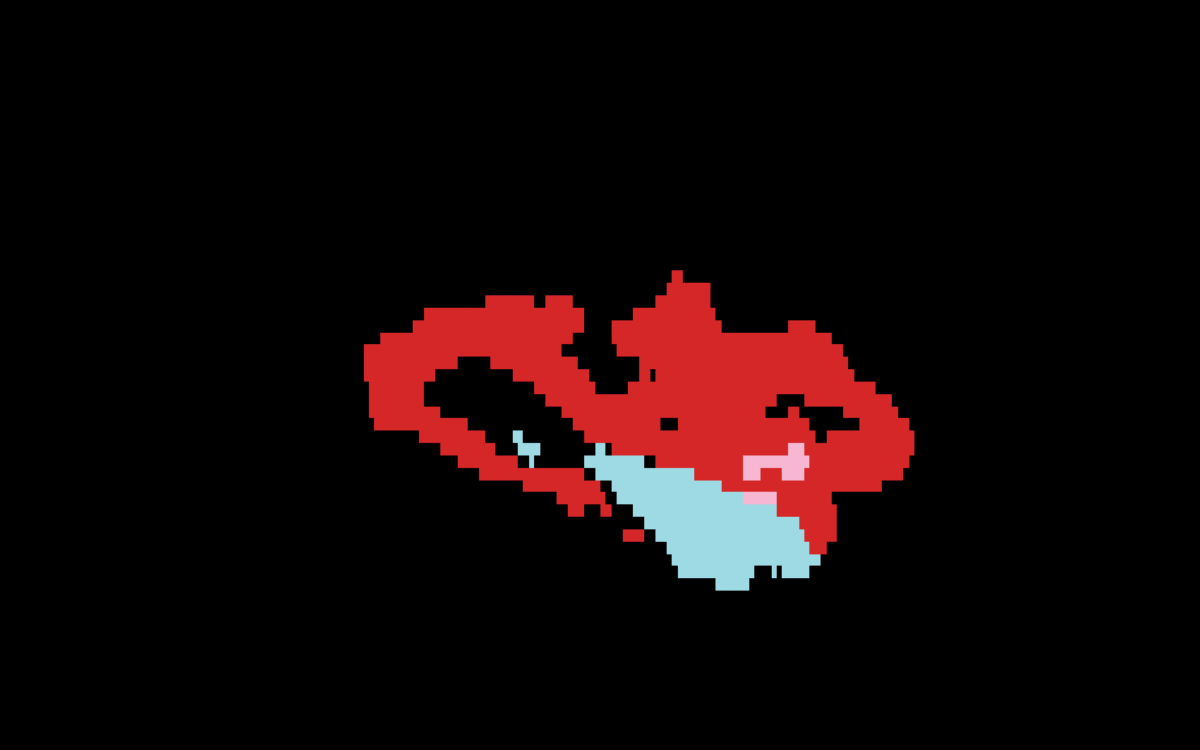}}
\hspace{-4pt}
\subfigure{\includegraphics[width=0.16\textwidth]{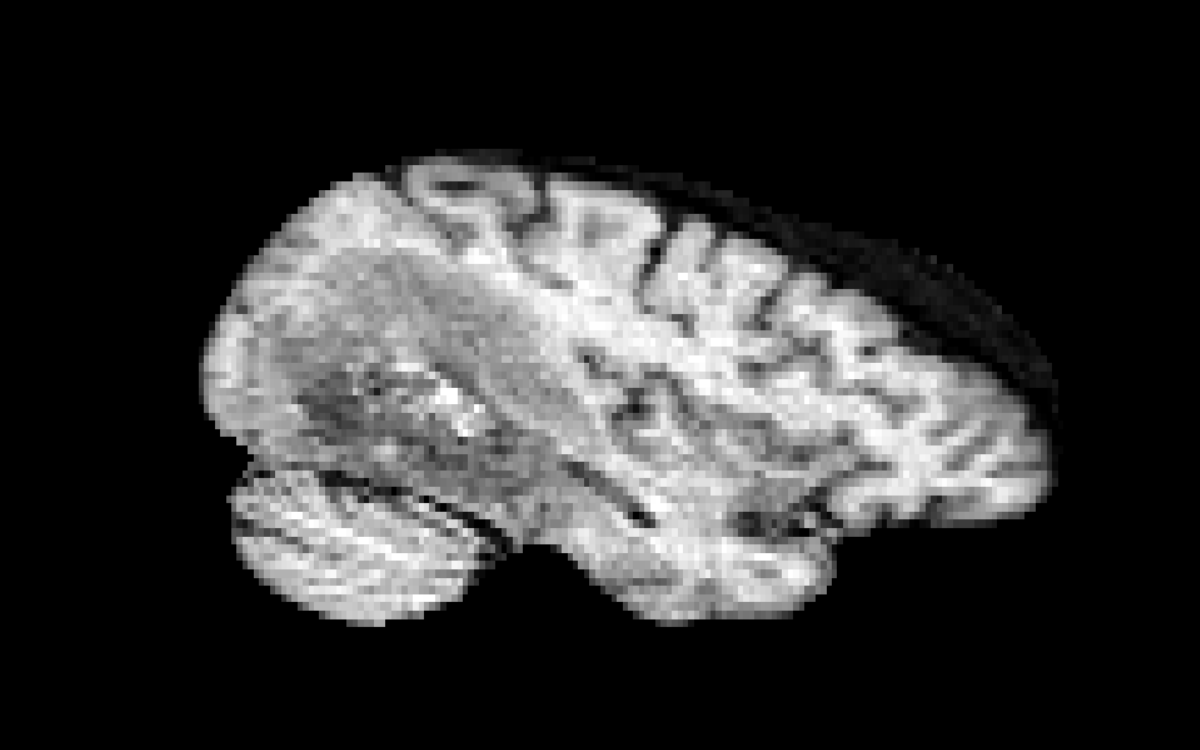}}
\hspace{-4pt}
\subfigure{\includegraphics[width=0.16\textwidth]{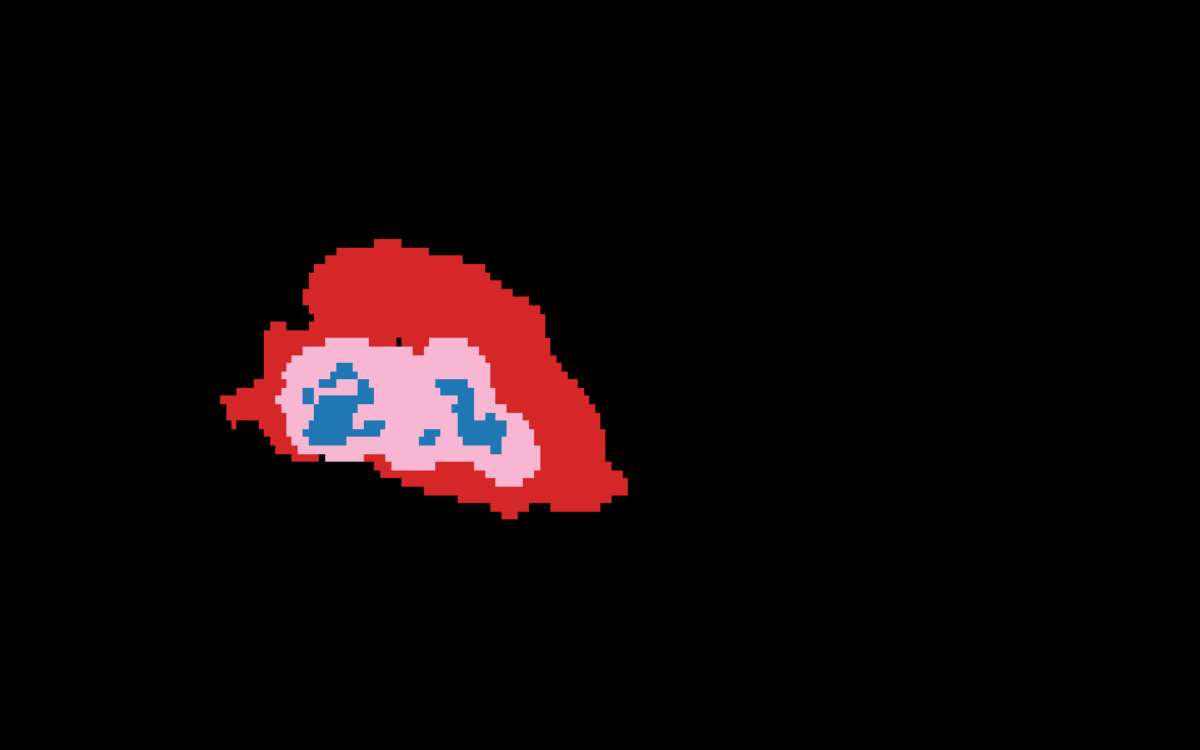}}
\\[-7pt]

\raisebox{0.04\textwidth}{\rotatebox{90}{\textbf{Ours}}}\hspace{2pt}%
\subfigure{\includegraphics[width=0.16\textwidth]{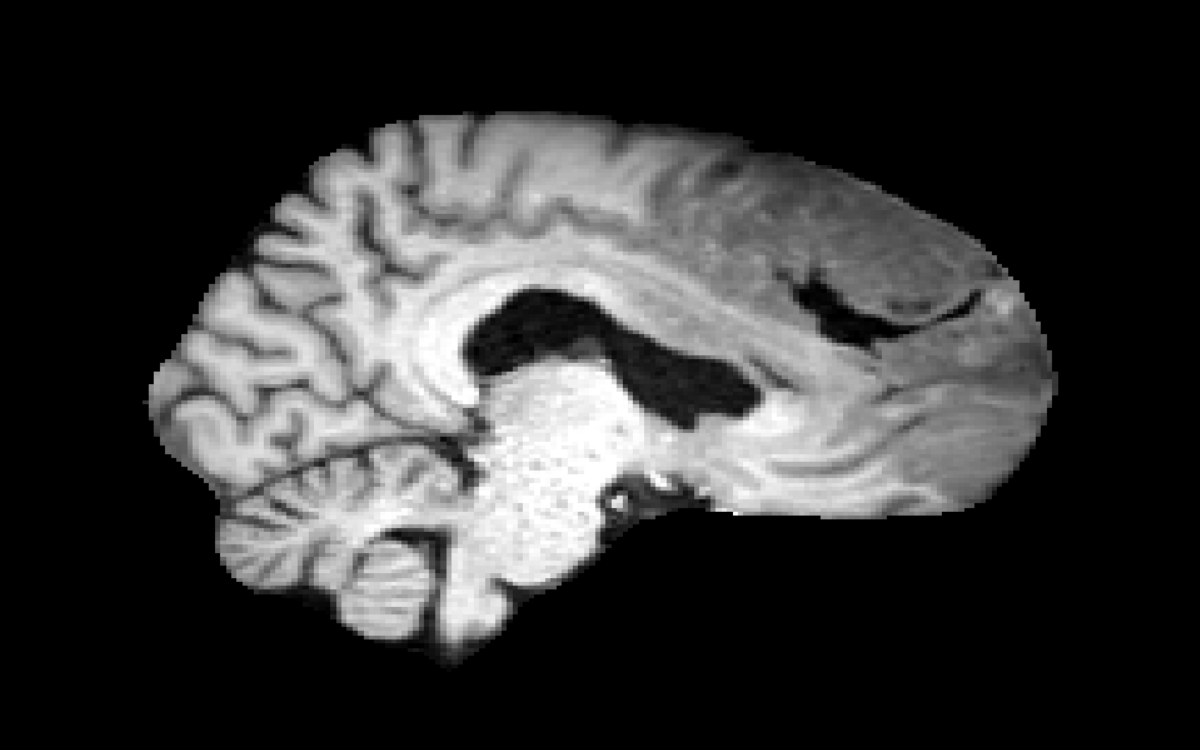}}
\hspace{-4pt}
\subfigure{\includegraphics[width=0.16\textwidth]{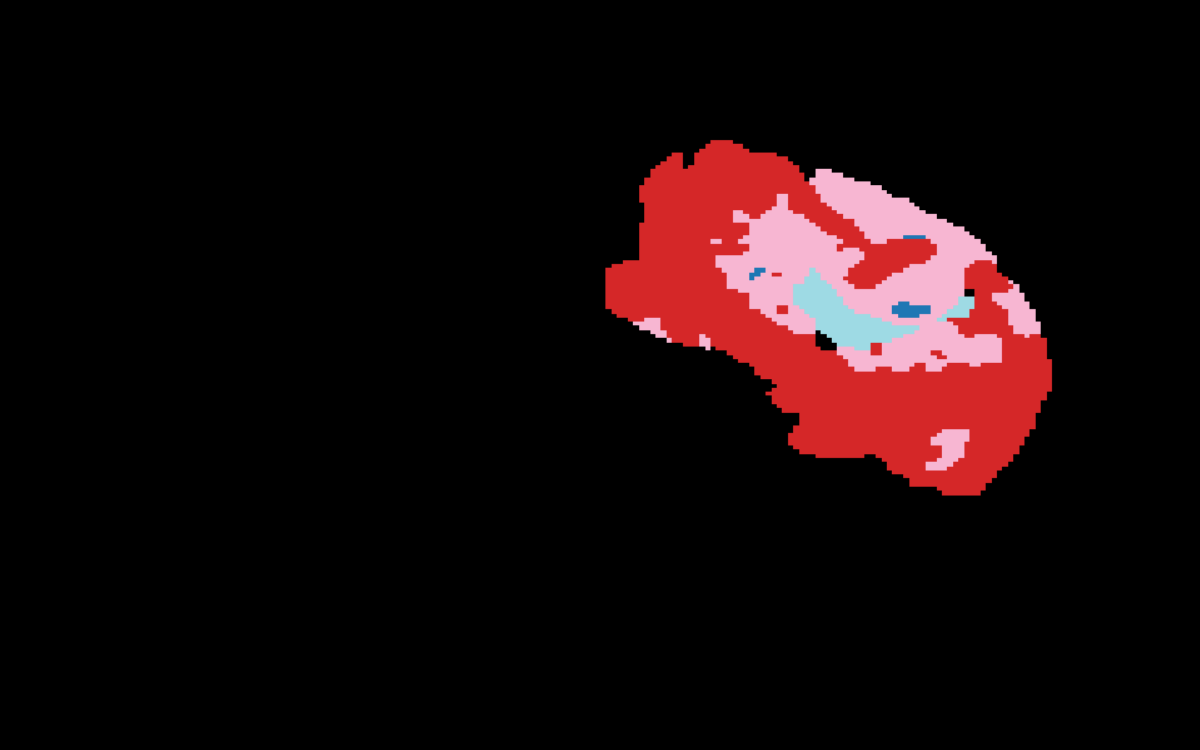}}
\hspace{-4pt}
\subfigure{\includegraphics[width=0.16\textwidth]{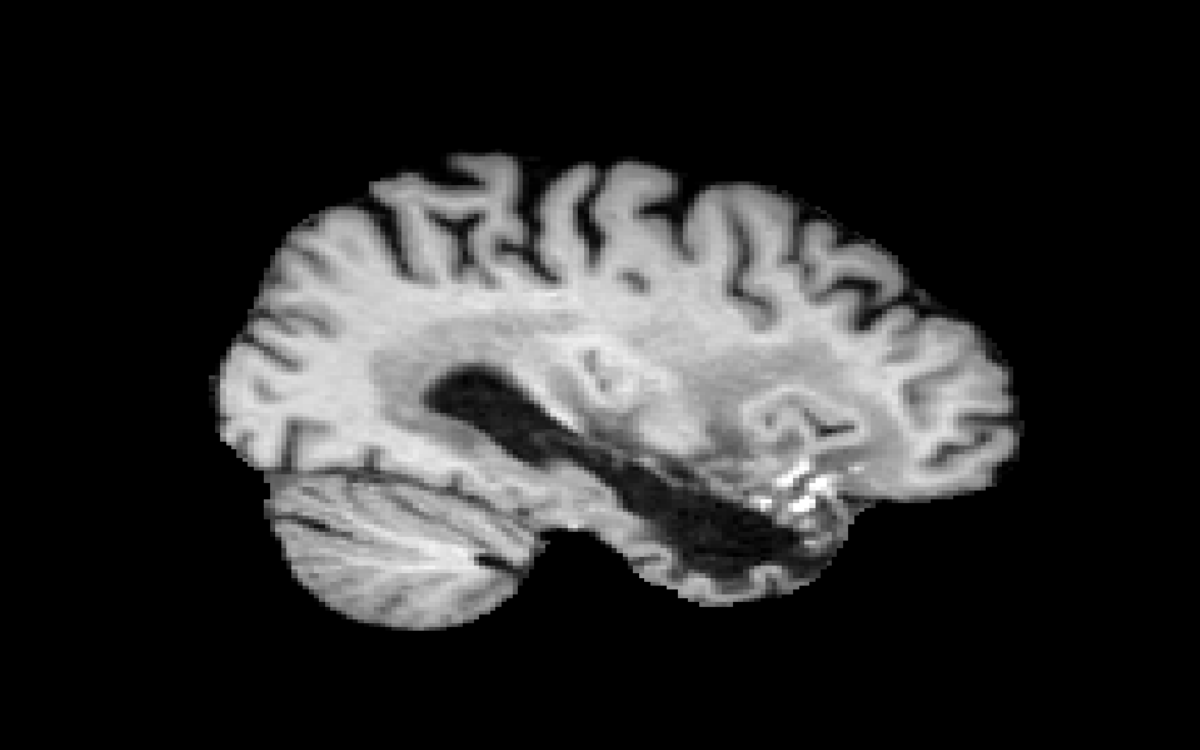}}
\hspace{-4pt}
\subfigure{\includegraphics[width=0.16\textwidth]{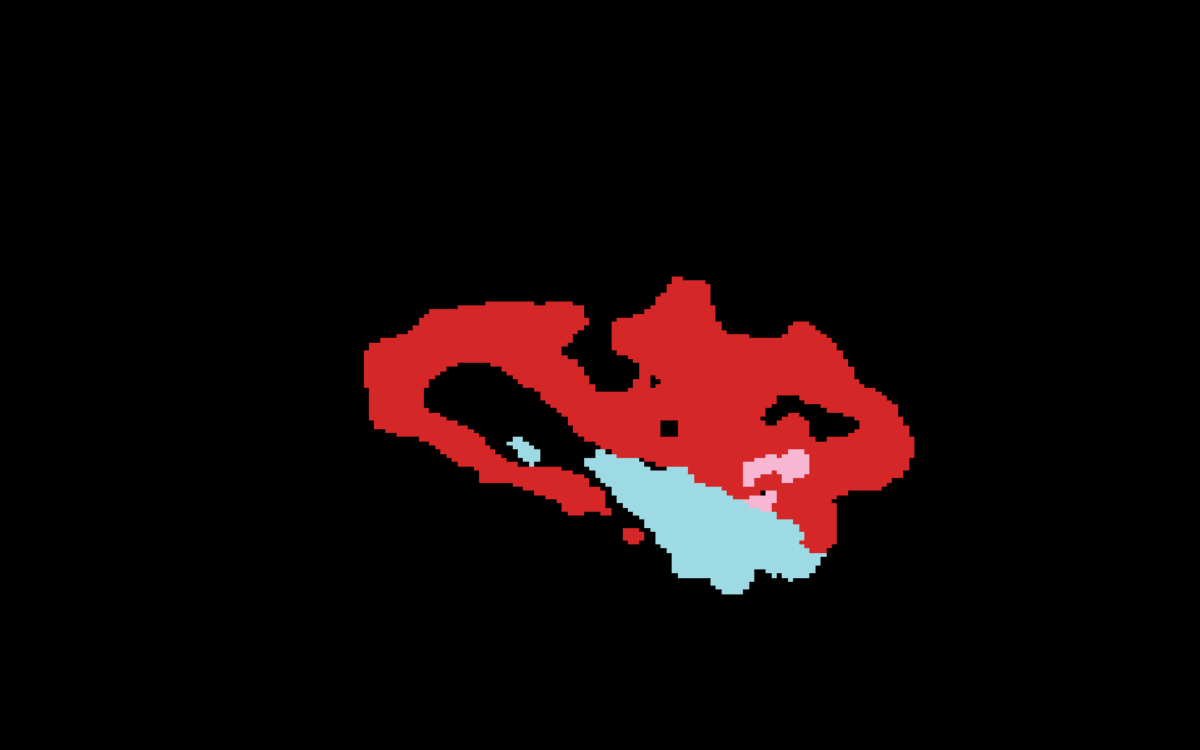}}
\hspace{-4pt}
\subfigure{\includegraphics[width=0.16\textwidth]{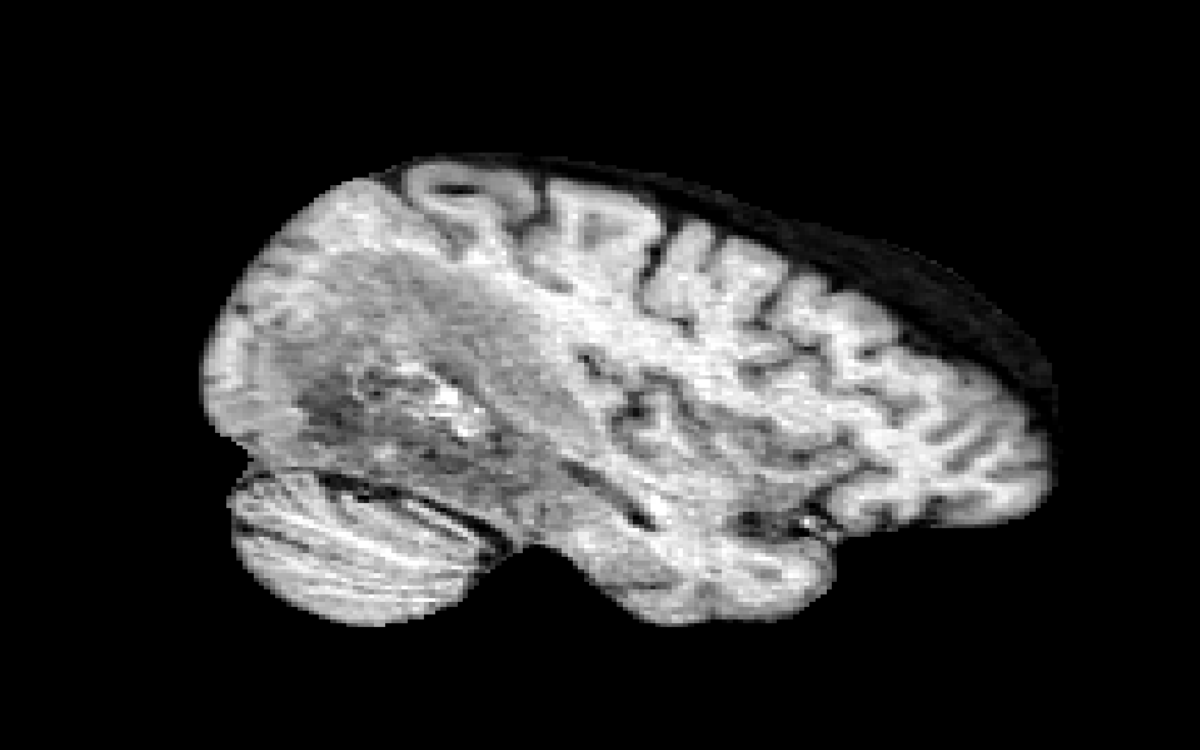}}
\hspace{-4pt}
\subfigure{\includegraphics[width=0.16\textwidth]{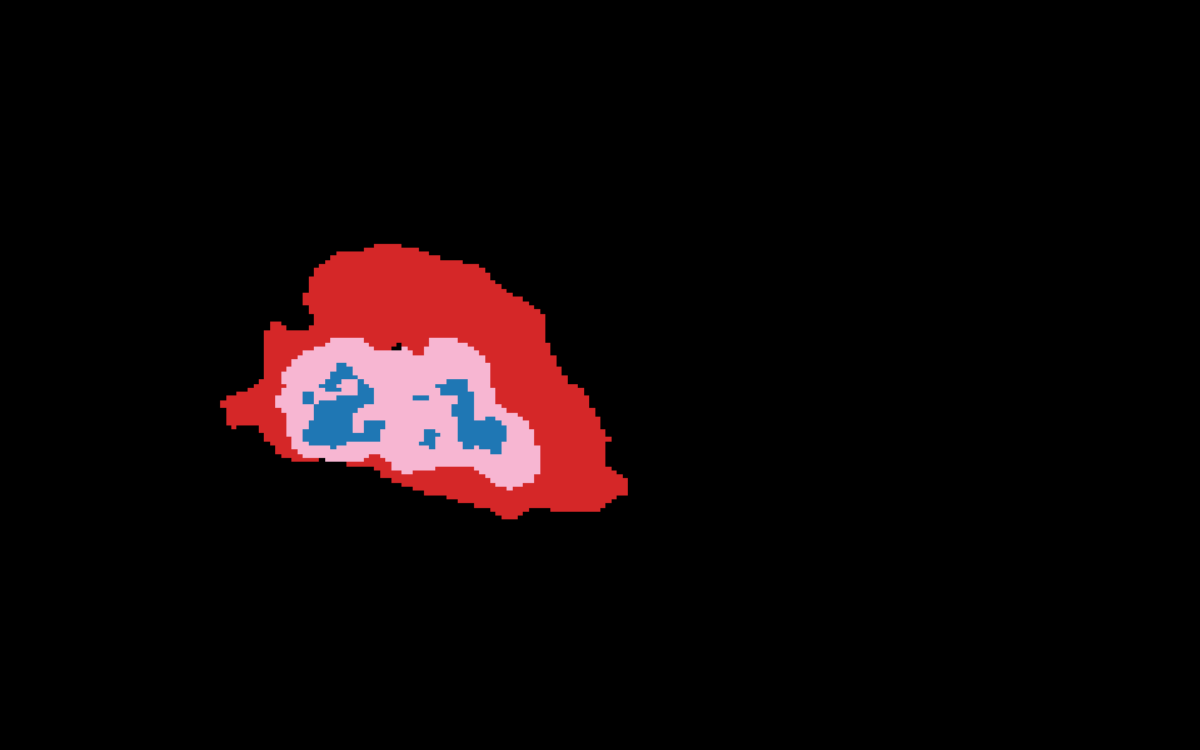}}
\\[-2pt]

\raisebox{0.04\textwidth}{\rotatebox{90}{\textbf{GT}}}\hspace{2pt}%
\subfigure{\includegraphics[width=0.16\textwidth]{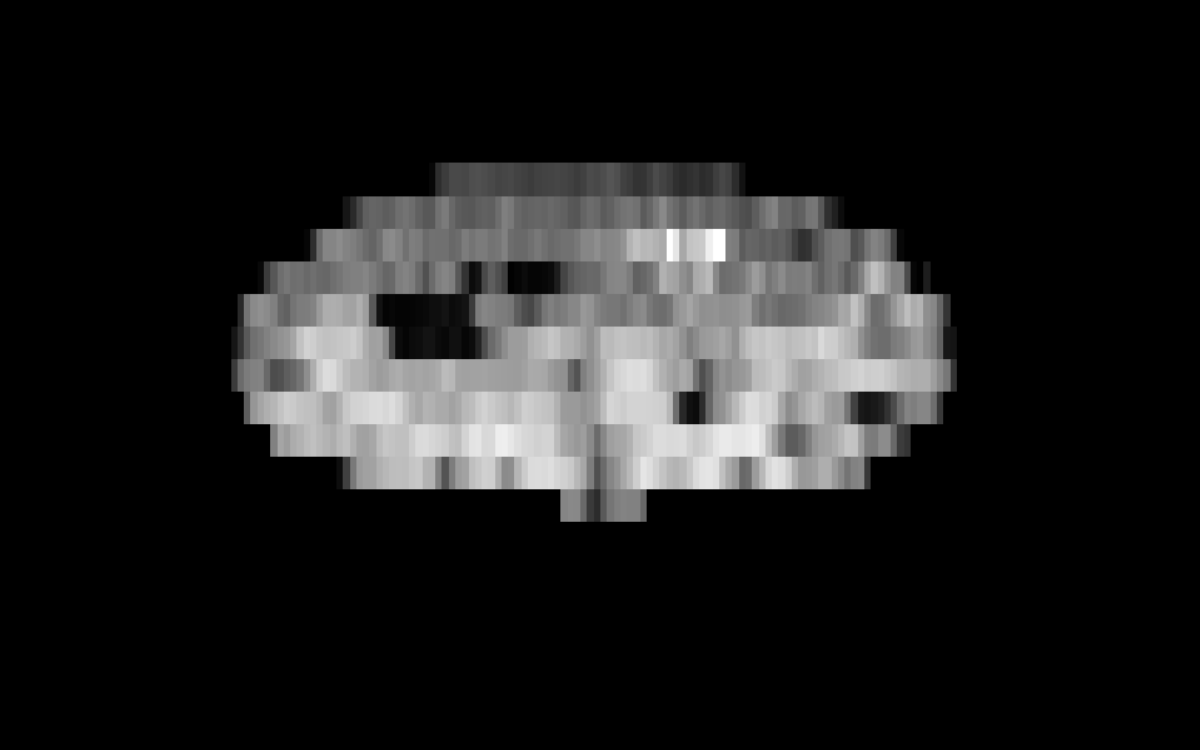}}
\hspace{-4pt}
\subfigure{\includegraphics[width=0.16\textwidth]{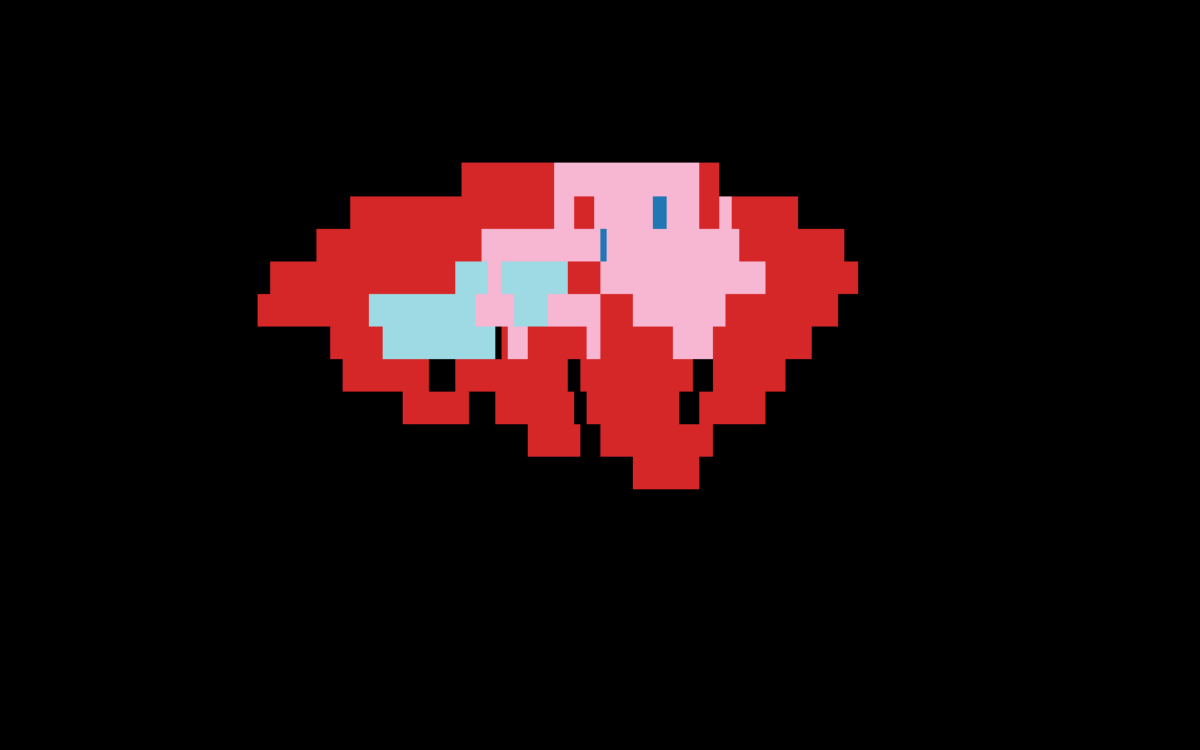}}
\hspace{-4pt}
\subfigure{\includegraphics[width=0.16\textwidth]{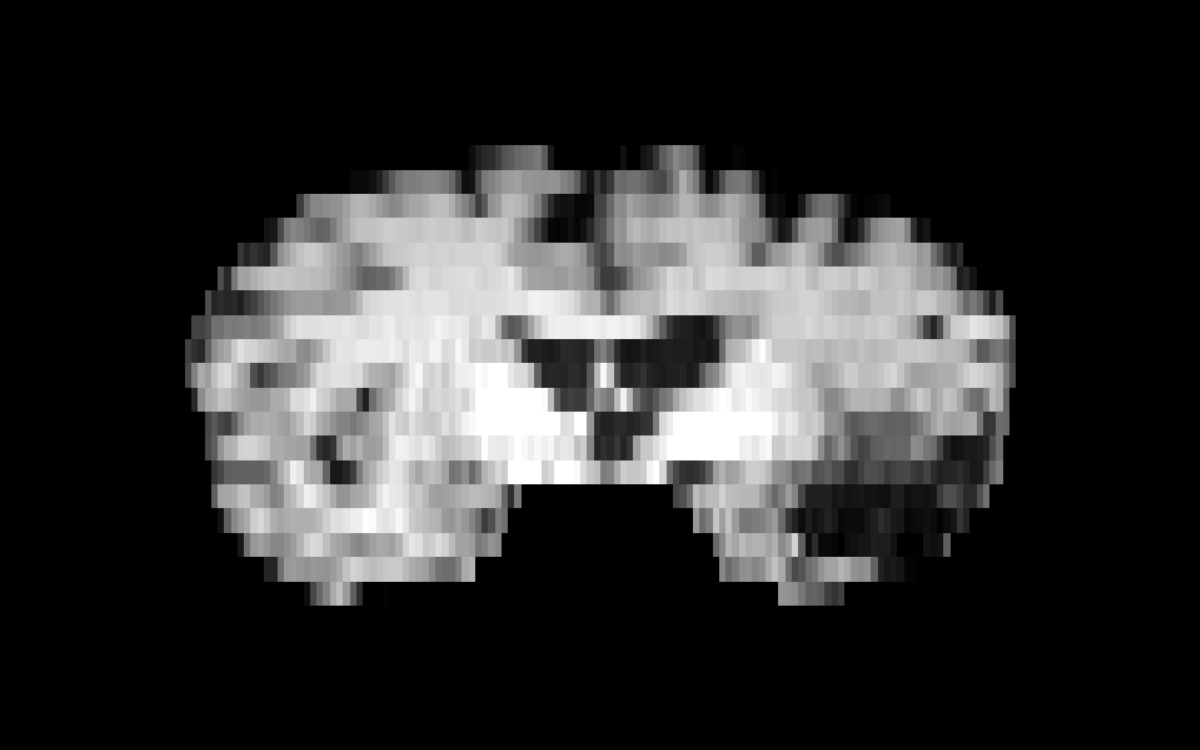}}
\hspace{-4pt}
\subfigure{\includegraphics[width=0.16\textwidth]{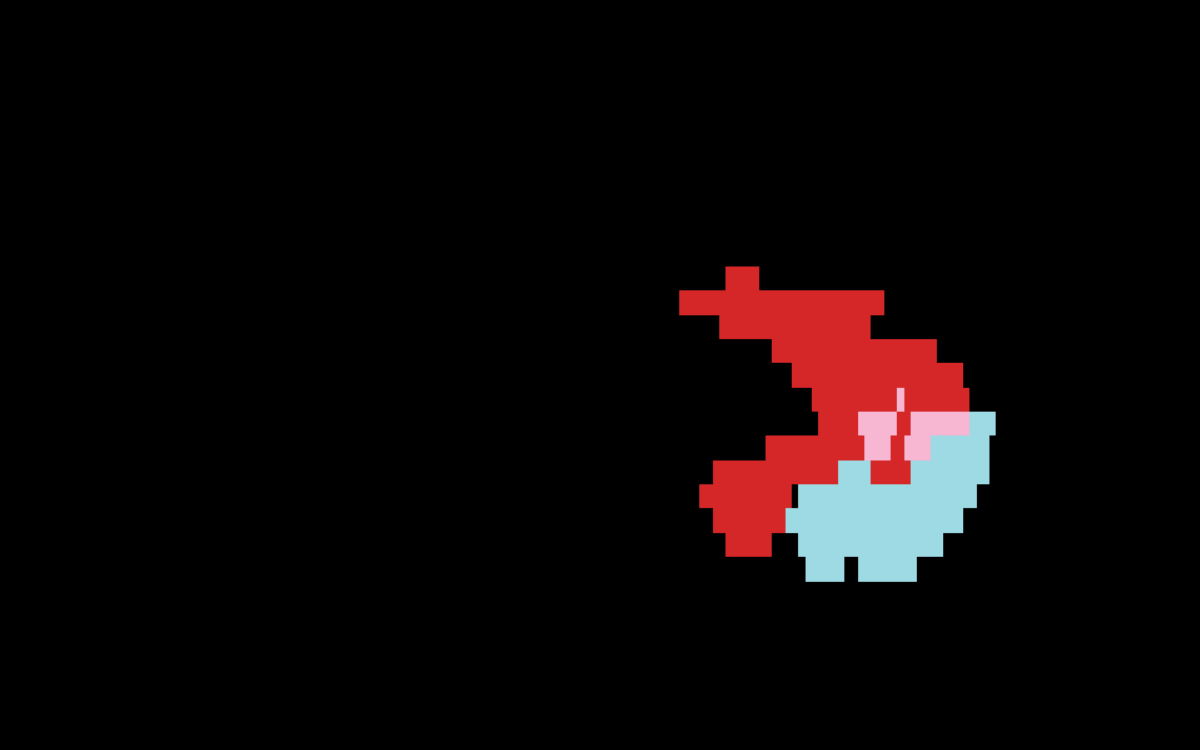}}
\hspace{-4pt}
\subfigure{\includegraphics[width=0.16\textwidth]{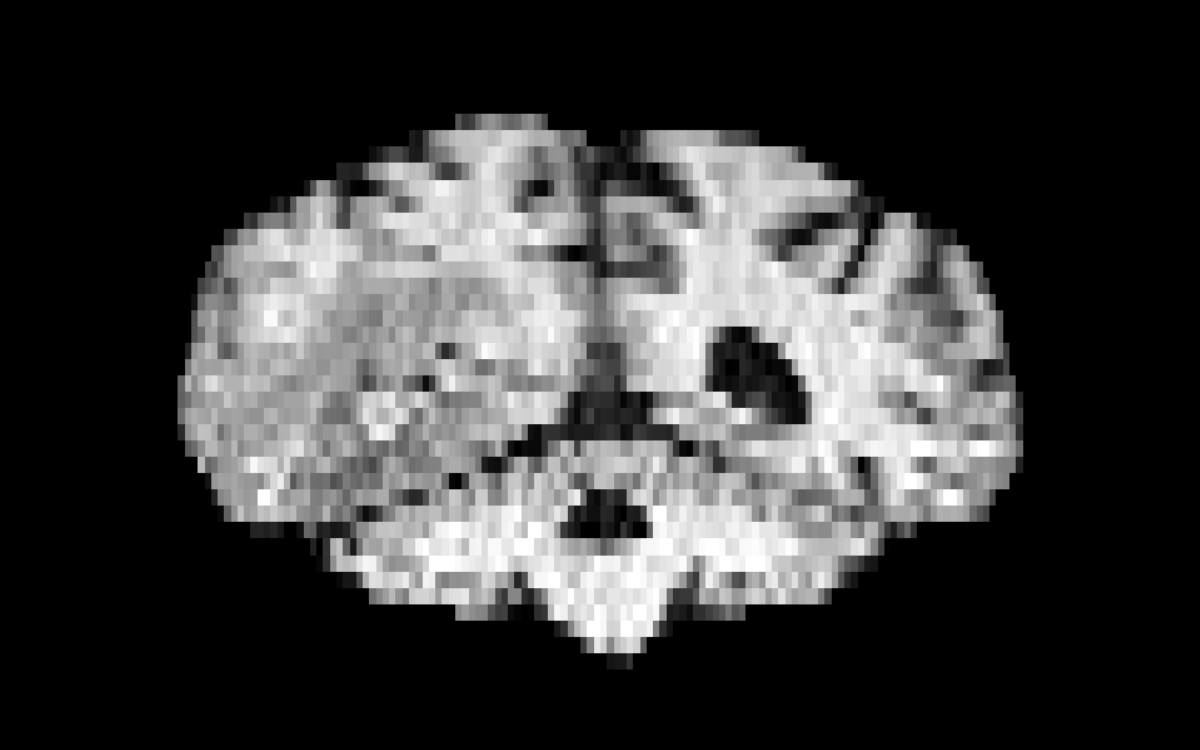}}
\hspace{-4pt}
\subfigure{\includegraphics[width=0.16\textwidth]{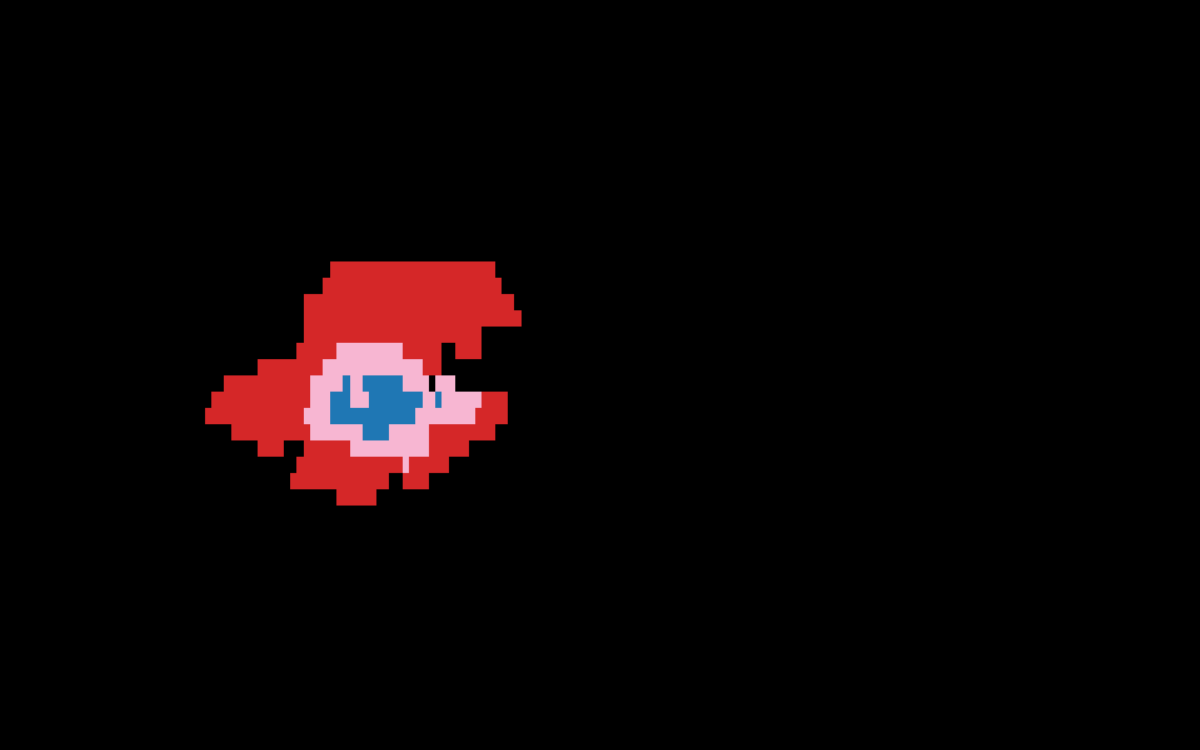}}
\\[-7pt]

\raisebox{0.0\textwidth}{\rotatebox{90}{\textbf{SFCLI+SicTTA}}}\hspace{2pt}%
\subfigure{\includegraphics[width=0.16\textwidth]{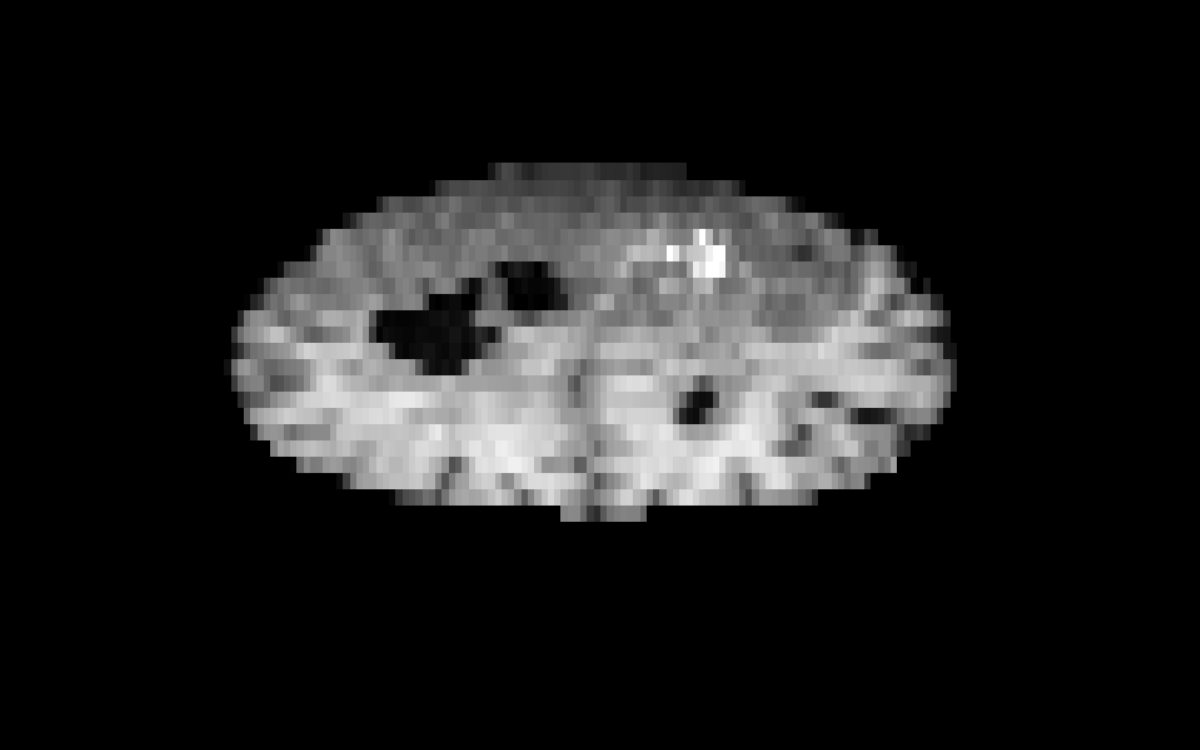}}
\hspace{-4pt}
\subfigure{\includegraphics[width=0.16\textwidth]{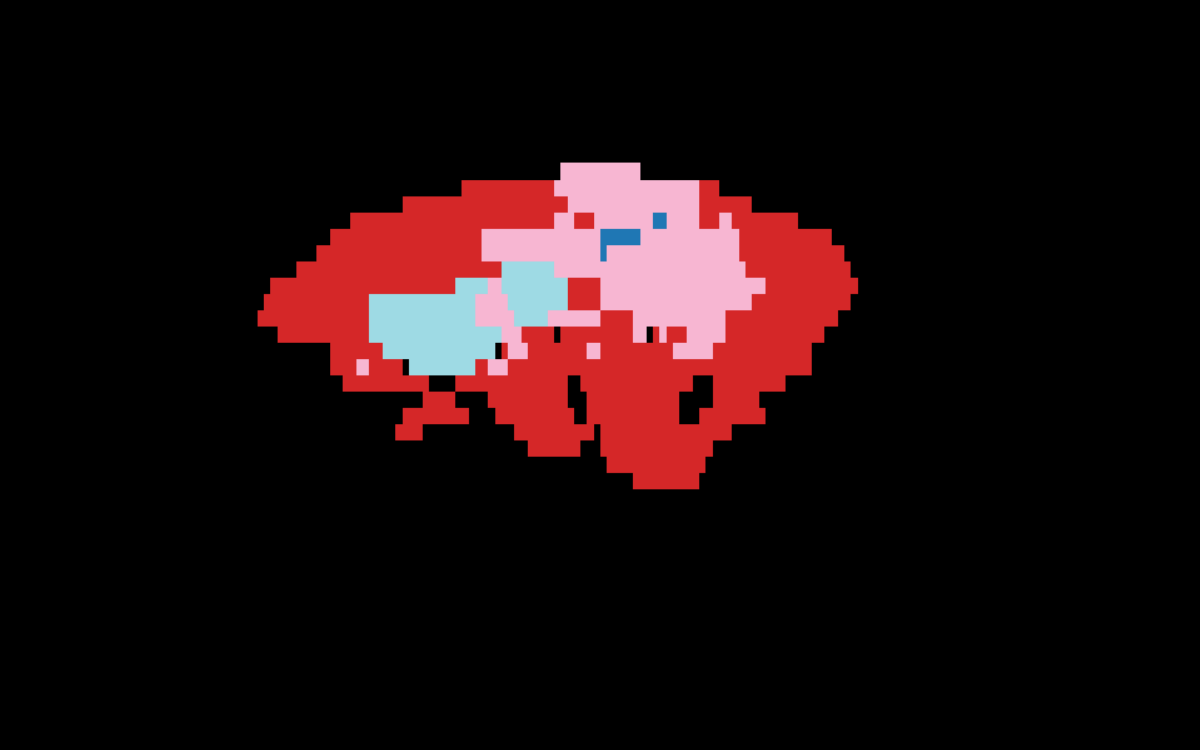}}
\hspace{-4pt}
\subfigure{\includegraphics[width=0.16\textwidth]{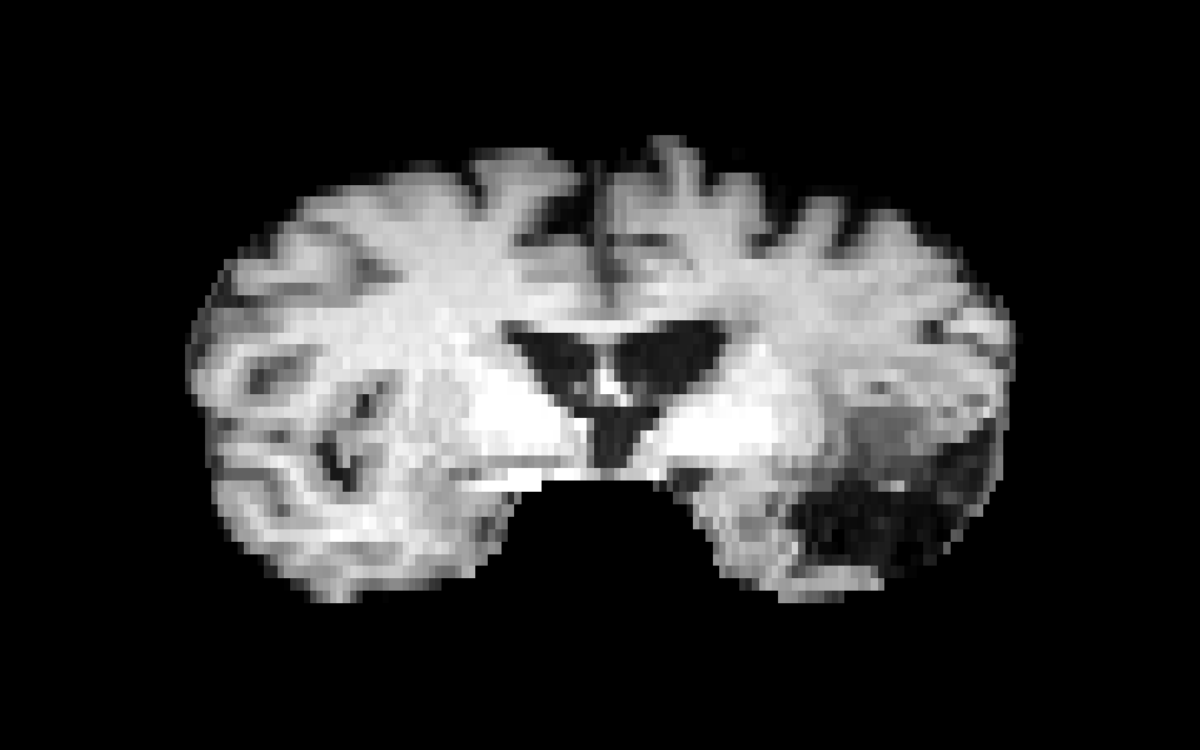}}
\hspace{-4pt}
\subfigure{\includegraphics[width=0.16\textwidth]{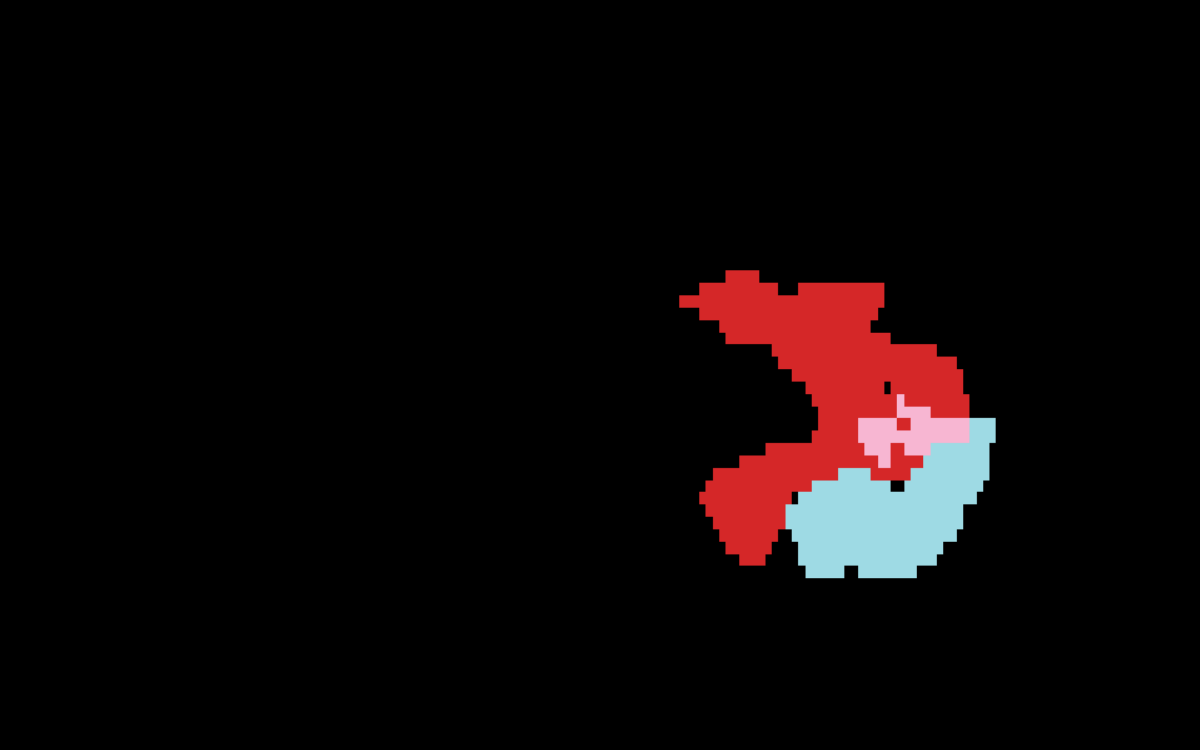}}
\hspace{-4pt}
\subfigure{\includegraphics[width=0.16\textwidth]{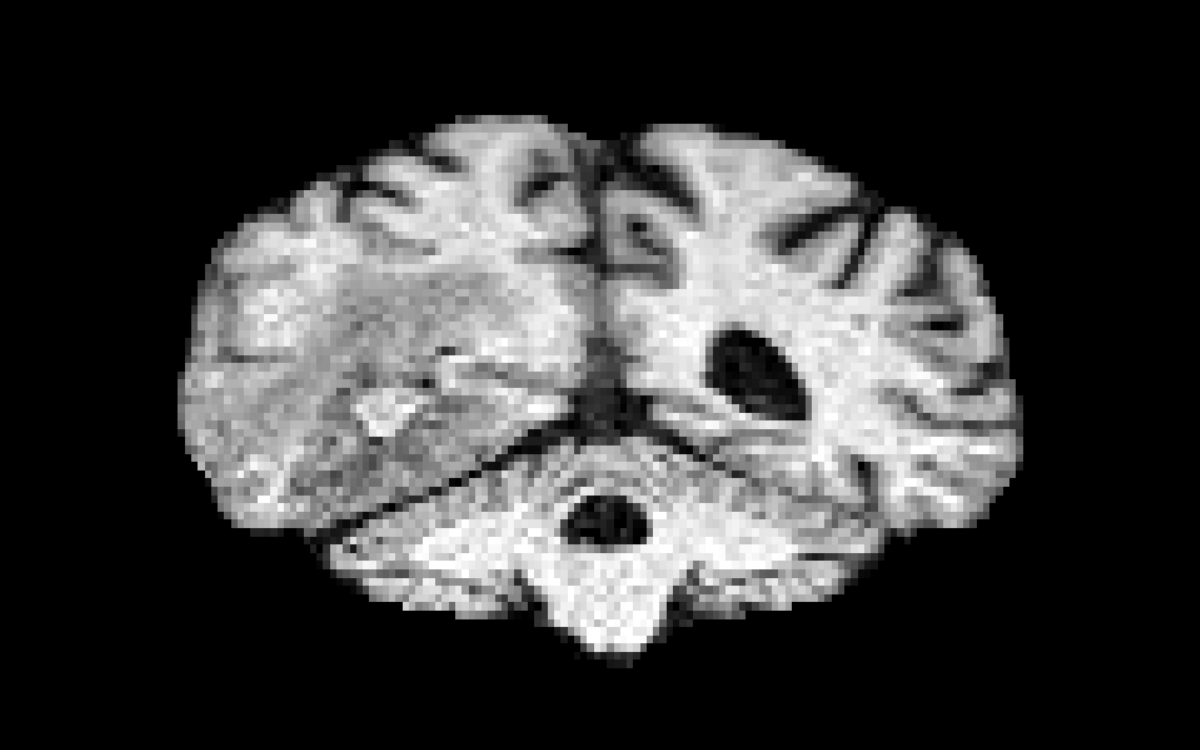}}
\hspace{-4pt}
\subfigure{\includegraphics[width=0.16\textwidth]{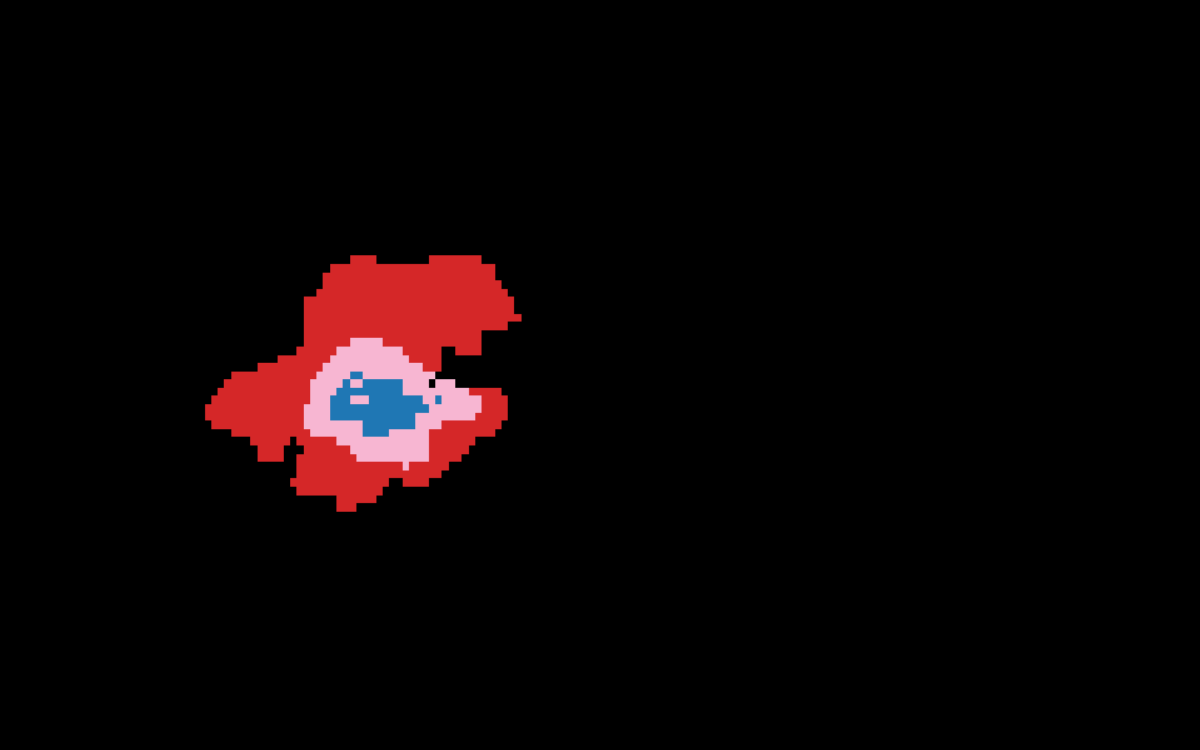}}
\\[-7pt]

\raisebox{0.04\textwidth}{\rotatebox{90}{\textbf{Ours}}}\hspace{2pt}%
\subfigure{\includegraphics[width=0.16\textwidth]{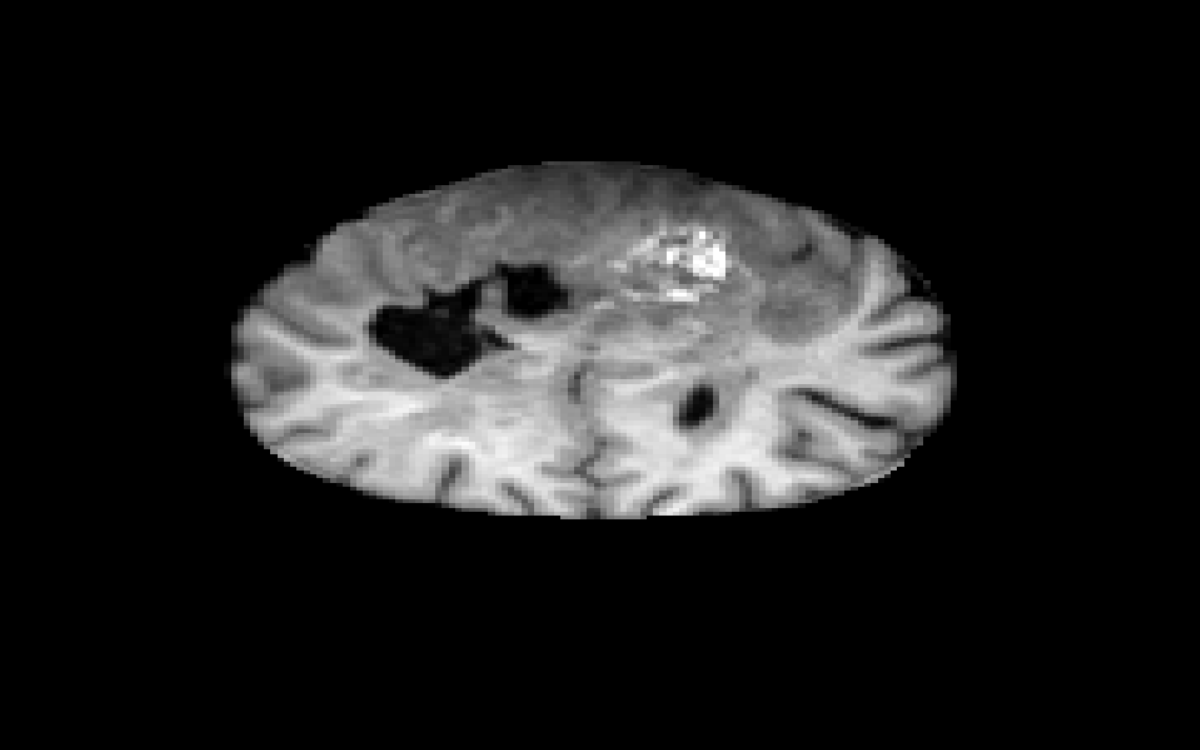}}
\hspace{-4pt}
\subfigure{\includegraphics[width=0.16\textwidth]{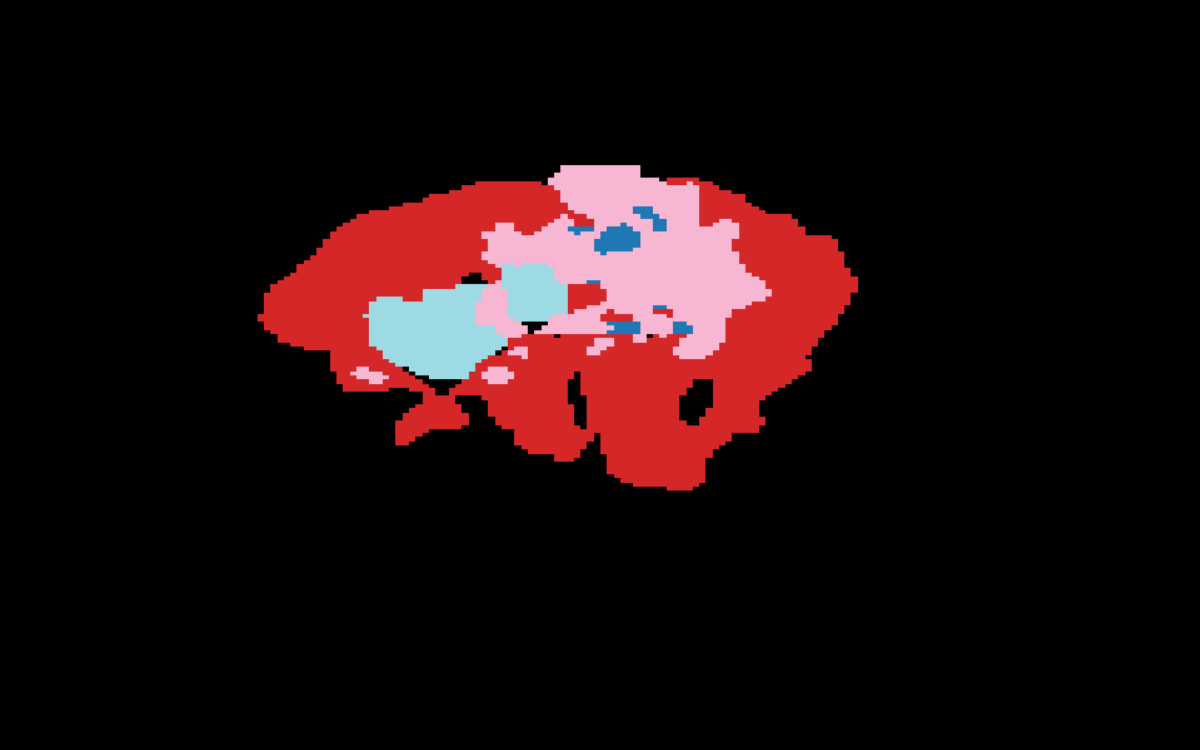}}
\hspace{-4pt}
\subfigure{\includegraphics[width=0.16\textwidth]{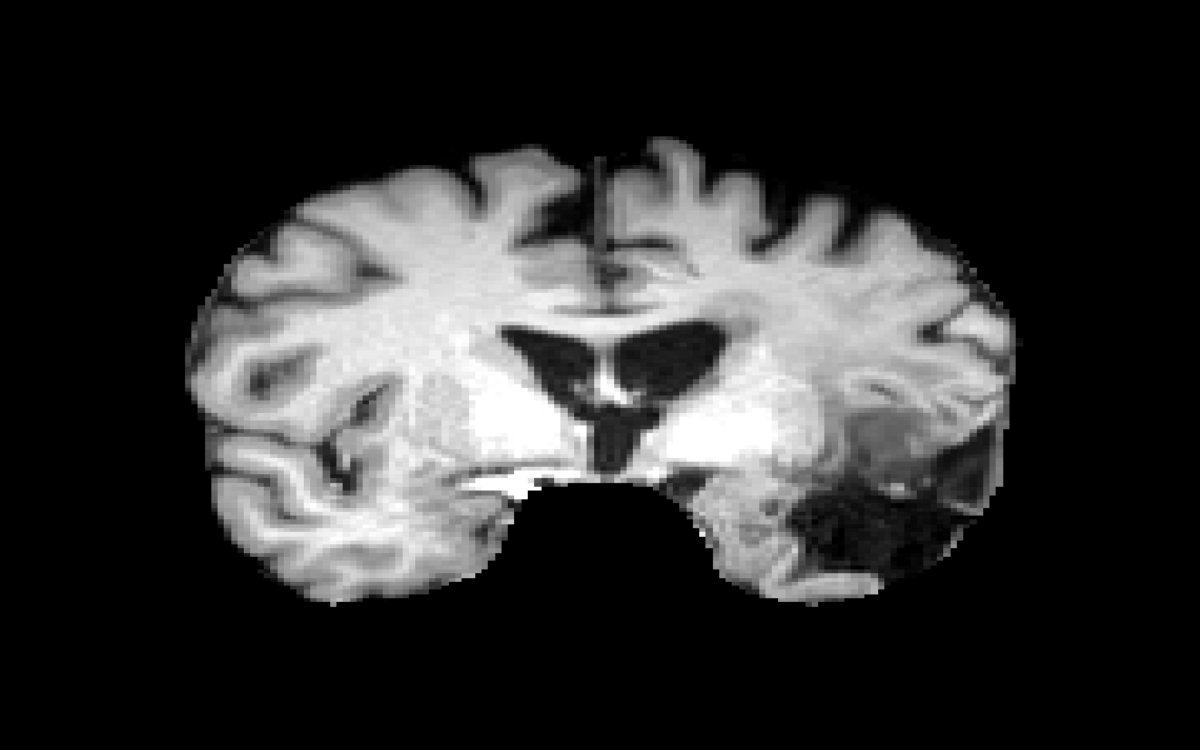}}
\hspace{-4pt}
\subfigure{\includegraphics[width=0.16\textwidth]{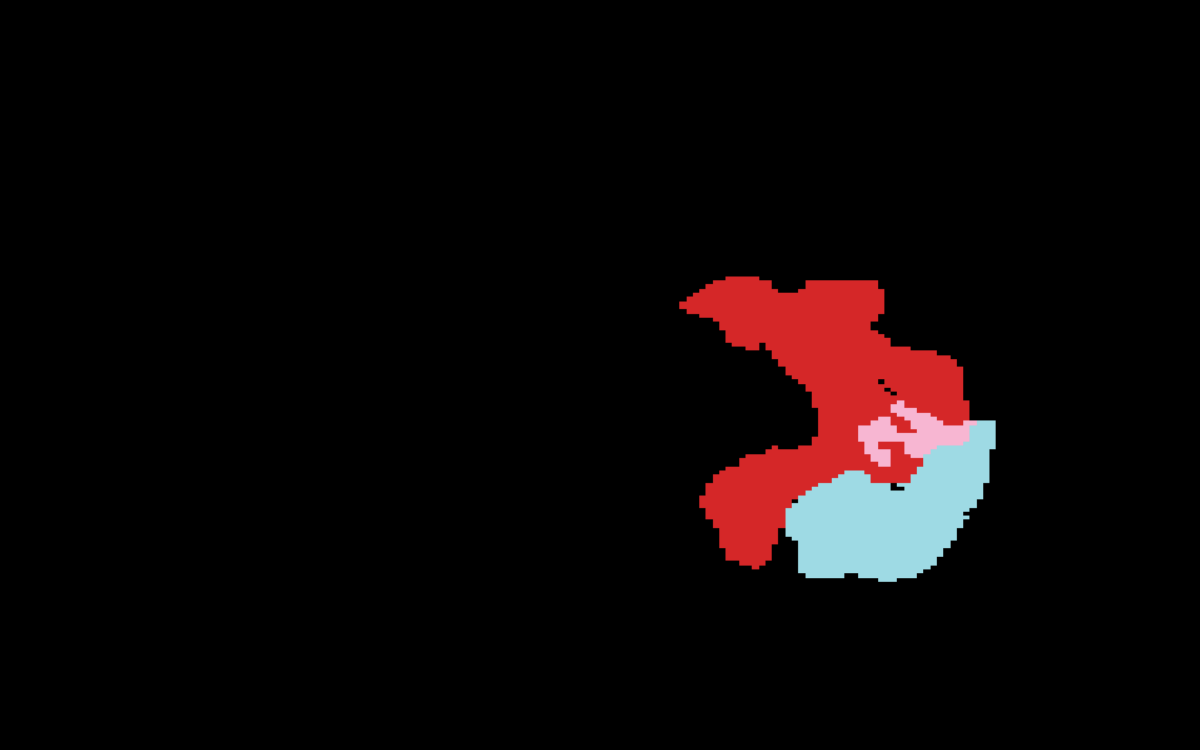}}
\hspace{-4pt}
\subfigure{\includegraphics[width=0.16\textwidth]{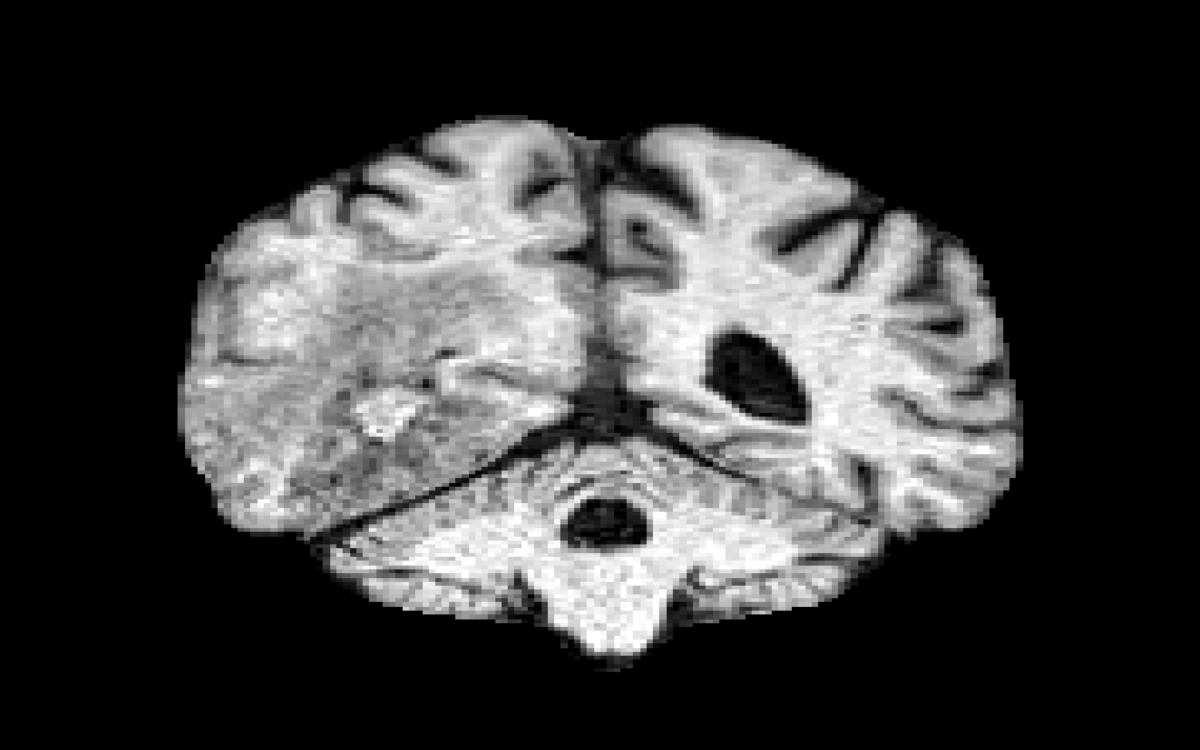}}
\hspace{-4pt}
\subfigure{\includegraphics[width=0.16\textwidth]{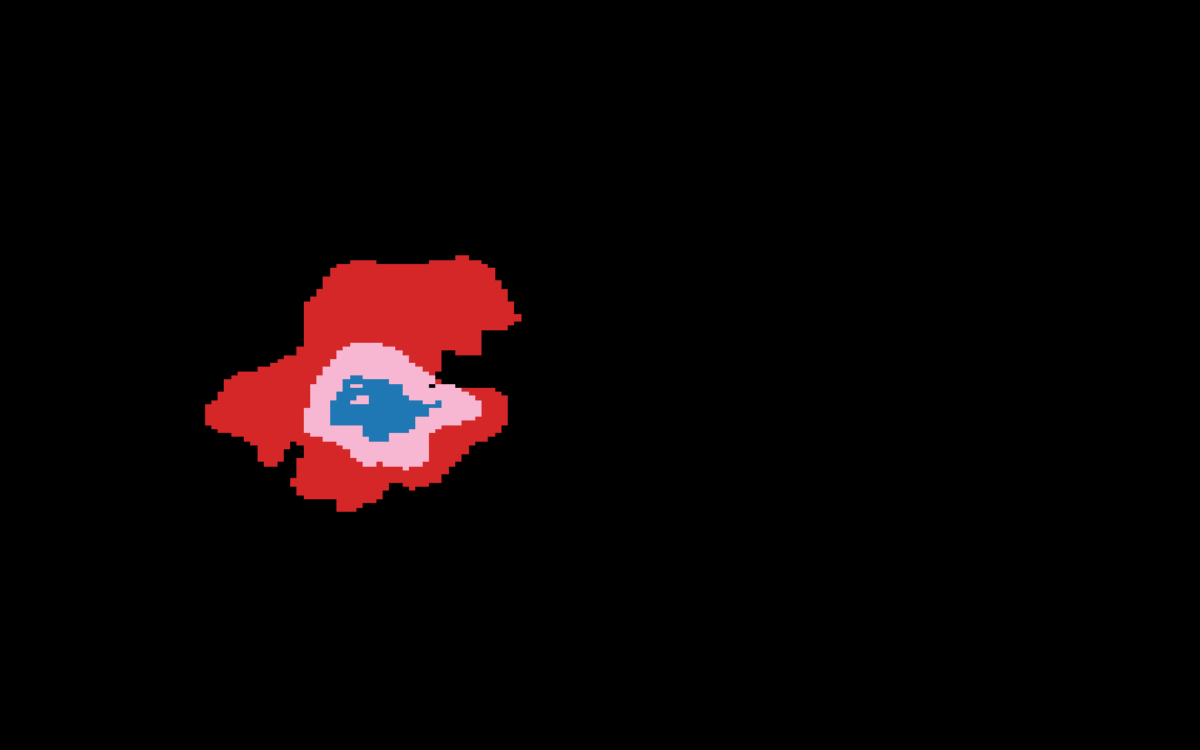}}

\caption{Qualitative results on subsampled BraTS2024 volumes under three z-axis sampling intervals. Rows 1-3 show sagittal views and rows 4-6 show coronal views.}
\label{fig:sparse_dense_vis}
\end{figure*}

We also present sagittal and coronal slice visualizations on subsampled BraTS2024 volumes to comprehensively compare interpolation quality and segmentation accuracy under different z-axis intervals. As shown in Fig. \ref{fig:sparse_dense_vis}, the two-stage baseline still suffers from structural discontinuity and blurry lesion boundaries under large z-axis gaps. Our Position-Aware bidirectional interaction mechanism consistently reconstructs continuous anatomical textures and yields sharp, reliable segmentation masks across all three sampling spacings in both sagittal and coronal views.

\subsection{Performance under Different Interpolation Positions}
To verify the generalization of the proposed joint framework across arbitrary slice positions, we evaluate interpolation SSIM and segmentation Dice on MSD Pancreas and AMOS 2022 at four relative target positions. All experiments are conducted under a fixed 4 mm through-plane sampling spacing. Three cascaded interpolation-segmentation baselines are adopted for comparison, where position 0.5 corresponds to the default mid-slice setting.

All methods undergo gradual performance degradation as the target position deviates from the slice center, owing to increased anatomical uncertainty. CoInS-Net consistently outperforms all cascaded baselines across all positions on both datasets, with a milder performance drop at edge positions. The stable performance benefits from the spatially continuous position interpolation module, which dynamically adjusts feature fusion weights according to relative target coordinates and physical spacing. Quantitative results are presented in Tables \ref{tab:position_1} and \ref{tab:position_2}.

\begin{table*}[!tbp]
\centering
\fontsize{7}{9}\selectfont
\setlength{\tabcolsep}{3pt}
\caption{Performance comparison at interpolation positions 0.2 and 0.33}
\label{tab:position_1}
\begin{tabular}{@{}lcccccccc@{}}
\toprule
\multirow{3}{*}{Model} & \multicolumn{4}{c}{Position 0.2} & \multicolumn{4}{c}{Position 0.33} \\
\cmidrule(lr){2-5} \cmidrule(lr){6-9}
& \multicolumn{2}{c}{MSD Pancreas} & \multicolumn{2}{c}{AMOS 2022} & \multicolumn{2}{c}{MSD Pancreas} & \multicolumn{2}{c}{AMOS 2022} \\
\cmidrule(lr){2-3} \cmidrule(lr){4-5} \cmidrule(lr){6-7} \cmidrule(lr){8-9}
& SSIM & Dice & SSIM & Dice & SSIM & Dice & SSIM & Dice \\
\midrule
McASSR+Swin UNETR & 0.924$\pm$0.0038$^{***}$ & 0.908$\pm$0.016$^{***}$ & 0.920$\pm$0.0045$^{***}$ & 0.779$\pm$0.021$^{***}$ & 0.940$\pm$0.0029$^{***}$ & 0.937$\pm$0.012$^{***}$ & 0.939$\pm$0.0034$^{***}$ & 0.816$\pm$0.015$^{***}$ \\
I$^3$Net+HiDiff & 0.919$\pm$0.0052$^{***}$ & 0.910$\pm$0.019$^{***}$ & 0.912$\pm$0.0061$^{***}$ & 0.782$\pm$0.024$^{**}$ & 0.935$\pm$0.0033$^{***}$ & 0.939$\pm$0.014$^{***}$ & 0.930$\pm$0.0041$^{***}$ & 0.819$\pm$0.018$^{***}$ \\
SFCLI-Net+SicTTA & 0.920$\pm$0.0044$^{***}$ & 0.917$\pm$0.017$^{**}$ & 0.917$\pm$0.0053$^{***}$ & 0.790$\pm$0.022$^{***}$ & 0.942$\pm$0.0025$^{***}$ & 0.946$\pm$0.011$^{**}$ & 0.936$\pm$0.0031$^{***}$ & 0.827$\pm$0.013$^{**}$ \\
Ours & \textbf{0.946$\pm$0.0011} & \textbf{0.956$\pm$0.009} & \textbf{0.946$\pm$0.0019} & \textbf{0.843$\pm$0.012} & \textbf{0.956$\pm$0.0008} & \textbf{0.972$\pm$0.008} & \textbf{0.956$\pm$0.0014} & \textbf{0.865$\pm$0.010} \\
\bottomrule
\end{tabular}
\end{table*}

\begin{table*}[!tbp]
\centering
\fontsize{7}{9}\selectfont
\setlength{\tabcolsep}{3pt}
\caption{Performance comparison at interpolation positions 0.5 and 0.66}
\label{tab:position_2}
\begin{tabular}{@{}lcccccccc@{}}
\toprule
\multirow{3}{*}{Model} & \multicolumn{4}{c}{Position 0.5} & \multicolumn{4}{c}{Position 0.66} \\
\cmidrule(lr){2-5} \cmidrule(lr){6-9}
& \multicolumn{2}{c}{MSD Pancreas} & \multicolumn{2}{c}{AMOS 2022} & \multicolumn{2}{c}{MSD Pancreas} & \multicolumn{2}{c}{AMOS 2022} \\
\cmidrule(lr){2-3} \cmidrule(lr){4-5} \cmidrule(lr){6-7} \cmidrule(lr){8-9}
& SSIM & Dice & SSIM & Dice & SSIM & Dice & SSIM & Dice \\
\midrule
McASSR+Swin UNETR & 0.942$\pm$0.0024$^{***}$ & 0.939$\pm$0.010$^{***}$ & 0.941$\pm$0.0030$^{***}$ & 0.818$\pm$0.013$^{***}$ & 0.938$\pm$0.0032$^{***}$ & 0.935$\pm$0.013$^{***}$ & 0.937$\pm$0.0039$^{***}$ & 0.814$\pm$0.017$^{***}$ \\
I$^3$Net+HiDiff & 0.937$\pm$0.0036$^{***}$ & 0.941$\pm$0.012$^{**}$ & 0.933$\pm$0.0042$^{***}$ & 0.821$\pm$0.014$^{***}$ & 0.933$\pm$0.0027$^{***}$ & 0.937$\pm$0.011$^{***}$ & 0.929$\pm$0.0035$^{***}$ & 0.817$\pm$0.014$^{**}$ \\
SFCLI-Net+SicTTA & 0.945$\pm$0.0020$^{***}$ & 0.948$\pm$0.009$^{***}$ & 0.944$\pm$0.0024$^{***}$ & 0.829$\pm$0.011$^{***}$ & 0.937$\pm$0.0041$^{***}$ & 0.944$\pm$0.015$^{***}$ & 0.934$\pm$0.0049$^{***}$ & 0.825$\pm$0.019$^{*}$ \\
Ours & \textbf{0.958$\pm$0.0005} & \textbf{0.974$\pm$0.006} & \textbf{0.958$\pm$0.0021} & \textbf{0.867$\pm$0.009} & \textbf{0.955$\pm$0.0009} & \textbf{0.970$\pm$0.007} & \textbf{0.955$\pm$0.0016} & \textbf{0.863$\pm$0.010} \\
\bottomrule
\end{tabular}
\end{table*}

\begin{figure*}[!tbp]
\centering
\fontsize{7}{9}\selectfont
\setlength{\tabcolsep}{0.1pt}
\includegraphics[width=1.0\textwidth]{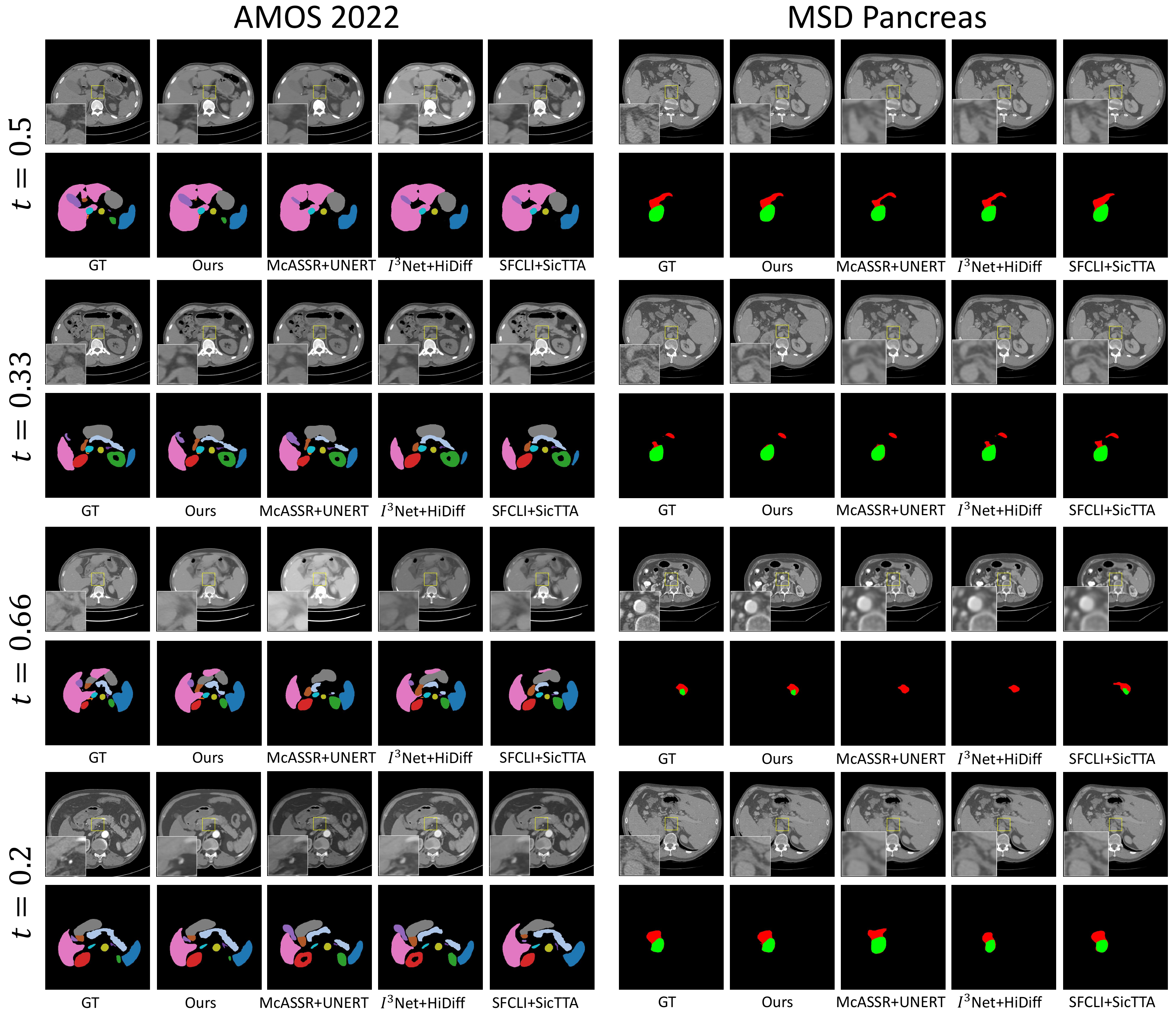}
\caption{Qualitative comparison at different interpolation positions. Left three columns show interpolation results; right three columns show segmentation results.}
\label{fig:position_vis}
\end{figure*}

Qualitative visualization of joint interpolation and segmentation at different positions is shown in Fig. \ref{fig:position_vis}. Compared with cascaded baselines, CoInS-Net preserves sharper anatomical boundaries and more accurate segmentation contours at non-central positions, effectively alleviating structural distortion and label misalignment when the target slice deviates from the midpoint.

\section{Ablation study}

To quantitatively verify the necessity and individual contribution of each core module, we conducted leave-one-out ablation studies, where each variant removed one module from the full model while keeping the same backbone architecture and training hyperparameters for fair comparison. As summarized in Table \ref{tab:ablation_modules}, removing any single component leads to consistent performance degradation across all datasets and both tasks, confirming the indispensability of the proposed design. We ablate the spatially continuous position interpolation module (SCPI), the two directions of the prototype-based task mutual interaction (segmentation-to-interpolation, $\text{PTMI}_{\mathrm{S}\rightarrow\mathrm{R}}$, and interpolation-to-segmentation, $\text{PTMI}_{\mathrm{R}\rightarrow\mathrm{S}}$), and the multi-scale task-cooperative decoder (MTC). The segmentation-to-interpolation prototype path contributes most to interpolation, since its removal causes the largest SSIM drop by withdrawing the anatomical structure that keeps synthesised boundaries plausible. For segmentation, the task-cooperative decoder is particularly important, as removing it produces the largest Dice reduction due to degraded fine-grained boundary modeling. Overall, the full model achieves the best performance on all metrics, demonstrating that the three modules jointly provide complementary benefits for the proposed framework.

\begin{table*}[!tbp]
\centering
\fontsize{7}{9}\selectfont
\setlength{\tabcolsep}{3pt}
\caption{Ablation Studies of Core Modules on Four Medical Image Datasets}\label{tab:ablation_modules}
\begin{tabular}{@{}lcccccccc@{}}
\toprule
\multirow{2}{*}{Method} & \multicolumn{2}{c}{MSD Heart} & \multicolumn{2}{c}{MSD Pancreas} & \multicolumn{2}{c}{AMOS2022} & \multicolumn{2}{c}{BraTS 2024} \\
\cmidrule(lr){2-3} \cmidrule(lr){4-5} \cmidrule(lr){6-7} \cmidrule(lr){8-9}
& SSIM & Dice & SSIM & Dice & SSIM & Dice & SSIM & Dice \\
\midrule
$\mathrm{w/o\ SCPI}$
& 0.968$\pm$0.0031$^{***}$ & 0.963$\pm$0.014
& 0.951$\pm$0.0016$^{***}$ & 0.969$\pm$0.018
& 0.951$\pm$0.0038$^{**}$ & 0.859$\pm$0.022
& 0.941$\pm$0.0046$^{**}$ & 0.834$\pm$0.025 \\
$\mathrm{w/o\ PTMI}_{\mathrm{S}\rightarrow\mathrm{R}}$
& 0.966$\pm$0.0044$^{**}$ & 0.964$\pm$0.020
& 0.949$\pm$0.0021$^{***}$ & 0.970$\pm$0.024
& 0.949$\pm$0.0052$^{***}$ & 0.861$\pm$0.031
& 0.939$\pm$0.0063$^{*}$ & 0.836$\pm$0.034 \\
$\mathrm{w/o\ PTMI}_{\mathrm{R}\rightarrow\mathrm{S}}$
& 0.969$\pm$0.0026$^{*}$ & 0.962$\pm$0.011
& 0.952$\pm$0.0011$^{***}$ & 0.968$\pm$0.015
& 0.952$\pm$0.0030$^{***}$ & 0.858$\pm$0.019$^{**}$
& 0.942$\pm$0.0037$^{***}$ & 0.833$\pm$0.023 \\
$\mathrm{w/o\ MTC}$
& 0.970$\pm$0.0037 & 0.959$\pm$0.017
& 0.953$\pm$0.0014$^{***}$ & 0.965$\pm$0.021
& 0.953$\pm$0.0045 & 0.854$\pm$0.027
& 0.943$\pm$0.0055$^{*}$ & 0.829$\pm$0.030 \\

Ours 
& \textbf{0.975$\pm$0.0013} & \textbf{0.968$\pm$0.003} 
& \textbf{0.958$\pm$0.0005} & \textbf{0.974$\pm$0.006} 
& \textbf{0.958$\pm$0.0021} & \textbf{0.867$\pm$0.009} 
& \textbf{0.949$\pm$0.0023} & \textbf{0.842$\pm$0.011} \\
\bottomrule
\end{tabular}
\end{table*}

\subsection{Multi-Task Learning Architecture Comparison}

To further validate the superiority of our multi-task design paradigm, we compared five classic multi-task learning architectures on the MSD Heart dataset. All variants shared the same Swin Transformer backbone and training hyperparameters for fair comparison.

Five representative multi-task learning schemes were compared, including Direct Feature Concatenation\cite{caruana1997multitask}, Shared-Bottom\cite{ruder2017overview}, Cross-Stitch\cite{misra2016crossstitch}, Mixture of Experts\cite{ma2018modeling}, and Progressive Layered Extraction\cite{tang2020progressive}. Quantitative results are shown in Table \ref{tab:mtl_arch}.

Among all compared architectures, Progressive Layered Extraction achieves the best performance among classic methods with 36.21 dB PSNR and 0.941 Dice, benefiting from its layered expert extraction mechanism that alleviates task conflict. Our CoInS-Net further outperforms this scheme by 1.67 dB in PSNR and 2.7 percentage points in Dice, which is attributed to the fine-grained position modulation and bidirectional prototype-based interaction mechanisms that fully exploit the intrinsic correlation between interpolation and segmentation tasks.

\begin{table}[!tbp]
\centering
\fontsize{7}{9}\selectfont
\setlength{\tabcolsep}{0.5pt}
\caption{Comparison of Multi-Task Learning Architectures}\label{tab:mtl_arch}
\begin{tabular}{@{}lllll@{}}
\toprule
Architecture & PSNR (dB) & SSIM & Dice & Recall \\
\midrule
Direct Concat & 34.38$\pm$0.34$^{***}$ & 0.933$\pm$0.0045$^{***}$ & 0.907$\pm$0.019$^{***}$ & 0.910$\pm$0.020$^{***}$ \\
Shared-Bottom & 35.71$\pm$0.26$^{***}$ & 0.949$\pm$0.0035$^{***}$ & 0.934$\pm$0.015$^{***}$ & 0.937$\pm$0.016$^{**}$ \\
Cross-Stitch & 35.24$\pm$0.30$^{***}$ & 0.944$\pm$0.0040$^{***}$ & 0.923$\pm$0.017$^{**}$ & 0.926$\pm$0.018$^{***}$ \\
MoE & 35.89$\pm$0.23$^{***}$ & 0.951$\pm$0.0031$^{***}$ & 0.937$\pm$0.013$^{**}$ & 0.940$\pm$0.014$^{***}$ \\
PLE & 36.21$\pm$0.18$^{***}$ & 0.955$\pm$0.0024$^{***}$ & 0.941$\pm$0.010$^{***}$ & 0.944$\pm$0.011$^{**}$ \\
Ours & \textbf{37.88$\pm$0.21} & \textbf{0.975$\pm$0.0013} & \textbf{0.968$\pm$0.003} & \textbf{0.970$\pm$0.004} \\
\bottomrule
\end{tabular}
\end{table}

\subsection{Loss Balancing Strategy Comparison}

We further compared four loss balancing strategies: fixed weighted loss, gradient normalization\cite{chen2018gradnorm}, dynamic weight average\cite{liu2019end}, and our learnable uncertainty weighting in Eq.~\eqref{eq:total_loss}. Quantitative results are listed in Table \ref{tab:loss_balance}.

Gradient normalization and dynamic weight average achieve slightly better interpolation performance than the fixed weight strategy, as they dynamically adjust weights according to gradient magnitude or loss decay rate. Their segmentation performance improvement is limited. In contrast, our uncertainty weighting achieves the best comprehensive performance, especially boosting segmentation Dice by 1.8 percentage points compared with fixed weight loss, while maintaining competitive interpolation accuracy. This indicates that adapting the task weights to each task's noise level effectively balances heterogeneous objectives and avoids the model being dominated by the easier interpolation task during training.

\begin{table}[!tbp]
\centering
\fontsize{7}{9}\selectfont
\setlength{\tabcolsep}{0.5pt}
\caption{Comparison of Loss Balancing Strategies}\label{tab:loss_balance}
\begin{tabular}{@{}lllll@{}}
\toprule
Strategy & PSNR (dB) & SSIM & Dice & Recall \\
\midrule
Fixed Weight & 37.37$\pm$0.23$^{**}$ & 0.970$\pm$0.0031$^{**}$ & 0.950$\pm$0.013 & 0.953$\pm$0.014$^{***}$ \\
GradNorm & 37.63$\pm$0.17$^{*}$ & 0.972$\pm$0.0023$^{*}$ & 0.956$\pm$0.010$^{*}$ & 0.959$\pm$0.011$^{*}$ \\
DWA & 37.69$\pm$0.14 & 0.973$\pm$0.0019$^{*}$ & 0.959$\pm$0.008$^{**}$ & 0.962$\pm$0.009 \\
Ours & \textbf{37.88$\pm$0.21} & \textbf{0.975$\pm$0.0013} & \textbf{0.968$\pm$0.003} & \textbf{0.970$\pm$0.004} \\
\bottomrule
\end{tabular}
\end{table}

\section{Model complexity}

To further verify the practical deployment value of the proposed CoInS-Net, we conducted comprehensive computational efficiency experiments on the MSD Heart dataset with a batch size of 4 and input size of 512 $\times$ 512. We compared with six groups of combined models formed by pairing state-of-the-art interpolation and segmentation models. Quantitative results are summarized in Table \ref{tab:efficiency}.

\begin{table*}[!tbp]
\centering
\fontsize{7}{9}\selectfont
\setlength{\tabcolsep}{5pt}
\caption{Computational Efficiency Comparison on MSD Heart MRI}
\label{tab:efficiency}
\begin{tabular}{@{}lllllll@{}}
\toprule
Model & Avg Infer Time (ms) & Avg FPS & Parameters (M) & Memory (MB) & FLOPs (G) & Model Size (MB) \\
\midrule
RCAN+UNet++ & 117.83$\pm$1.47 & 16.98$\pm$0.23 & 162.74$\pm$0.00 & 1396.54$\pm$14.73 & 386.92$\pm$0.00 & 1248.37$\pm$0.00 \\
SAINT+nnU-Net & 108.46$\pm$1.13 & 18.44$\pm$0.19 & 148.39$\pm$0.00 & 1248.76$\pm$12.58 & 342.67$\pm$0.00 & 1095.82$\pm$0.00 \\
McASSR+Swin UNETR & 96.72$\pm$0.91 & 20.68$\pm$0.17 & 134.86$\pm$0.00 & 1086.43$\pm$10.92 & 304.58$\pm$0.00 & 918.64$\pm$0.00 \\
ACVTT+SegMamba-V2 & 88.35$\pm$0.74 & 22.64$\pm$0.15 & 119.57$\pm$0.00 & 924.71$\pm$8.67 & 268.34$\pm$0.00 & 736.29$\pm$0.00 \\
I$^3$Net+HiDiff & 79.28$\pm$0.63 & 25.23$\pm$0.13 & 103.42$\pm$0.00 & 746.38$\pm$7.21 & 231.76$\pm$0.00 & 512.53$\pm$0.00 \\
CycleINR+SicTTA & 72.64$\pm$0.52 & 27.53$\pm$0.11 & 91.85$\pm$0.00 & 612.47$\pm$5.86 & 189.43$\pm$0.00 & 386.71$\pm$0.00 \\
Ours & \textbf{37.70$\pm$0.22} & \textbf{53.09$\pm$0.28} & \textbf{42.98$\pm$0.00} & \textbf{408.92$\pm$4.61} & \textbf{118.23$\pm$0.00} & \textbf{163.96$\pm$0.00} \\
\bottomrule
\end{tabular}
\end{table*}

CoInS-Net achieves an average inference time of 37.70 milliseconds per frame, reducing inference latency by 48.1 percent compared with the best combined model. The average inference frame rate reaches 53.09 FPS, meeting the real-time processing requirements of clinical medical image analysis. In terms of model complexity, our model has only 42.98M parameters, 118.23G FLOPs, and 163.96MB model size, all of which are significantly lower than all combined models. For memory consumption during inference, CoInS-Net only occupies 408.92 MB of GPU memory, which is 33.2 percent lower than the best combined model. The significant computational efficiency advantage benefits from the shared encoder architecture that avoids redundant feature extraction, as well as the sparse token interaction mechanism that reduces unnecessary computational overhead, making our model more suitable for deployment in clinical environments with limited hardware resources.

\section{Discussion}

This study introduces CoInS-Net, an end-to-end single-stage multi-task network for joint medical image interpolation and segmentation. The model achieves state-of-the-art performance on four mainstream medical imaging datasets covering cardiac MRI, abdominal CT, cross-modal abdominal CT/MRI, and intracranial glioma MRI modalities. The proposed framework fully exploits the intrinsic complementarity between interpolation and segmentation tasks via a shared encoder-dual decoder architecture, combined with continuous position interpolation and bidirectional prototype-based interaction mechanisms, achieving significant performance gains over both single-task baselines and classic multi-task learning architectures.

The core technical innovations of CoInS-Net address three key challenges in joint medical image interpolation and segmentation. First, the spatially continuous position interpolation module solves the arbitrary-position modeling problem for interpolated frames, generating a physical-spacing-aware position prior and correcting a linear feature blend through a gated residual at every scale. Unlike fixed-frequency positional encoding used in existing methods, this module incorporates the physical spacing of anisotropic volumes and guarantees exact endpoint recovery by construction, significantly improving the accuracy of lesion boundary reconstruction in interpolated frames. Second, the prototype-based task mutual interaction module establishes lightweight bidirectional communication. Segmentation prototypes constrain interpolation to preserve structural integrity, while interpolation prototypes convey inter-slice transition cues that guide segmentation to perceive z-axis deformation. Because the two branches meet only through a few prototypes, the module introduces negligible computational overhead while delivering consistent performance gains. Third, the multi-scale task-cooperative decoder selectively routes cross-task spatial information through learned sigmoid gates at each upsampling scale, refining local details and boundaries without extra supervision. Ablation studies show that these components contribute consistently to both tasks, especially improving the boundary-sensitive recall metric.

\begin{figure*}[!tbp]
    \centering
    \setlength{\tabcolsep}{2pt}
    \subfigure[Fold 1 training loss]{\includegraphics[width=0.19\textwidth]{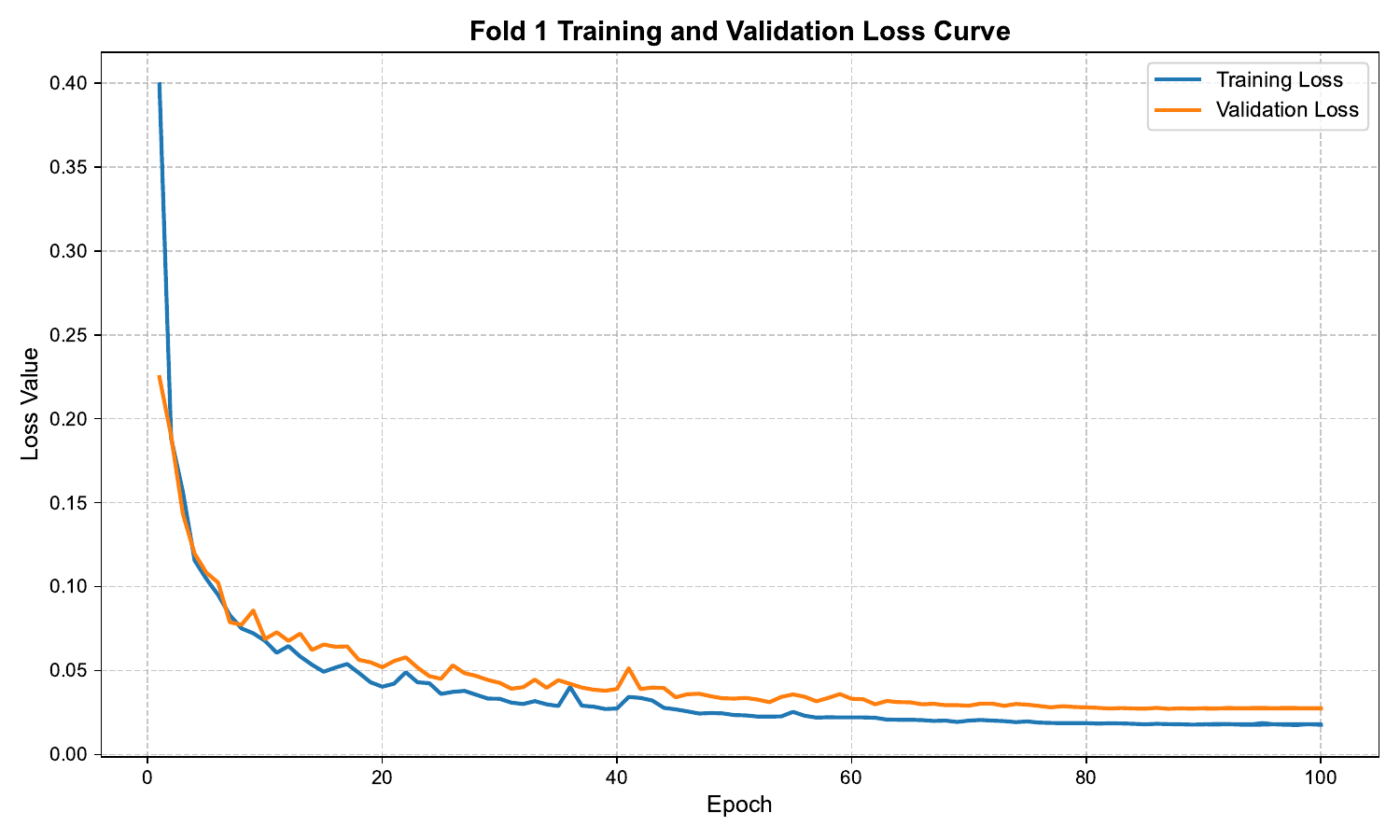}\label{fig:conv_train_1}}
    \hfill
    \subfigure[Fold 2 training loss]{\includegraphics[width=0.19\textwidth]{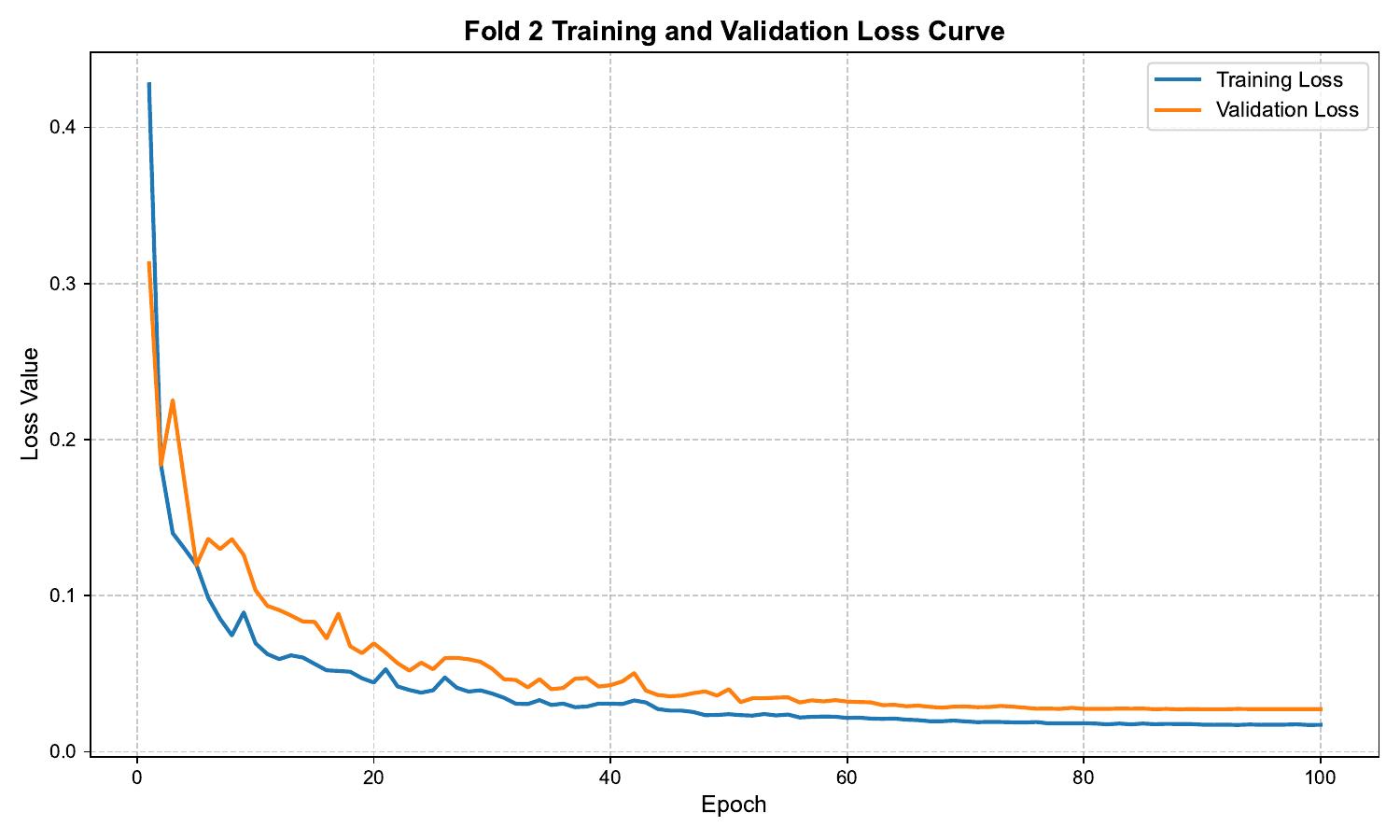}\label{fig:conv_train_2}}
    \hfill
    \subfigure[Fold 3 training loss]{\includegraphics[width=0.19\textwidth]{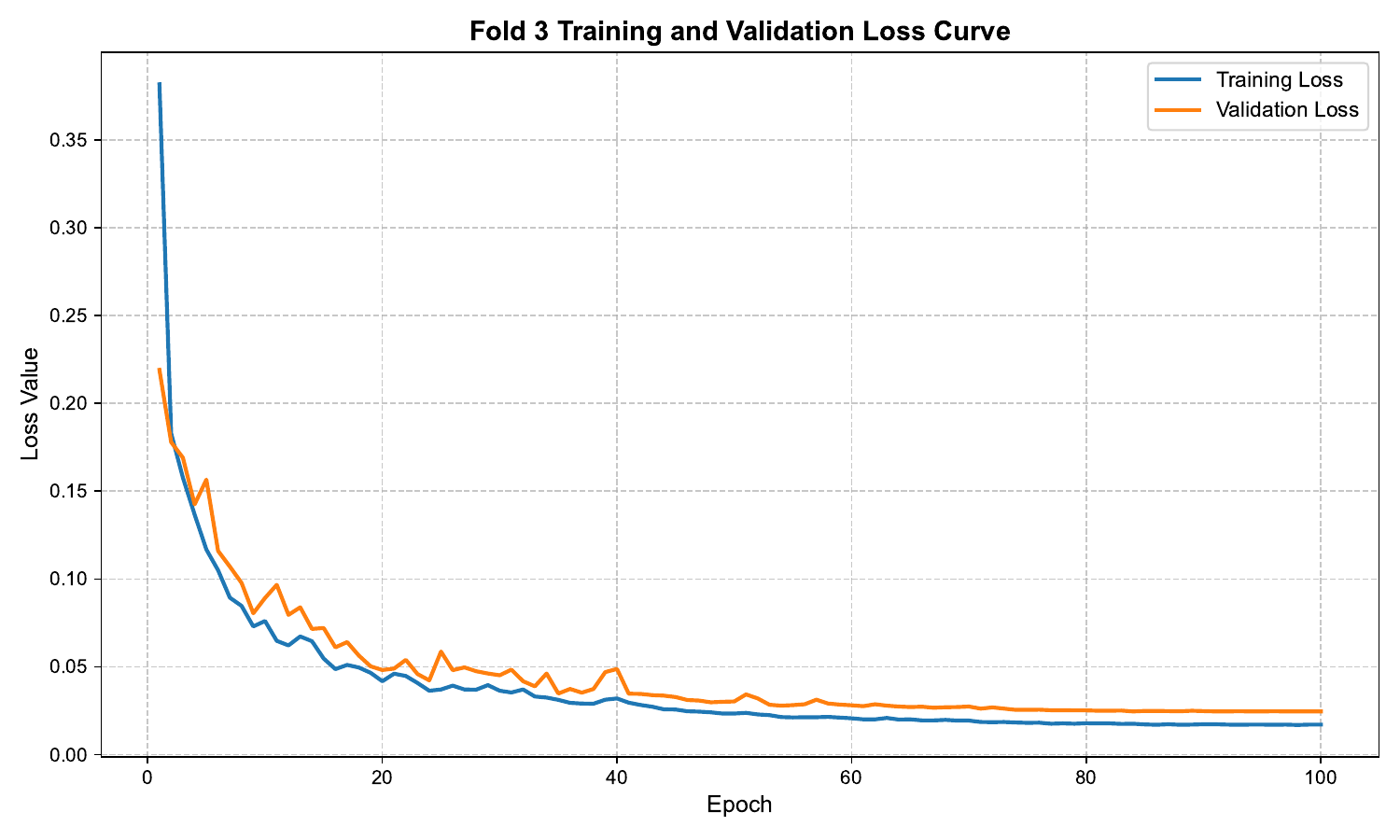}\label{fig:conv_train_3}}
    \hfill
    \subfigure[Fold 4 training loss]{\includegraphics[width=0.19\textwidth]{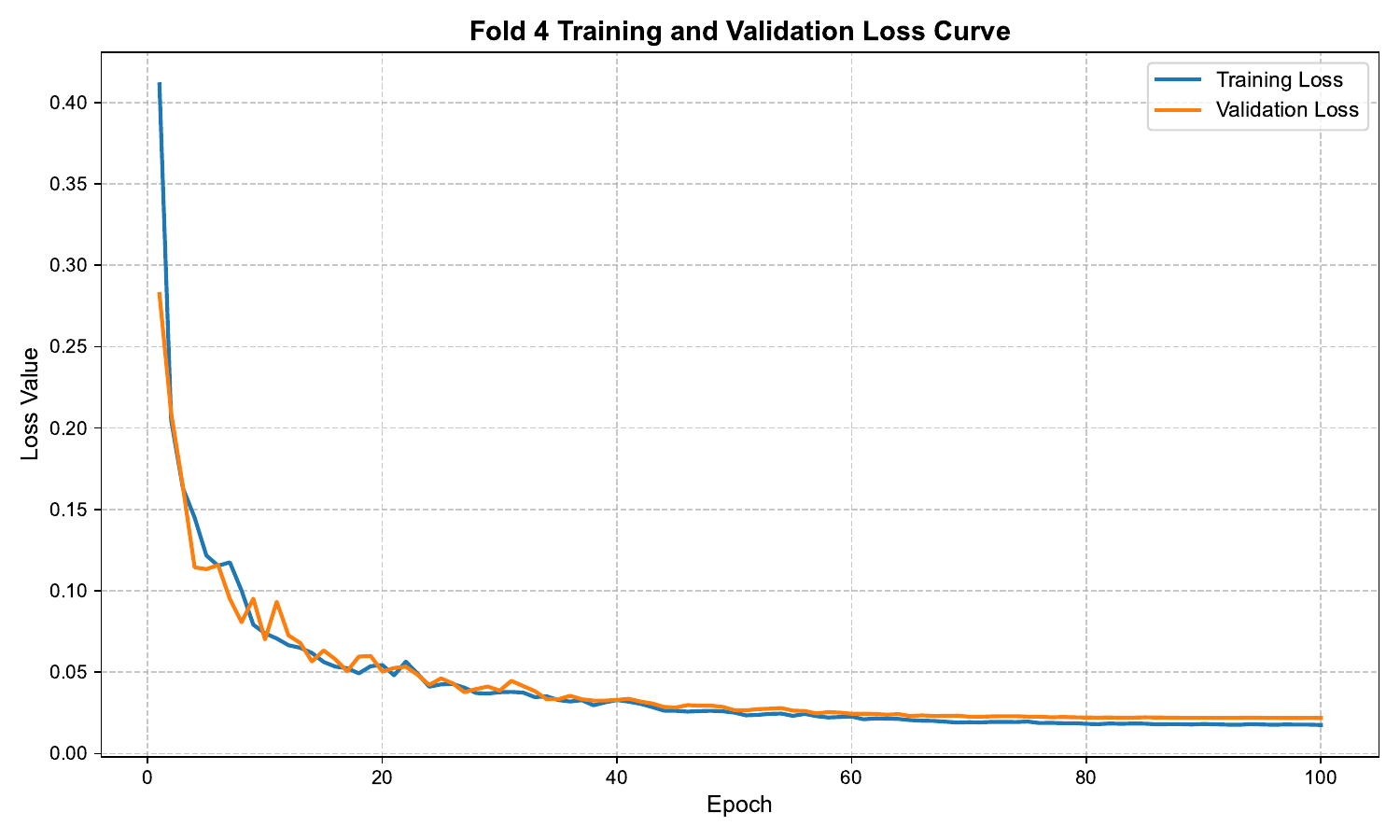}\label{fig:conv_train_4}}
    \hfill
    \subfigure[Fold 5 training loss]{\includegraphics[width=0.19\textwidth]{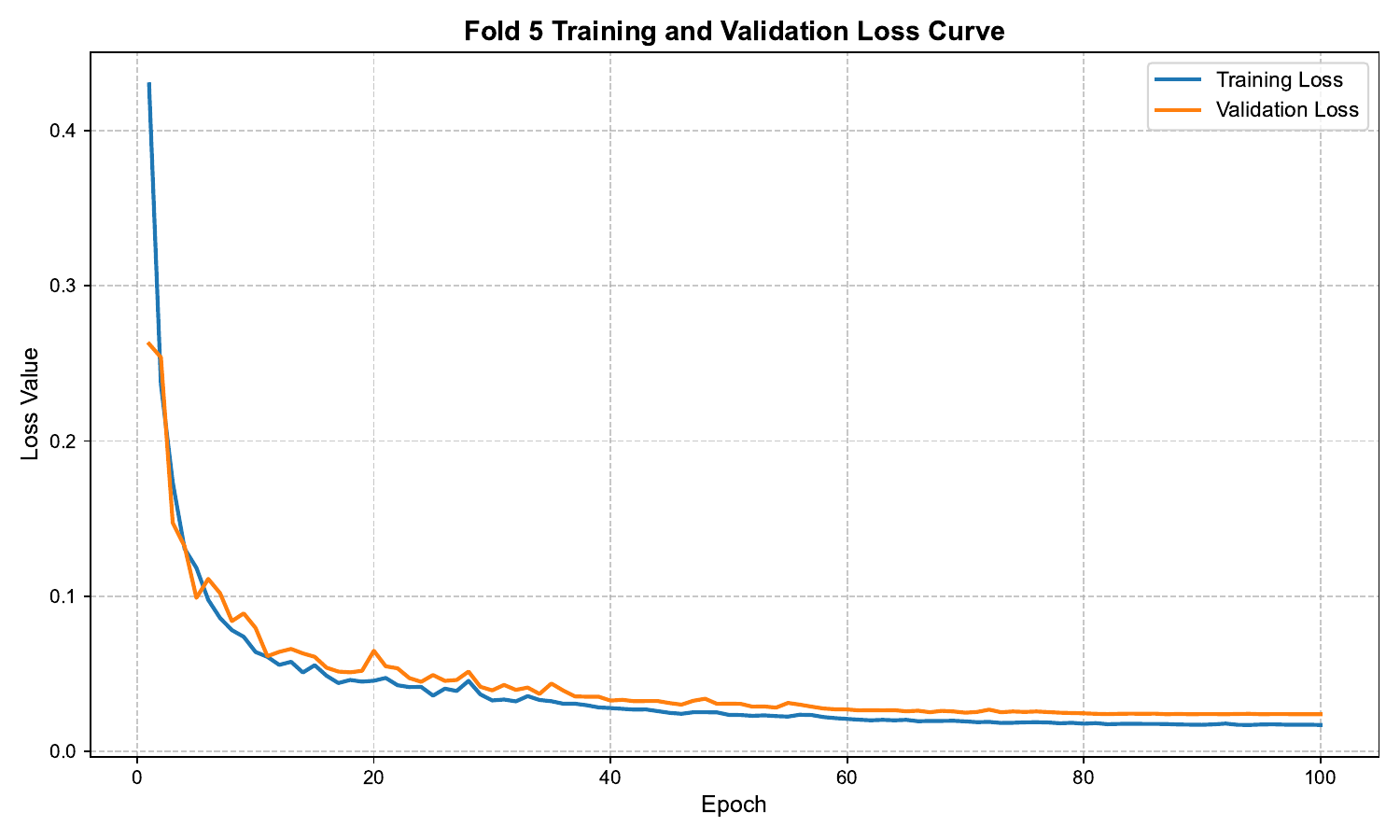}\label{fig:conv_train_5}}
    \\[1pt]
    \subfigure[Fold 1 validation metrics]{\includegraphics[width=0.19\textwidth]{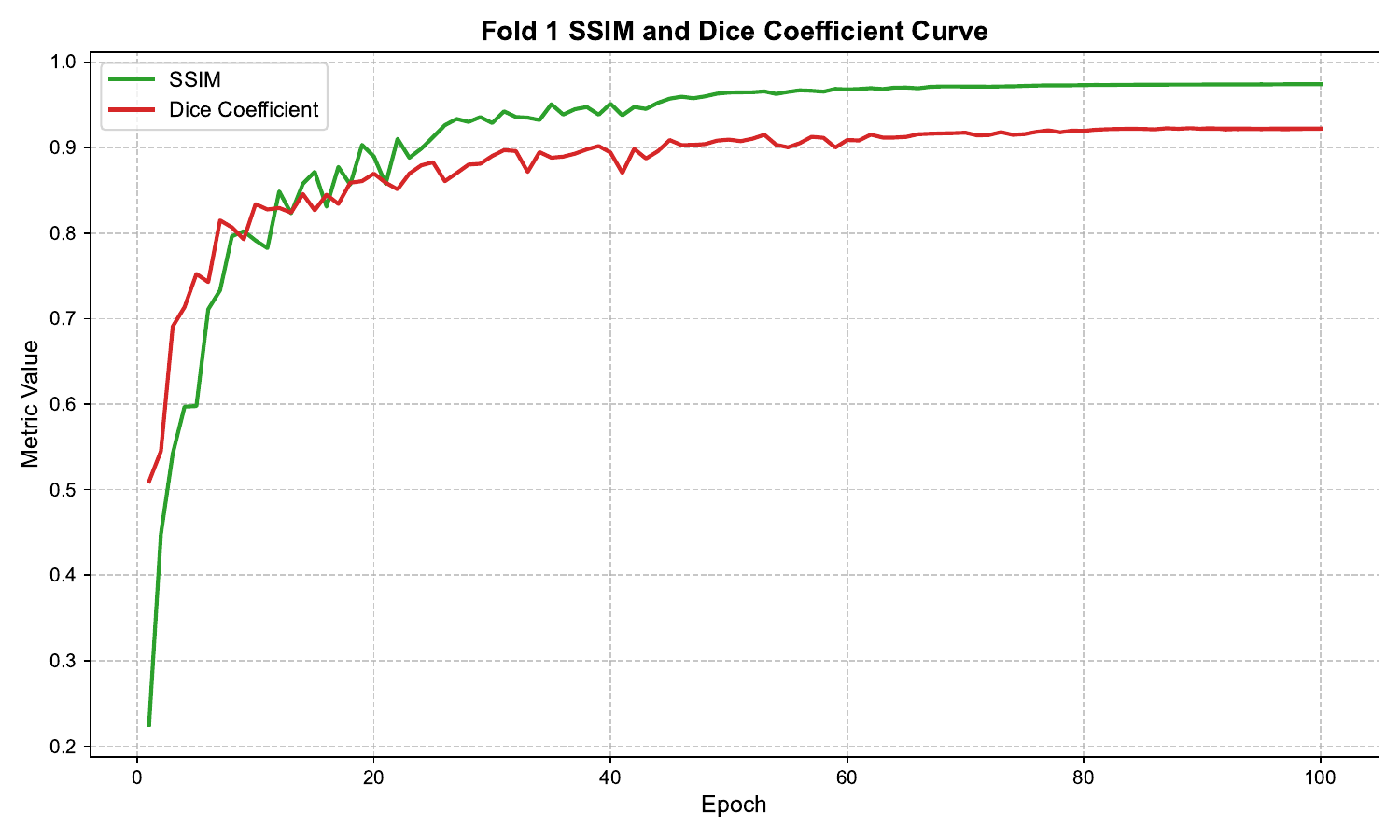}\label{fig:conv_val_1}}
    \hfill
    \subfigure[Fold 2 validation metrics]{\includegraphics[width=0.19\textwidth]{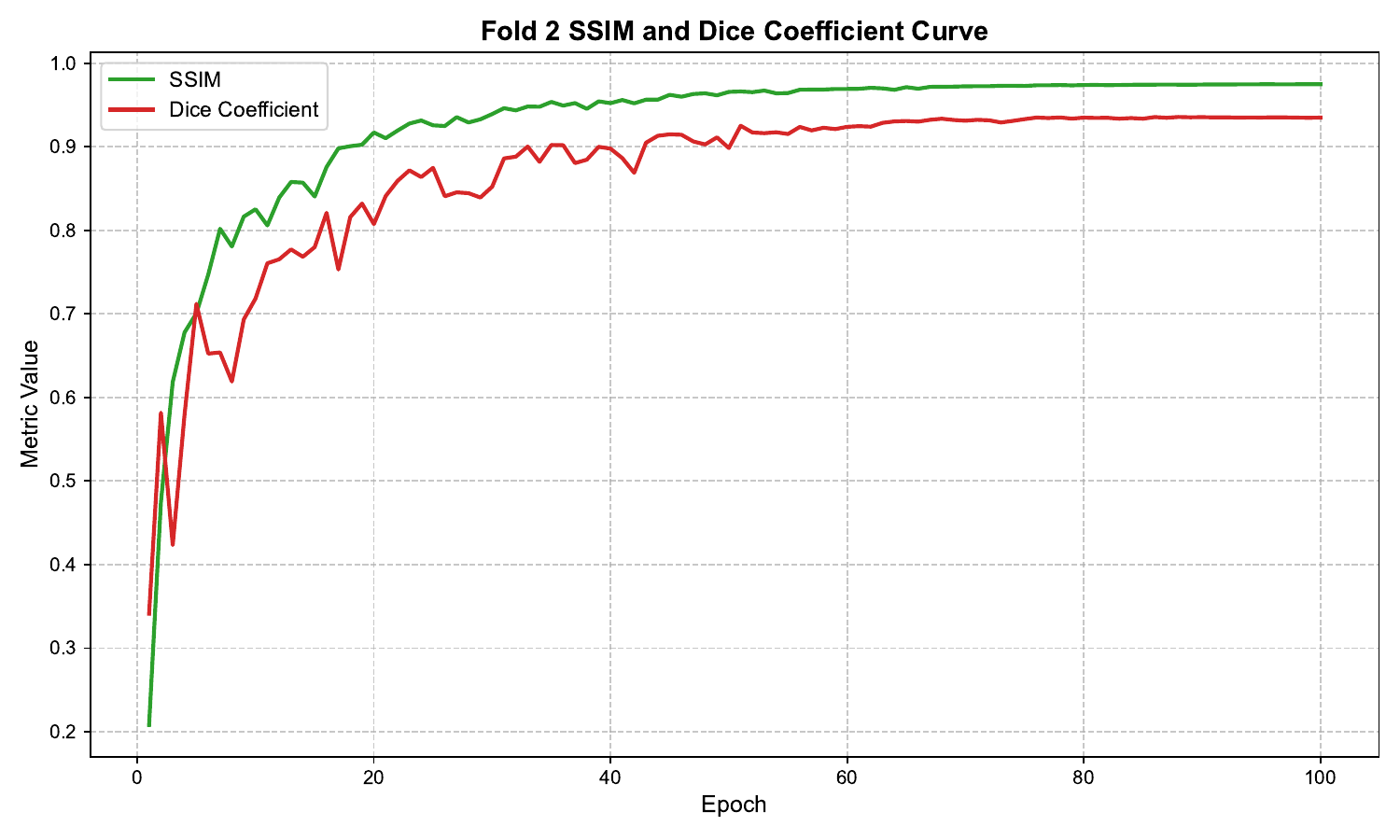}\label{fig:conv_val_2}}
    \hfill
    \subfigure[Fold 3 validation metrics]{\includegraphics[width=0.19\textwidth]{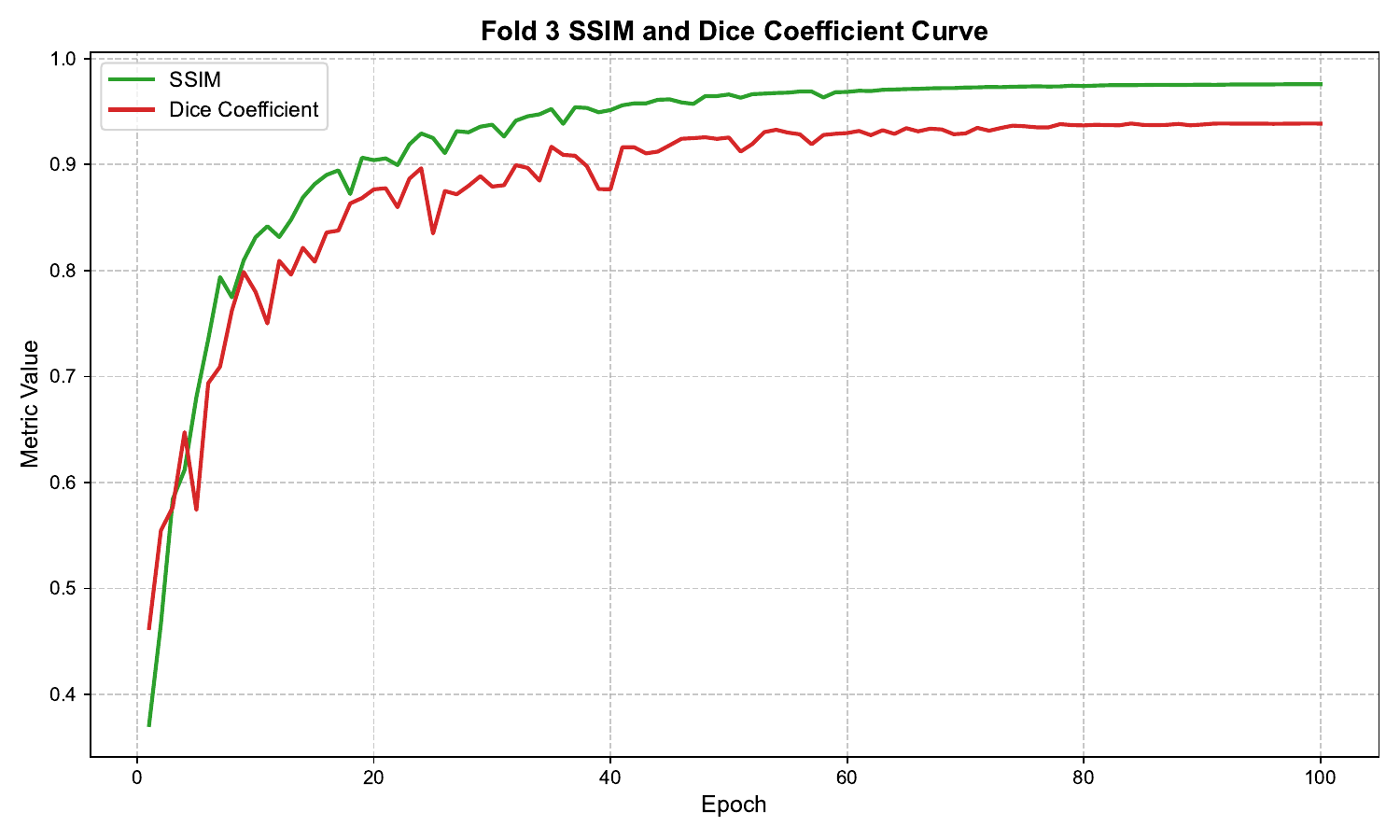}\label{fig:conv_val_3}}
    \hfill
    \subfigure[Fold 4 validation metrics]{\includegraphics[width=0.19\textwidth]{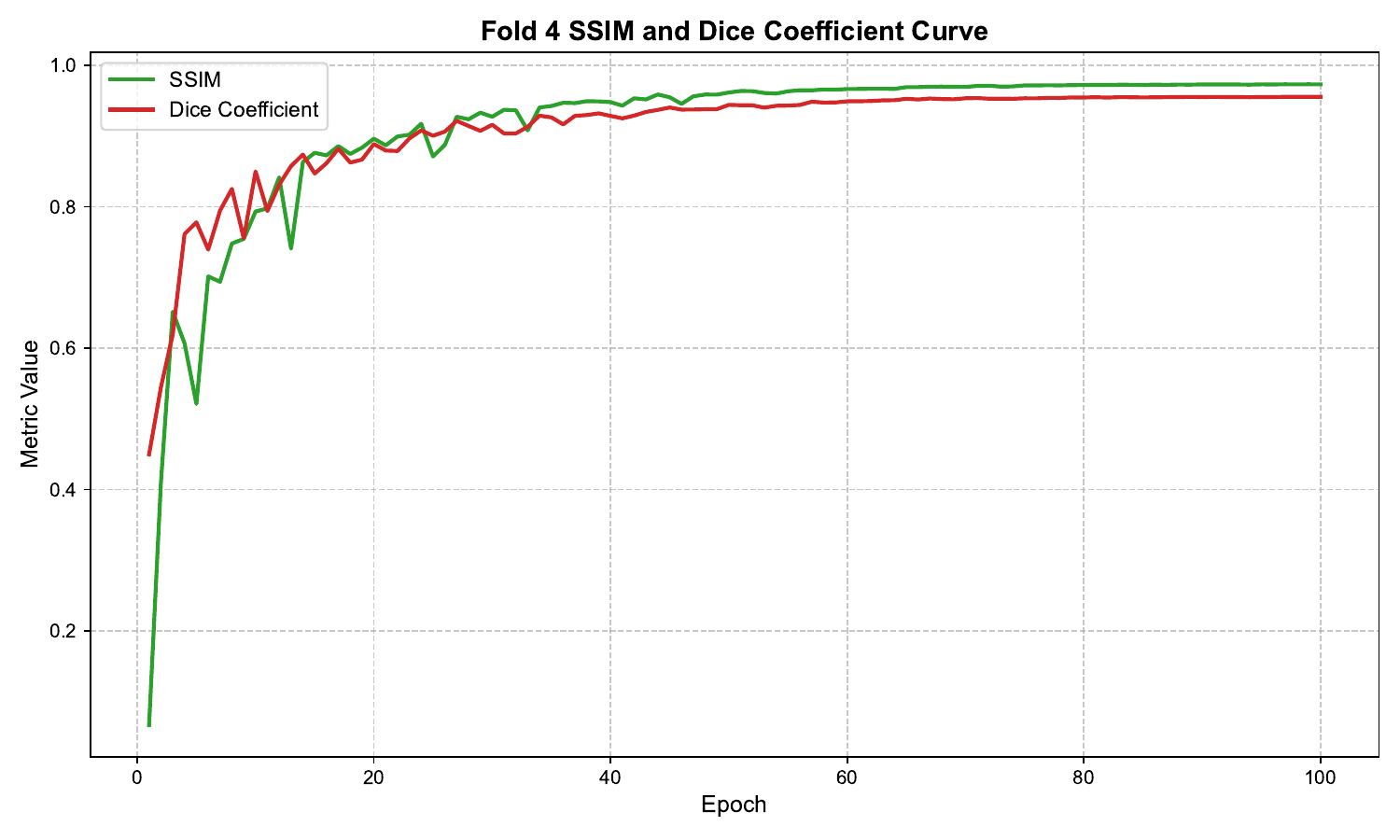}\label{fig:conv_val_4}}
    \hfill
    \subfigure[Fold 5 validation metrics]{\includegraphics[width=0.19\textwidth]{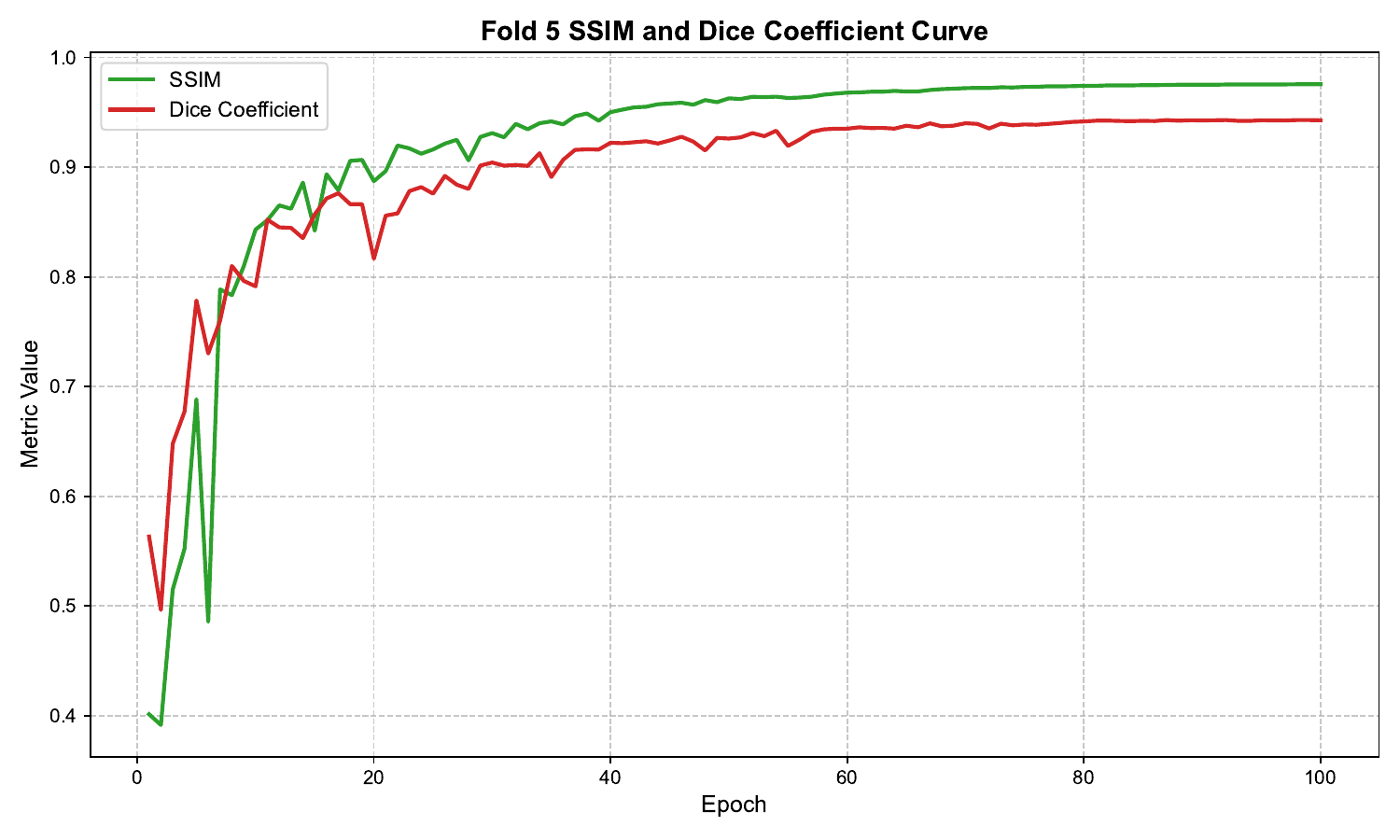}\label{fig:conv_val_5}}
    \caption{Multi-task training convergence curves of five-fold cross-validation on the MSD Heart dataset. (a)--(e) Total training loss curves across the five folds; (f)--(j) Validation curves of interpolation SSIM and segmentation Dice across the five folds.}
    \label{fig:convergence}
\end{figure*}

Figure \ref{fig:convergence} displays five-fold cross-validation training curves on the MSD Heart dataset. The multi-task total training loss steadily declines and converges while validation interpolation SSIM and segmentation Dice metrics keep rising and stabilize at high values throughout model training.

From a clinical standpoint, the joint optimization framework brings substantial practical value. The improved boundary precision of both interpolated images and segmentation masks benefits radiotherapy planning and longitudinal monitoring, where small contour deviations may affect target definition and dosimetry. The reduced inference latency and memory footprint also enable deployment on resource-limited clinical workstations, facilitating routine clinical workflow integration. The strong performance across diverse anatomical regions and modalities further suggests broad applicability for different clinical scenarios.

Nevertheless, this study has several limitations. First, all evaluations are conducted on publicly available benchmark datasets, which do not fully represent the diversity of real-world clinical data acquired under varying scanning protocols and vendor distributions. External validation on independent, institution-specific clinical datasets is required to further confirm generalizability. Second, the current framework operates on 2D slices, which does not fully exploit 3D volumetric context for spatial consistency. Future work will extend the architecture to pure 3D volumetric processing to further enhance inter-slice continuity. Third, the model performance on extremely small lesions remains inferior to that on medium and large lesions, consistent with known limitations of overlap-based metrics for small volumes. Future studies could incorporate size-aware sampling strategies or uncertainty estimation to improve performance on low-volume cases.

\section{Conclusion}

This paper presents CoInS-Net, an advanced multi-task framework for joint medical image interpolation and segmentation that integrates spatially continuous position interpolation, prototype-based bidirectional task interaction, and a multi-scale task-cooperative decoder. The results from extensive evaluation on four diverse medical imaging benchmarks showcase the model’s superiority in terms of both reconstruction accuracy and segmentation precision, making it a significant advancement in multi-task medical image analysis. CoInS-Net consistently outperforms existing single-task methods across various anatomical regions and modalities, demonstrating its ability to exploit cross-task complementarity and providing clinically relevant image analysis results. The framework’s robust performance across diverse datasets, coupled with its high computational efficiency, indicates potential for real-world clinical deployment. Future work will focus on extending the model to 3D volumetric processing and validating performance on multi-center clinical datasets.

% \section*{CRediT authorship contribution statement}
% Yujia Sun: Writing – original draft, Visualization, Validation, Methodology, Funding acquisition, Conceptualization.
% Peiting Shi: Writing – original draft, Visualization, Software, Data curation, Conceptualization.
% Ningfeng Que: Writing – review \& editing, Validation, Resources.
% Rongrong Fu: Validation, Supervision, Project administration, Formal analysis.
% Yingying Yang: Writing – original draft, Visualization, Software, Methodology.
% Xinhang Li: Supervision, Funding acquisition, Conceptualization.

\section*{Declaration of competing interest}
The authors declare that they have no known competing financial interests or personal relationships that could have appeared to influence the work reported in this paper.

\section*{Acknowledgments}
This work was not supported by any fundation.

\bibliographystyle{unsrt}
\bibliography{references}

\end{document}